\documentclass[10pt,twocolumn,letterpaper]{article}

\usepackage{cvpr}              

\newcommand{\change}[1]{#1}

\usepackage{overpic}
\newcommand{\myparagraph}[1]{\noindent\textbf{#1}\quad}

\newcommand{\themethod}[0]{\textsc{LiMo}\xspace}

\DeclareMathOperator*{\argmin}{arg\,min}

\definecolor{color1}{HTML}{264653}

\definecolor{color2}{HTML}{2a9d8f}

\definecolor{color3}{HTML}{e9c46a}

\definecolor{color4}{HTML}{f4a261}

\definecolor{color5}{HTML}{e76f51}

\newcommand{\best}[1]{\colorbox{red!25}{#1}}

\newcommand{\second}[1]{\colorbox{orange!25}{#1}}

\newcommand{\third}[1]{\colorbox{yellow!25}{#1}}

\definecolor{ligreen}{RGB}{0,128,0}

\definecolor{cvprblue}{rgb}{0.21,0.49,0.74}
\usepackage[pagebackref,breaklinks,colorlinks,allcolors=cvprblue]{hyperref}

\usepackage{microtype}
\usepackage{multirow}
\usepackage{color}
\usepackage{siunitx}
\usepackage{xcolor}
\usepackage{rotating}
\newcommand{\rot}[1]{\begin{turn}{90}#1\enspace\end{turn}}
\usepackage{capt-of}
\usepackage{float}
\usepackage{colortbl}
\usepackage{array}
\usepackage[super]{nth}

\usepackage{pifont}

\def\paperID{18241} 
\def\confName{CVPR}
\def\confYear{2026}

\title{Lighting in Motion: Spatiotemporal HDR Lighting Estimation}

\author{Christophe Bolduc$^1$ \quad
Julien Philip$^2$ \quad
Li Ma$^2$ \quad
Mingming He$^2$ \quad \\
Paul Debevec$^2$ \quad
Jean-François Lalonde$^1$ \\[1.0em]
$^1$Université Laval, $^2$Eyeline Labs
}

\begin{document}

\twocolumn[{
\maketitle
\vspace{-10mm} 

\begin{center}
    \includegraphics[width=\linewidth]{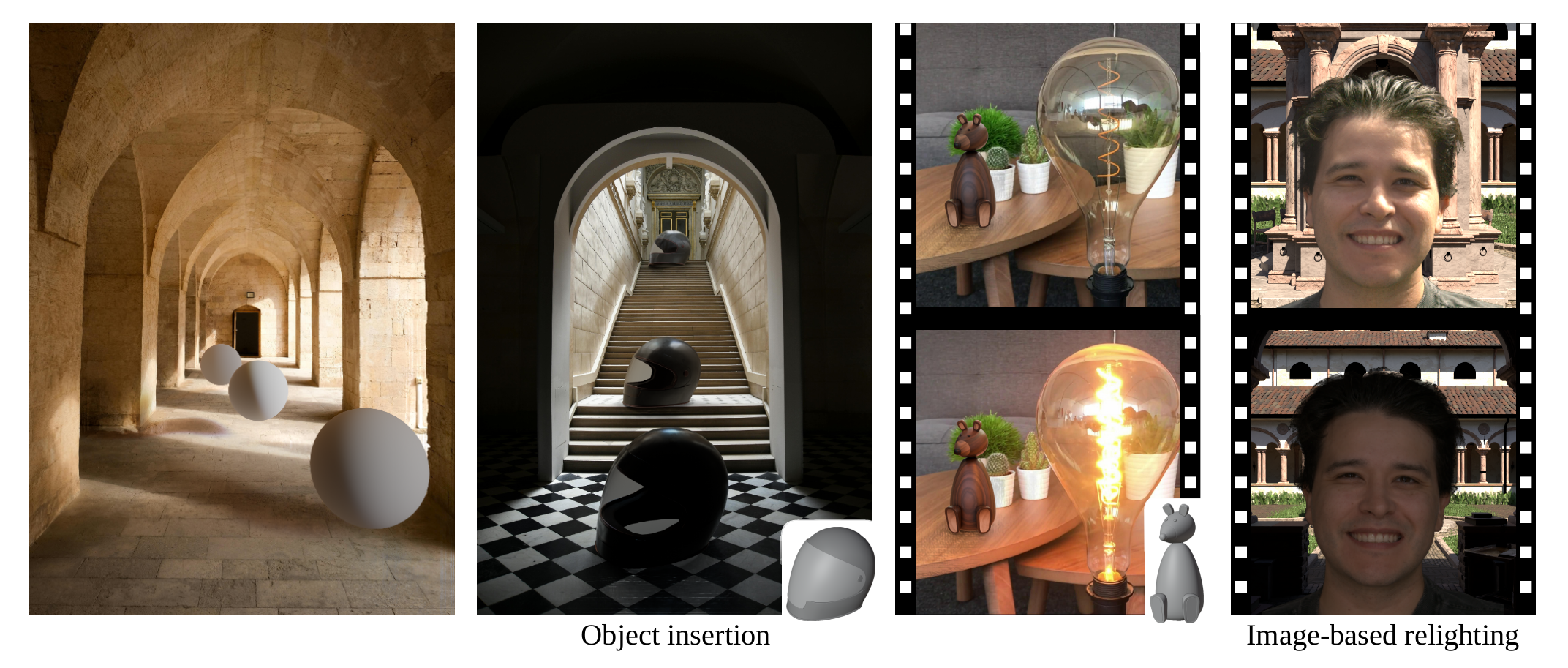}
    \vspace{-6mm}
    \captionof{figure}{We present \themethod: a spatiotemporal lighting estimation method with accurate spatial grounding, full HDR and realistic reflections. \themethod accurately grounds virtual objects over different spatial positions (\nth{1} and \nth{2} left), and over time (\nth{3} left). \themethod can readily be used in virtual production (right), for instance by inserting actors captured in light domes in real sets.}
    \label{fig:teaser}
\end{center}

\vspace{6mm} 
}]


\begin{abstract}
We present \textbf{Li}ghting in \textbf{Mo}tion (\themethod), a diffusion-based approach to spatiotemporal lighting estimation.
\themethod targets both realistic high-frequency detail prediction and accurate illuminance estimation.
To account for both, we propose generating a set of mirrored and diffuse spheres at different exposures, based on their 3D positions in the input.
Making use of diffusion priors, we fine-tune powerful existing diffusion models on a large-scale customized dataset of indoor and outdoor scenes, paired with spatiotemporal light probes.
For accurate spatial conditioning, we demonstrate that depth alone is insufficient and we introduce a new geometric condition to provide the relative position of the scene to the target 3D position.
Finally, we combine diffuse and mirror predictions at different exposures into a single HDRI map leveraging differentiable rendering.
We thoroughly evaluate our method and design choices to establish \themethod as state-of-the-art for both spatial control and prediction accuracy. 
\end{abstract}    
\section{Introduction}
\label{sec:intro}

Humans have the inherent ability to determine whether a virtual object inserted into an image belongs in the scene or not \cite{ferwerda2010perception,10.1145/2810038}.
When an object's shading does not harmonize well with its surroundings, it creates a ``pasted-in'' effect where the object appears out of place, breaking realism~\cite{reinhard2002color}.
Whether the task is to composite an actor into an environment or add an object to an image sequence, having access to accurate lighting information is critical.
Capturing lighting has long been the default solution~\cite{debevec1998rendering}, notably by inserting \emph{light probes}---typically, mirror and diffuse spheres reflecting the incoming light rays towards the camera---in the scene and capturing them using high dynamic range (HDR) photography~\cite{Debevec97}.
However, this process requires physical access to the location, the proper equipment to capture HDR images, and time.
Hence, the ability for computers to automatically estimate lighting given an image, or a sequence of images, has potential impact for virtual and augmented reality, filmmaking, and design. 



We assert that a generally applicable lighting estimation technique should have the five following capabilities:
\begin{enumerate}
    \item Allow for ``grounding'' its prediction in a specific location in the scene, since lighting varies as a function of the relative position to the light sources and occlusions~\cite{gardner2017learning};
    \item Adapt to temporal variations: a moving camera revealing unseen light sources, moving objects causing occlusions, or changing lighting conditions;
    \item Predict accurate HDR luminance values, including for large areas of indirect light reflecting from objects in the scene as well as for concentrated, orders-of-magnitude brighter light sources;
    \item Estimate near-field light sources indoors as well as distant environmental light outside;
    \item Estimate plausible lighting distributions including high-frequency environmental detail and low-frequency directional illuminance, even though estimating such information is typically under-constrained.
\end{enumerate}


Previous methods focused on subsets of these specifications, as shown in \cref{tab:lighting_representation_comparison}.
For example, methods proposed to estimate a single global lighting estimate from images~\cite{gardner2017learning,legendre2019deeplight,Phongthawee2023DiffusionLight} or videos~\cite{liang2025luxdit}.
Others predict spatially-varying lighting but are targeted for indoors~\cite{song2019neural,li2020inverse} or outdoors~\cite{zhu2021spatially}.
Recent works deal with spatiotemporal variations~\cite{li2023spatiotemporally,4DLighting}, but we find they struggle in properly grounding the predictions within the local context of the scene.

In this paper, we present what we believe to be the first approach to effectively address these five capabilities in a single framework.
Our approach provides lighting predictions that can be grounded at a specific 3D position, vary through time, predict accurate HDR values, works both indoors and outdoors, and generates realistic details for reflections.

\newcommand{\gdot}{\textcolor{green!60!black}{\ding{51}}}

\newcommand{\rdot}{\textcolor{red!70!black}{\ding{55}}}

\newcommand{\ydot}{\textcolor{yellow!50!orange}{$\blacktriangle$}}

\setlength{\tabcolsep}{3pt} 

\begin{table}[t]
\centering
\small
\begin{tabular}{lccccc}
\toprule
 &
{\scriptsize\shortstack{1.Ground.}} &
{\scriptsize\shortstack{2.Temp.}} &
{\scriptsize\shortstack{3.HDR}} &
{\scriptsize\shortstack{4.Near}} &
{\scriptsize\shortstack{5.High freq.}} \\
\midrule
Gardner~\etal~\cite{gardner2017learning}   & \rdot & \rdot & \gdot & \rdot & \rdot \\
Garon~\etal~\cite{garon2019fast}           & \ydot & \rdot & \gdot & \gdot & \rdot \\
Irisformer~\cite{zhu2022irisformer}      & \ydot & \rdot & \gdot & \gdot & \rdot \\
LuxDiT~\cite{liang2025luxdit}         & \rdot & \gdot & \gdot & \rdot & \gdot \\
DiffusionLight~\cite{Phongthawee2023DiffusionLight} & \rdot & \rdot & \ydot & \rdot & \gdot \\
4D Lighting~\cite{4DLighting}     & \ydot & \gdot & \ydot & \gdot & \ydot \\
\themethod (Ours)           & \gdot & \gdot & \gdot & \gdot & \gdot \\
\bottomrule
\end{tabular}
\caption{\change{Comparison of different lighting estimation methods on five key capabilities (see text). A ``\ydot'' indicates that the method addresses the corresponding property but with limited effectiveness.}}
    \label{tab:lighting_representation_comparison}
\end{table}

Our method, dubbed \textbf{Li}ghting in \textbf{Mo}tion (\themethod), works as follows: given a monocular image/video and a sequence of positions in the scene, we first use an off-the-shelf predictor~\cite{chen2025video} to recover per-pixel depth. Using the depth and lighting estimation positions we compute a set of geometric maps that are used to condition a diffusion model.
The network, specifically fine-tuned for the task, outputs either a mirror or diffuse sphere at the desired locations, and at a specific exposure value.
By querying the network for multiple combinations of mirror/diffuse spheres and exposure values, we obtain a stack of exposure brackets for both the diffuse and mirror spheres at each position.
These outputs are subsequently fused for each position into a single HDRI, which are combined into a sequence.
To evaluate the method, we present a novel video lighting estimation test dataset made from realistic synthetic data.


Our work makes the following contributions:
\begin{itemize}[leftmargin=2em]
    \item \themethod, a diffusion-based method to predict full HDR lighting at any 3D point, and at any time in a video;
    \item New geometric maps to condition the diffusion-based generator, which we demonstrate are critical for accurate spatially-varying predictions;
    \item A mirror and diffuse multi-bracket approach to lighting estimation which provides both realistic and more physically accurate estimation.
\end{itemize}


\begin{figure*}[t]
    \centering
    \includegraphics[width=0.95\linewidth]{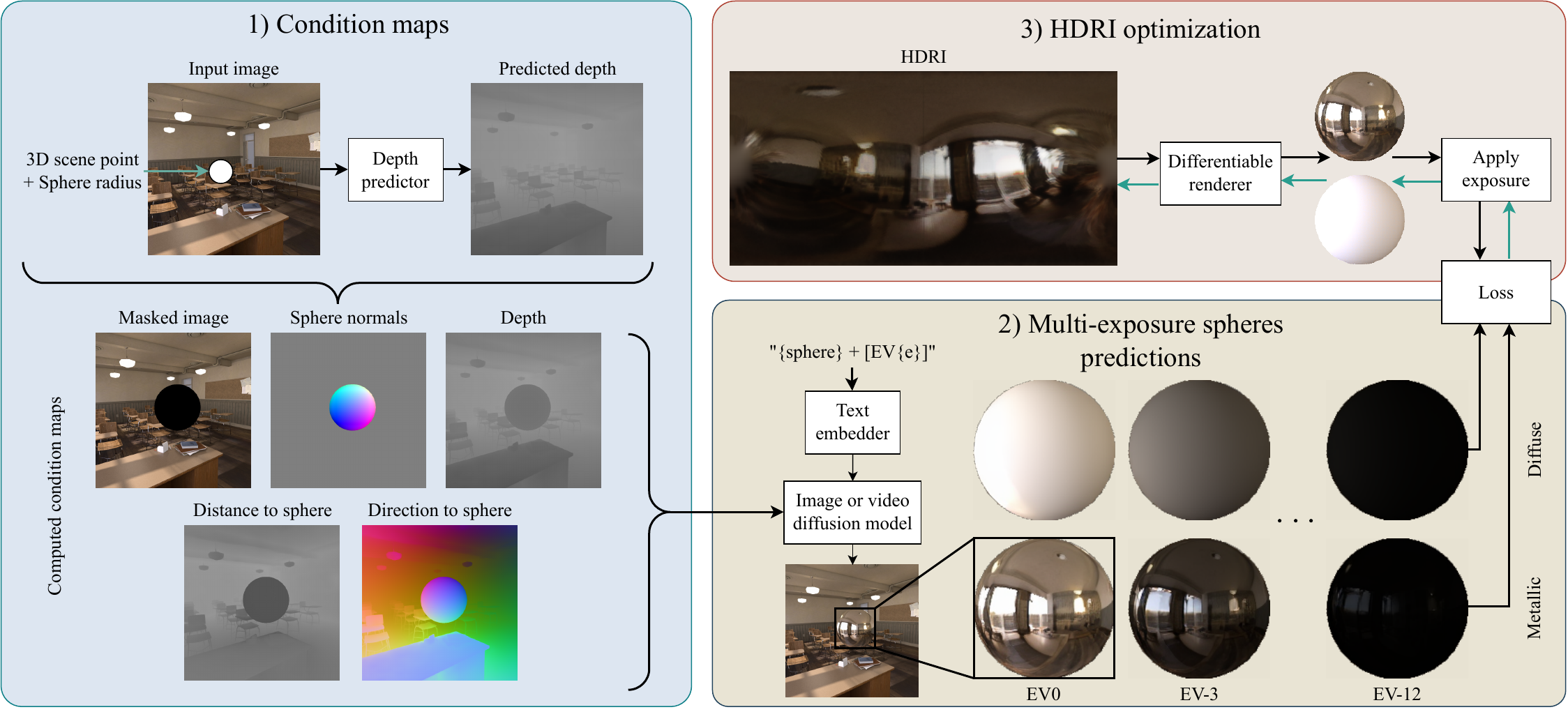}
    \caption{Overview of \themethod, our proposed diffusion-based spatiotemporal lighting estimation method. From an input image (or video sequence) and a 3D scene point, we first obtain an estimate of the per-pixel depth (1, top). From this, a set of condition maps are computed (1, bottom). These maps, along with a text prompt, are used to condition a diffusion model (2) which is trained to inpaint a sphere at the desired 3D scene point. The model learns to predict spheres at different exposures and materials (metallic or diffuse). The predicted spheres are merged into a single HDRI map (3) through a differentiable rendering approach.}
    \label{fig:method_overview}
\end{figure*}

\section{Related work}
\label{sec:related_works}


\myparagraph{Image-based lighting (IBL)} photographs light probes, typically mirror and diffuse spheres~\cite{debevec1998rendering,reinhard2005high}, to construct HDRI maps and render realistic virtual objects. Hardware and physical constraints prohibit the use of such techniques if physical access to the scene is impossible. 

\myparagraph{Single image lighting estimation}
Early approaches estimated environment lighting from images using cues such as reflections, shadows, and geometry~\cite{sato2003illumination,lalonde2012estimating}. Of course, learning-based methods outperformed their predecessors by proposing approaches that directly regress HDR lighting representations from single images of indoor scenes~\cite{gardner2017learning,gardner2019deep,zhan2021emlight}, outdoor scenes~\cite{hold2019deep,zhang2019all}, or both~\cite{legendre2019deeplight,dastjerdi2023everlight,Phongthawee2023DiffusionLight,Chinchuthakun2025DiffusionLightTurbo}, or a scene with a human face \cite{LeGendre:2020:FaceLight}, see \cite{einabadi2021deep} for a recent survey. These approaches typically estimate lighting at a single point in the scene (usually the image center).

\myparagraph{Spatially-varying lighting estimation} 
Since lighting can vary drastically across the field of view, some methods accept a specific location as input, and predict the lighting at that point as an HDRI map~\cite{song2019neural,bai2023local} or spherical harmonics~\cite{garon2019fast}. Others predict a dense lighting representation, for example at each pixel location in a 2D grid of spherical gaussians~\cite{li2020inverse,zhu2022irisformer} or as a volumetric, either using a voxel grid of spherical gaussians~\cite{wang2021learning}, or an implicit representation~\cite{srinivasan2020lighthouse}. Other approaches have been proposed for outdoor scenes~\cite{zhu2021spatially,tang2022estimating}.

\myparagraph{Spatiotemporal lighting estimation}
Lighting can also change over time: flipping a light switch, or panning the camera to a bright window, create drastic changes in lighting that can be estimated by constructing a spatiotemporal volume~\cite{li2023spatiotemporally}. Concurrent to this work, LuxDiT~\cite{liang2025luxdit} predict temporally-varying HDRI maps. Finally, and most closely related to our work, Tong~\etal~\cite{4DLighting} adapt a diffusion model for generating multiple spheres across the field of view, and subsequently build a unique implicit representation over the video through a NeRF-like approach. However, as we will demonstrate, it tends to generate results that are overly smooth and do not properly capture the dynamics of lighting.

\myparagraph{Diffusion-based rendering}
It has been recently shown that virtual object insertion can be performed as a learning task,  through diffusion-based rendering~\cite{zeng2024rgbx,liang2025diffusionrenderer,zhang2025zerocomp} or harmonization~\cite{careaga2023intrinsic,guo2021intrinsic}, bypassing the need for acquiring or estimating an HDR map. While these approaches offer a promising new paradigm for image compositing, they lack the artistic control offered by the traditional IBL pipeline. Here, we focus on explicit HDR lighting recovery in the form of HDRI maps, as they can readily be used in existing, production-ready object compositing frameworks.

\section{Method}

We present \themethod, which adapts image and video generative models to predict spatially-varying HDR illumination in a single image, or a video sequence. 
 
\subsection{General approach}
\label{sec:general_approach}


Our approach consists of using priors from a diffusion model to inpaint a diffuse or mirror sphere at a specific 3D position in space, and at a given exposure value.
At test time, the model is run multiple times to generate both diffuse and mirror spheres at multiple exposures, and predictions are subsequently merged to a single HDRI.

Contrary to some recent work \cite{Phongthawee2023DiffusionLight,Chinchuthakun2025DiffusionLightTurbo} that only rely on mirror spheres, we argue that accurate HDR lighting estimation is greatly facilitated by using diffuse spheres.
Mirror spheres are a great way to obtain plausible reflections but the estimation of luminance values for concentrated light sources relies on a very small number of pixels and requires many exposure values \cite{10.1145/1185657.1185687}.
With a diffuse sphere, the energy of concentrated light sources is integrated to moderate exposure levels by the diffuse surface and becomes easier to estimate accurately from fewer exposure levels \cite{diffusemirror}.
\Cref{fig:method_overview} shows an overview of our proposed method.
Next, we elaborate on how we condition the diffusion model (\cref{sec:conditions}), the dataset used for fine-tuning it (\cref{sec:dataset}), and the HDRI reconstruction at test time (\cref{sec:hdri}).

\subsection{Model conditioning}
\label{sec:conditions}

We fine-tune a diffusion model, conditioning it on an input image or video, additional maps, and text. For clarity, we describe what is done for single images, but our approach handles videos by using a video model or applying the image model independently on each image sequentially. 

\myparagraph{RGB and geometry conditioning}
To allow accurate lighting estimation, we curate multiple input maps used as conditioning to the diffusion model.
These input maps are channel-wise concatenated to the input noise and the first layer of the model is expanded to accommodate these extra channels.

First, we provide the RGB image $I_\text{rgb}$ as input.
The region corresponding to the sphere to inpaint is set to black to prevent the background from spilling into the sphere.
Second, we provide the depth of the background image $I_\text{d}$, with the depth of the sphere to inpaint. 

As we will demonstrate in \cref{sec:ablation}, providing only the depth of the scene and sphere as in \cite{4DLighting} is insufficient for accurately grounding the light prediction at a specific 3D point.
We provide three additional maps: a normal map of the sphere $I_\text{n}$, and two novel maps capturing the geometric relations between the scene and the sphere. 

Those geometric maps relate the scene's surfaces to the sphere's position: for a given pixel $i$ in the image, we define the direction $I_{\text{dir},i}$ and distance $I_{\text{dist},i}$ from the 3D point corresponding to $i$ to the sphere as
\begin{equation}
I_{\text{dir},i} =
\begin{cases}
\dfrac{\mathbf{p}_i - \mathbf{c}}{\|\mathbf{p}_i - \mathbf{c}\|}, & \text{if not on sphere}, \\[1.2em]
\mathbf{v}_i - 2(\mathbf{v}_i \cdot \mathbf{n}_i)\,\mathbf{n}_i, & \text{if on sphere},
\end{cases}
\end{equation}
and 
\begin{equation}
    I_{\text{dist},i} = \|\mathbf{p}_i - \mathbf{c}_i\| \,.
\end{equation}
Here, $\mathbf{p}_i$ are the $(x,y,z)$ pixel world coordinates, computed from the pixel depth and view direction vector $\mathbf{v}_i$ obtained using the camera field of view, and $\mathbf{c}$ is the coordinates of the sphere center.
For pixels on the sphere, we use the reflected incoming ray direction, computed using $\mathbf{n}_i$, the sphere surface normal and $\mathbf{v}_i$.
Intuitively, this map allows the model to match the reflected direction at a given pixel on the sphere and points in the scene that are in the same direction from the sphere center. These geometric quantities are expressed in the camera coordinate system.

The image $I_\text{rgb}$ is assumed to be in sRGB color space, and both the depth map $I_d$ and distance to sphere map $I_\text{dist}$ are log-normalized. All maps $\{I_\text{rgb}, I_\text{n}, I_\text{d}, I_\text{dir}, I_\text{dist}\}$ are individually encoded to latent space using the pre-trained VAE encoder, and subsequently channel-wise concatenated. To allow more channels, the patch embedding convolution of the denoising network is adapted by copying and dividing the initial pre-trained weights by the number of added channels. 

At training time, the ground truth per-pixel depth $I_d$ and camera field of view are used to compute the maps. At test time, we rely on pre-trained estimators (\citet{chen2025video} for depth and \citet{wang2025moge2} for field of view).

\myparagraph{Exposure conditioning}
Similar to \cite{bolduc2025gaslight}, we train the model to accept an exposure value (EV) as text prompt, given to the network using the pre-trained text encoder. Contrary to methods which interpolate between 2 extreme exposures~\cite{Phongthawee2023DiffusionLight, 4DLighting}, we found that giving a discrete number of EVs is effective for accurate EV predictions. In practice, we set EVs to $\{0,-3,-6,-9,-12\}$. During training, the corresponding EV image sphere is given as target, keeping regions outside the sphere at the original EV0.

\myparagraph{Multi-sphere predictions}
An additional text prompt is given to condition on which sphere to inpaint (mirror or diffuse). 
Including both the type of sphere and the exposure value, the final text prompt has the form ``\{sphere type\} [EV{value}]'' (e.g., ``Diffuse sphere [EV0]'').

\newlength{\mywidth}

\subsection{Dataset}
\label{sec:dataset}

To train our approach, physically acquiring a dataset of ground truth HDR lighting moving in space in dynamic lighting would require a highly controllable environment. Instead, we turn to synthetic scenes to generate sequences and their associated ground truth illumination. We use Blender~\cite{blender}, paired with BlenderKit~\cite{blenderkit} assets to procedurally generate both indoor and outdoor renderings. More details on this process are available in the supplementary materials.

To render a data sample, we randomly select a scene (indoor or outdoor), select a random camera pose, and first obtain a normal RGB rendering.
We then generate training targets by placing a grid of spheres in the scene, disabling their visibility to both other spheres and the scene. They are equally-spaced and sized in image space and of random depth.
Even though our training is done on a single sphere, packing a sphere grid provides multiple data points per render and viewpoints, allowing for a random sphere selection as training sample.
Each sphere depth is computed using
\begin{equation}
    \delta = d_\text{min} + (d_\text{max} - d_\text{min})u^{\alpha},
\end{equation}
with $d_\text{min}=0.25$, $d_\text{max}=0.98$, $\alpha=0.4$ and $u \sim \mathcal{U}(0,1)$.
To obtain the scene scale depth for each sphere, the sampled depth factor is multiplied by the scene's minimum depth over the area of the sphere:
\begin{equation}
    d_\text{sph} = \delta \min(I_d \odot M_\text{sph})  ,
\end{equation}
where $I_d$ is the first render's depth and $M_\text{sph}$ is a binary mask of the sphere's footprint in image space.
Finally, the 3D radius is computed according to the sampled depth and camera's field of view to allow fixed image space radius.

The sphere is rendered with two materials: perfect mirror (\textit{roughness} = 0, \textit{metallic} = 1, \textit{albedo} = (1,1,1) and perfect diffuse (\textit{roughness} = 1, \textit{metallic} = 0, \textit{albedo} = (1,1,1)).
Images are rendered as float16 EXR, retaining the high dynamic range.
We also save the depth and sphere masks layers, along with the intrinsic camera matrix and the sphere's 3D position and world radius. 
These are used to compute the condition maps.
For video sequence data generation, the approach is adapted to 3 scenarios (see \cref{fig:data_example}).

\myparagraph{Dynamic sphere position}
In this scenario the camera is static and the spheres are moving.
A random image space offset $(x',y')$ is selected $x' \sim \mathcal{U}(0,W), \, y' \sim \mathcal{U}(0,H)$, where $W$ and $H$ are the image width and height respectively. The spheres will move by this offset in image space over the sequence.
For depth, an additional factor is sampled for the last frame and the 3D position and world size of the sphere is interpolated through the sequence.

\myparagraph{Dynamic camera}
We procedurally generate two camera poses and interpolate between them. The absolute distance to the sphere is interpolated between the first and last frame with a fixed depth factor to insure smooth movement.

\myparagraph{Dynamic lighting}
The lighting is made dynamic by randomly rotating the azimuth of the HDRI map, randomly changing the scene's light source intensities, and randomly rotating the ``Sun'' light.

\subsection{Equirectangular HDRI map optimization}
\label{sec:hdri}

At inference, the diffusion model is queried multiple times to generate both mirrored and diffuse spheres at different exposures.
We obtain a final HDRI by merging the predictions. 

Let us define a function $\mathcal{R}(L_t,m)$, which renders a sphere of material $m \in \mathcal{M}$, with $\mathcal{M} = \{\text{mirror}, \text{diffuse}\}$, under HDRI lighting $L_t$ at frame $t$. We then seek to find 
\begin{equation}
\argmin_L \sum_{t \in \mathcal{T}}\sum_{e \in \mathcal{E}} \sum_{m \in \mathcal{M}} \ell \left(\pi(e, m, t) - e\mathcal{R}(L_t, m) \right)\,,
\label{eqn:hdrfusion}
\end{equation}
where $\pi(e,m,t)$ is the fine-tuned generative model conditioned on frame $t$, exposure $e \in \mathcal{E}$, where $\mathcal{E} = 2^{\{\text{EV}_0, \text{EV}_{-3}, ...\}}$ and material $m$, and $\ell$ is the loss function 
\begin{equation}
\ell = \ell_2(\hat{y}_t, y_t) + \frac{\lambda}{2} (\ell_1(\hat{y}_t, \hat{y}_{t-1}) + \ell_1(\hat{y}_t, \hat{y}_{t+1})) \,.
\end{equation}
Here, $L$ is a spatio-temporal HDRI volume initialized at a constant gray value (0.5), and Adam is used to optimize \cref{eqn:hdrfusion} through gradient descent. $\mathcal{R}$ is implemented in PyTorch as a two-modes renderer: reflective and diffuse with cosine and light multi-importance sampling. In practice, we randomly alternate between exposures and materials at each iteration instead of summing over all possibilities. To speedup convergence, we represent $L$ with a Laplacian pyramid~\cite{gomez2024rrm}. 






\begin{table*}
\centering
\footnotesize
\setlength{\tabcolsep}{1pt}
\begin{tabular}{llcccccccccccccccc}
\midrule
& & \multicolumn{4}{c}{RMSE$_\downarrow$} &  \multicolumn{4}{c}{SI-RMSE$_\downarrow$} &
\multicolumn{4}{c}{SSIM$_\uparrow$} &
\multicolumn{4}{c}{Ang. Err.$_\downarrow$} \\
\midrule
 Dataset & Method      & Mirr & Diff & Gloss & Mat  & Mirr & Diff & Gloss & Mat  & Mirr & Diff & Gloss & Mat  & Mirr & Diff & Gloss & Matte \\
 \midrule
 \multirow{4}{*}{Infinigen} 
 & Diff.Light & \num{0.40} & \num{0.47} & \num{0.48} & \num{0.43} & \num{1.52} & \num{0.70} & \num{0.70} &\num{0.85} & \num{0.68}  & \num{0.83} & \num{0.81} & \num{0.83} & \third{\num{14.3}} & \third{\num{9.7}} & \third{\num{9.4}} & \third{\num{9.8}}  \\
 & 4D Lighting    & \third{\num{0.34}} & \third{\num{0.36}} & \third{\num{0.38}} & \third{\num{0.38}} & \third{\num{1.36}} & \third{\num{0.62}} & \third{\num{0.64}} & \third{\num{0.74}} & \third{\num{0.72}}  & \third{\num{0.86}} & \third{\num{0.85}} & \third{\num{0.84}} & \num{14.7} & \num{11.2} & \num{11.5} & \num{11.6}  \\
 & \themethod (image) & \best{\num{0.25}} & \best{\num{0.16}} & \best{\num{0.16}} & \best{\num{0.17}} & \best{\num{0.41}} & \best{\num{0.11}} & \best{\num{0.13}} & \best{\num{0.18}} & \second{\num{0.78}}  & \best{\num{0.95}} & \best{\num{0.94}} & \best{\num{0.95}} & \best{\num{4.4}} & \best{\num{2.3}} & \best{\num{2.2}} & \best{\num{2.4}}  \\
 & \themethod (video)  & \second{\num{0.26}} & \second{\num{0.22}} & \second{\num{0.22}} & \second{\num{0.21}} & \second{\num{0.42}} & \second{\num{0.13}} & \second{\num{0.16}} & \second{\num{0.21}} & \best{\num{0.79}}  & \best{\num{0.95}} & \second{\num{0.93}} & \second{\num{0.94}} & \best{\num{4.4}} & \second{\num{2.7}} & \second{\num{2.8}} & \second{\num{2.9}}  \\
 \midrule
 \multirow{4}{*}{\shortstack[l]{Laval\\Indoor SV}} 
 & Diff.Light & \num{0.50} & \num{0.49} & \num{0.51} & \num{0.49} & \num{1.40} & \num{0.91} & \num{0.89} &\num{0.99} & \num{0.70} & \num{0.80} & \num{0.78} & \num{0.80} & \num{10.3} & \num{8.2} & \num{8.2} & \num{7.8} \\
 & 4D Lighting     & \second{\num{0.35}} & \second{\num{0.27}} & \second{\num{0.28}} & \third{\num{0.31}} & \third{\num{0.91}} & \third{\num{0.22}} & \second{\num{0.23}} & \third{\num{0.39}} & \second{\num{0.80}} & \second{\num{0.94}} & \second{\num{0.93}} & \third{\num{0.92}} & \third{\num{6.8}} & \third{\num{5.0}} & \third{\num{5.0}} & \third{\num{5.1}} \\
 & \themethod (image) & \best{\num{0.30}} & \best{\num{0.20}} & \best{\num{0.21}} & \best{\num{0.24}} & \best{\num{0.60}} & \best{\num{0.17}} & \best{\num{0.18}} & \best{\num{0.28}} & \best{\num{0.81}} & \best{\num{0.97}} & \best{\num{0.95}} & \best{\num{0.95}} & \best{\num{4.6}} & \best{\num{2.6}} & \best{\num{2.6}} & \best{\num{2.9}} \\
 & \themethod (video)  &  \second{\num{0.35}} & \second{\num{0.27}} & \second{\num{0.28}} & \second{\num{0.30}} & \second{\num{0.66}} & \second{\num{0.21}} & \second{\num{0.23}} & \second{\num{0.34}} & \second{\num{0.80}} & \second{\num{0.94}} & \second{\num{0.93}} & \second{\num{0.93}} & \second{\num{5.5}} & \second{\num{3.7}} & \second{\num{3.6}} & \second{\num{3.9}} \\
 \midrule
 \multirow{4}{*}{\shortstack[l]{Laval\\Outdoor}} 
 & Diff.Light & \third{\num{0.36}} & \third{\num{0.23}} & \third{\num{0.24}} & \third{\num{0.24}} & \num{0.89} & \num{0.13} & \third{\num{0.11}} & \third{\num{0.18}} & \num{0.75} & \second{\num{0.98}} & \third{\num{0.97}} & \third{\num{0.97}} & \third{\num{7.3}} & \third{\num{2.3}} & \third{\num{2.1}} & \third{\num{2.9}} \\
 & 4D Lighting     & \num{0.37} & \num{0.25} & \num{0.28} & \num{0.29} & \third{\num{0.69}} & \third{\num{0.12}} & \num{0.13} & \num{0.21} & \third{\num{0.77}} & \num{0.97} & \num{0.96} & \num{0.95} & \num{7.4} & \num{2.9} & \num{3.0} & \num{3.4} \\
 & \themethod (image) & \second{\num{0.27}} & \best{\num{0.12}} & \best{\num{0.13}} & \best{\num{0.13}} & \second{\num{0.37}} & \best{\num{0.07}} & \best{\num{0.07}} & \best{\num{0.11}} & \second{\num{0.79}} & \best{\num{0.99}} & \best{\num{0.99}} & \best{\num{0.99}} & \best{\num{3.9}} & \best{\num{1.2}} & \best{\num{1.1}} & \best{\num{1.5}} \\
 & \themethod (video)  & \best{\num{0.26}} & \second{\num{0.19}} & \second{\num{0.20}} & \second{\num{0.19}} & \best{\num{0.34}} & \second{\num{0.09}} & \second{\num{0.09}} & \second{\num{0.13}} & \best{\num{0.82}} & \second{\num{0.98}} & \second{\num{0.98}} & \second{\num{0.98}} & \second{\num{4.0}} & \second{\num{1.9}} & \second{\num{1.9}} & \second{\num{2.4}} \\
\end{tabular}
\caption{Quantitative evaluation of lighting estimation on single images from the Infinigen~\cite{infinigen2024indoors} (top), the Laval Indoor SV dataset~\cite{garon2019fast} (middle) \change{and the Laval Outdoor HDR Dataset~\cite{holdgeoffroy-cvpr-19} (bottom)}. We compare the image and video versions of \themethod with ``Diff.Light''~\cite{Phongthawee2023DiffusionLight} and ``4D Lighting''~\cite{4DLighting}. ``Mirr'' (mirror), ``Diff'' (diffuse), ``Gloss'' (glossy) and ``Mat'' (matte) refer to the different test spheres (see \cref{sec:metrics_datasets}). Results are color coded by \best{best}, \second{second} and \third{third} best. We observe that \themethod (image) outperforms the previous work in all cases.}
\label{tab:quant_single_img}
\end{table*}

\begin{figure*}[!th]
   \centering
   \footnotesize
   \setlength{\tabcolsep}{0.5pt}
   \newlength{\tmplength}
   \newlength{\cbarheight}
   \setlength{\tmplength}{0.11\linewidth}
   \setlength{\cbarheight}{1.9cm}
    \begin{tabular}{ccccccc}
    Scene & & DiffusionLight & 4D Lighting & \themethod (image) & \themethod (video) & GT \\
    
    \multirow[t]{2}{*}{\raisebox{-0.5\height}{\includegraphics[width=2\tmplength]{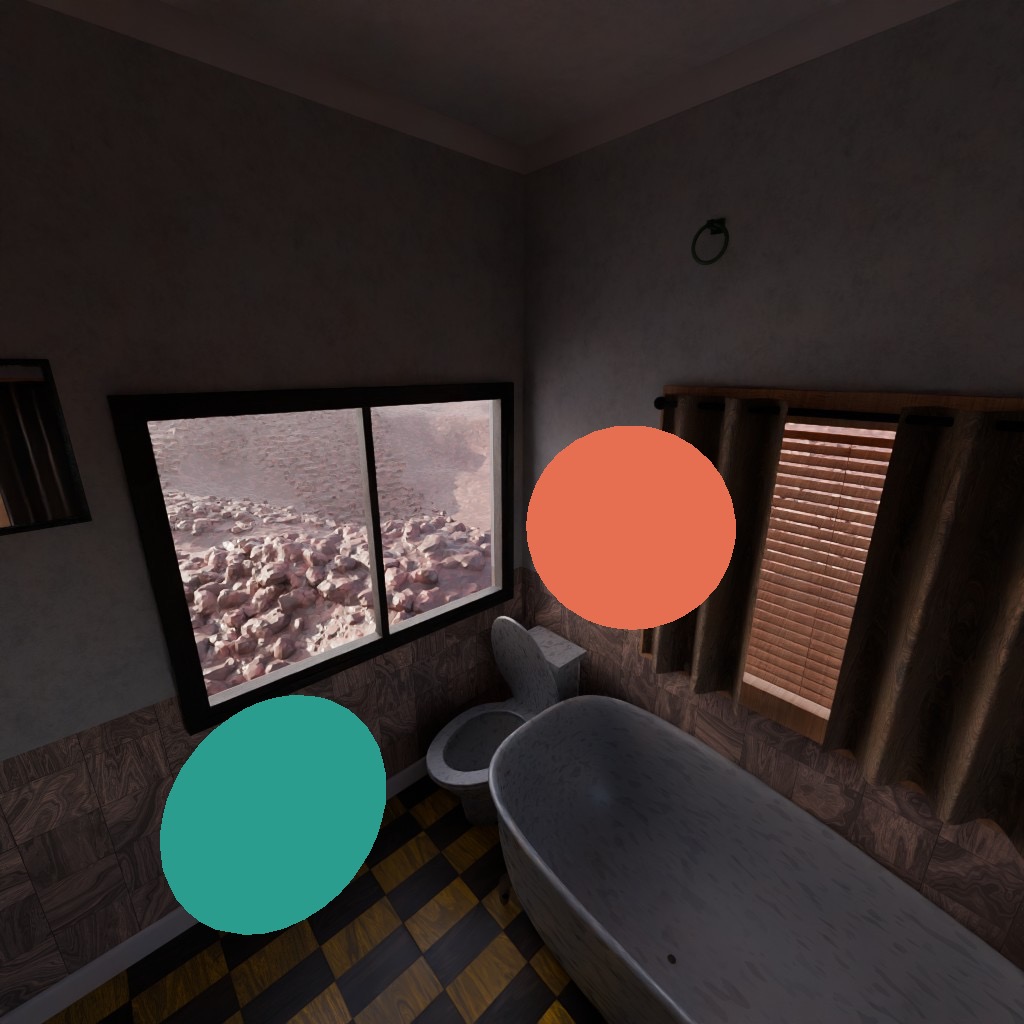}}}&
         \color{color2}\rule[4pt]{1.2pt}{1.7cm} 
    &
    \includegraphics[width=\tmplength]{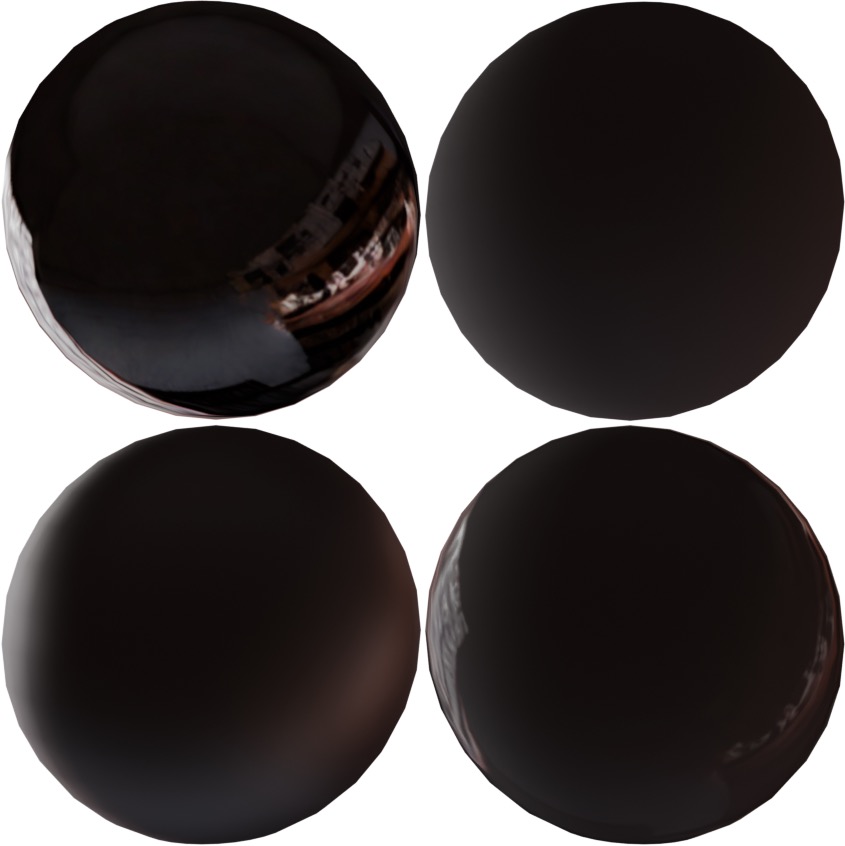}&
    \includegraphics[width=\tmplength]{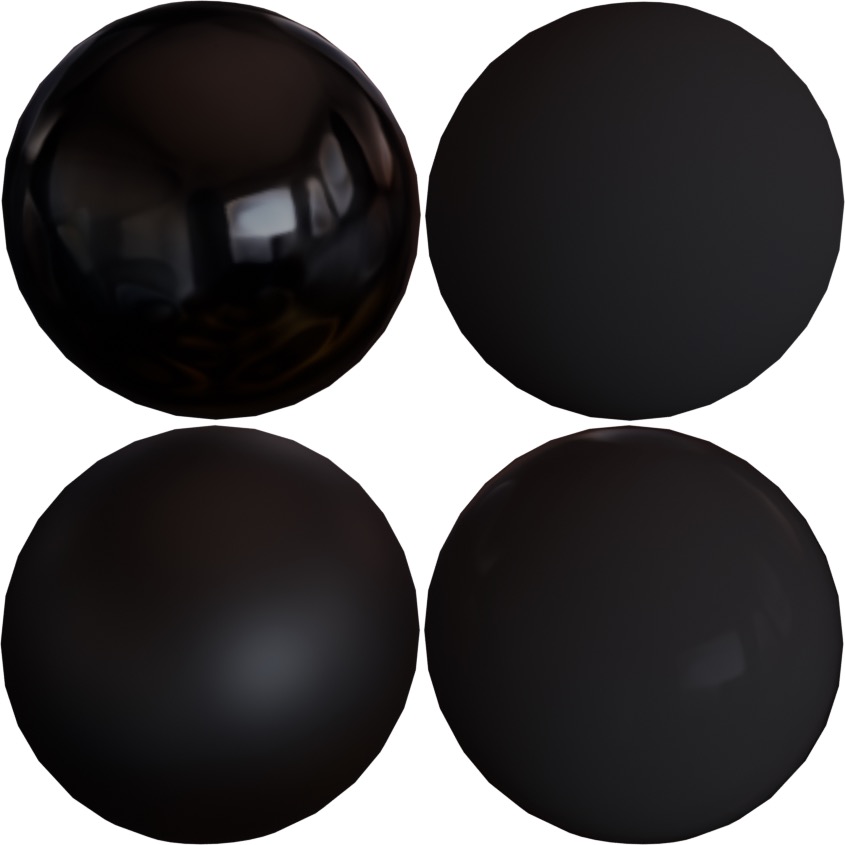}&
    \includegraphics[width=\tmplength]{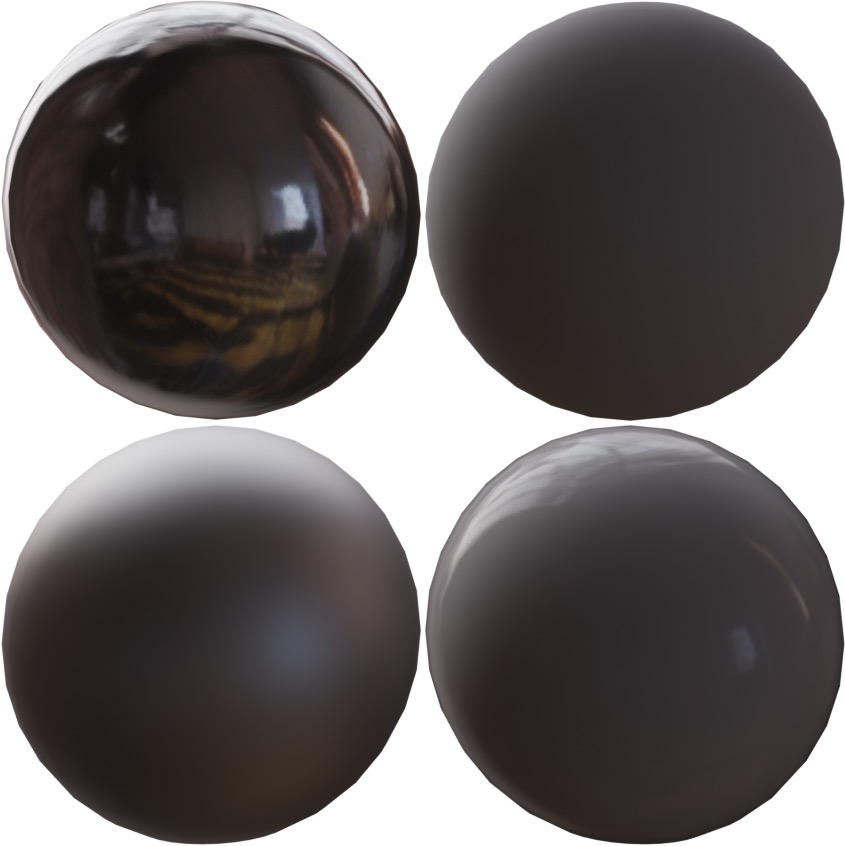}&
    \includegraphics[width=\tmplength]{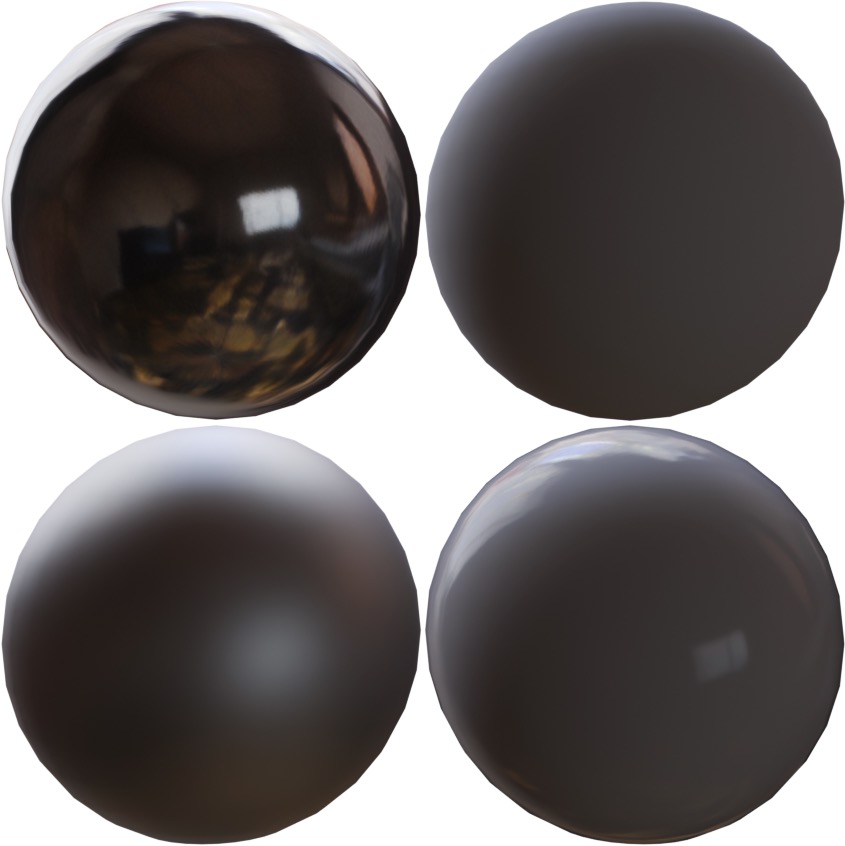}&
    \includegraphics[width=\tmplength]{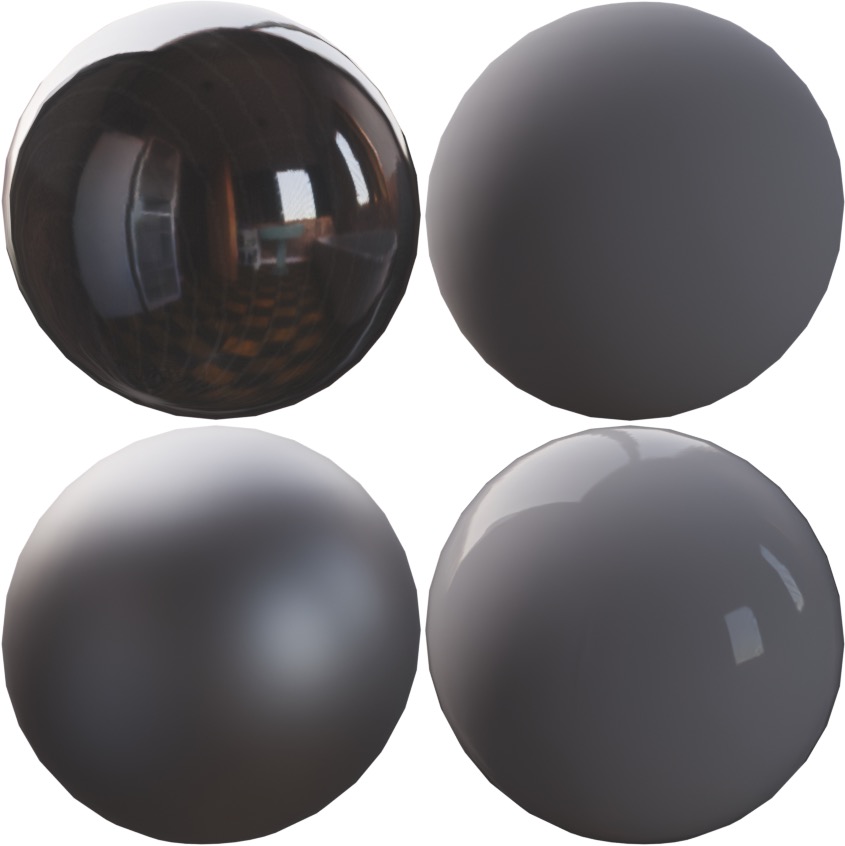}\\
    &
         \color{color5}\rule[4pt]{1.2pt}{1.7cm}
    &
    \includegraphics[width=\tmplength]{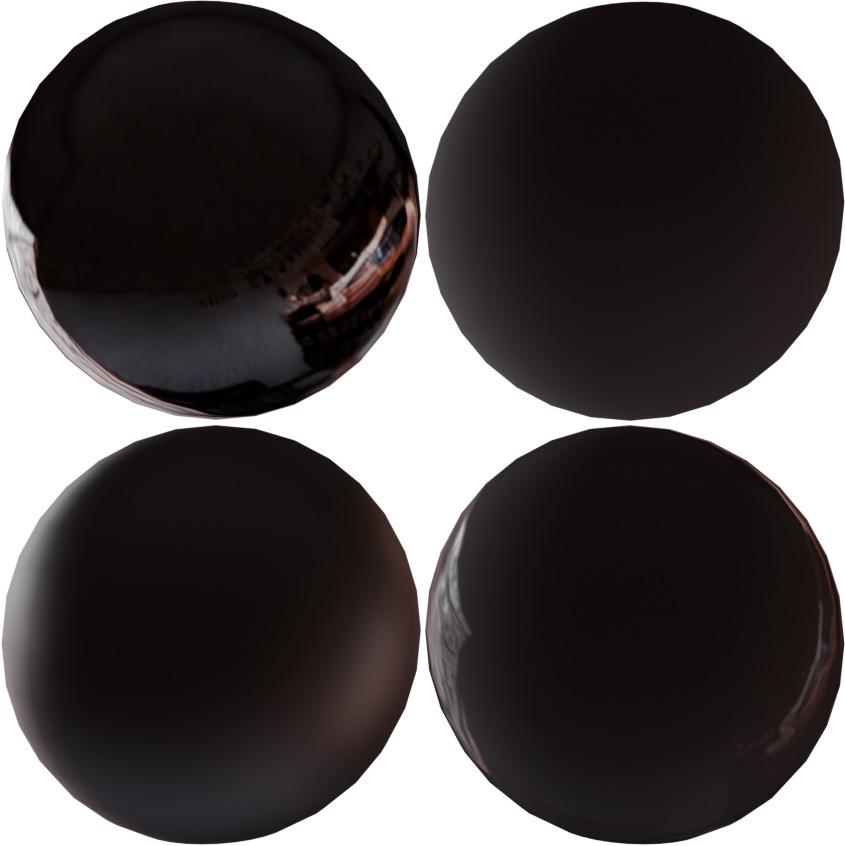}&
    \includegraphics[width=\tmplength]{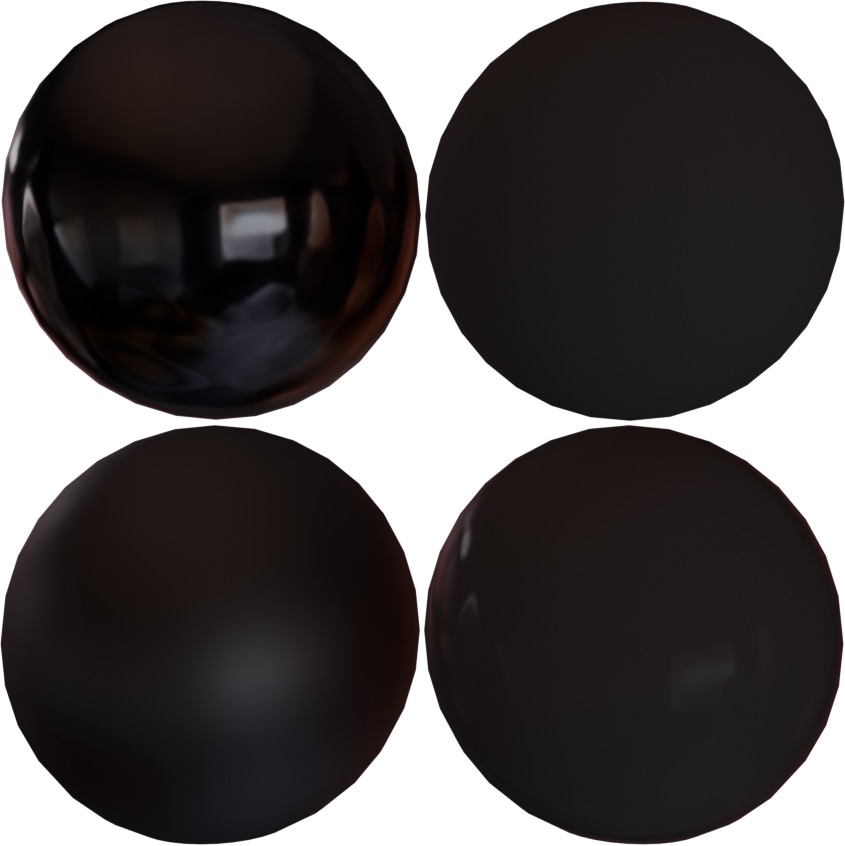}&
    \includegraphics[width=\tmplength]{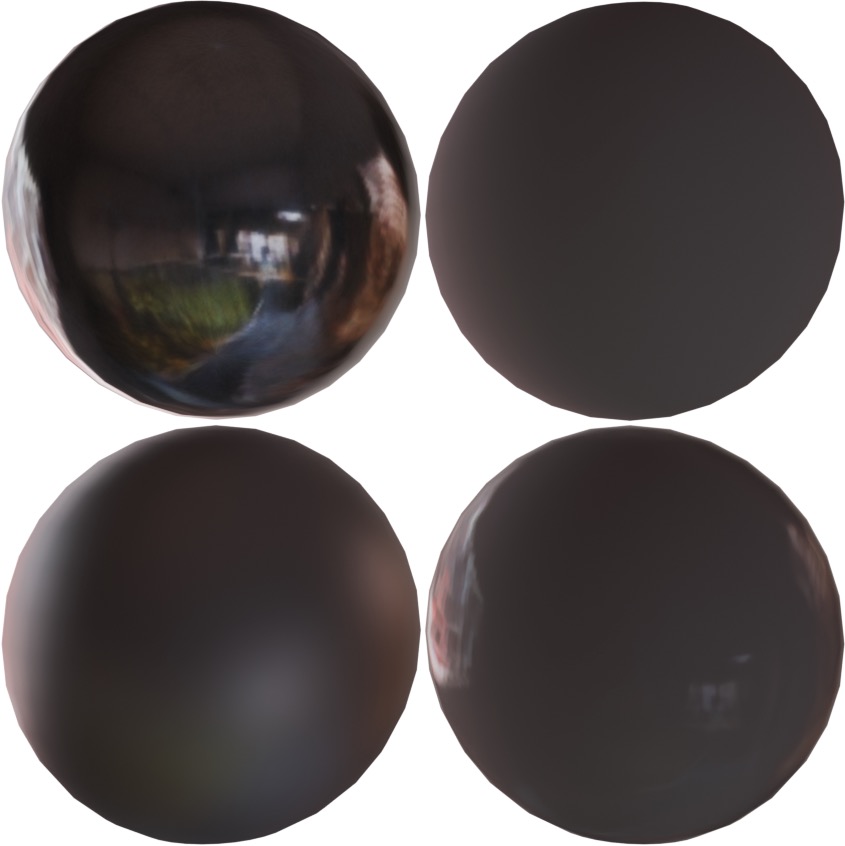}&
    \includegraphics[width=\tmplength]{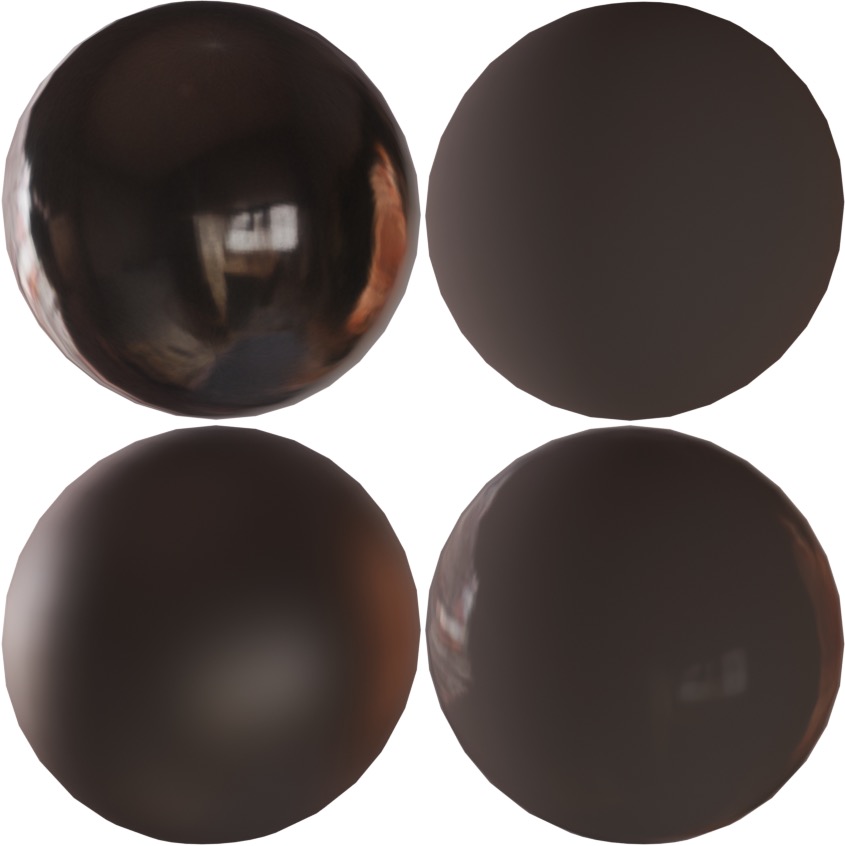}&
    \includegraphics[width=\tmplength]{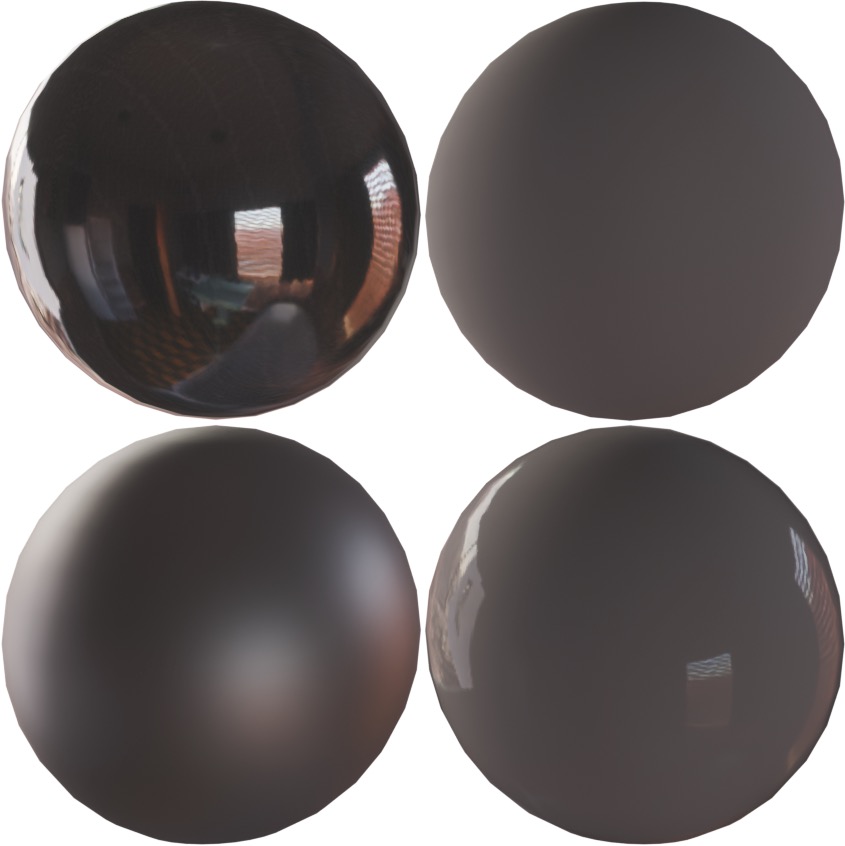}\\
    \multirow[t]{2}{*}{\raisebox{-0.5\height}{\includegraphics[width=2\tmplength]{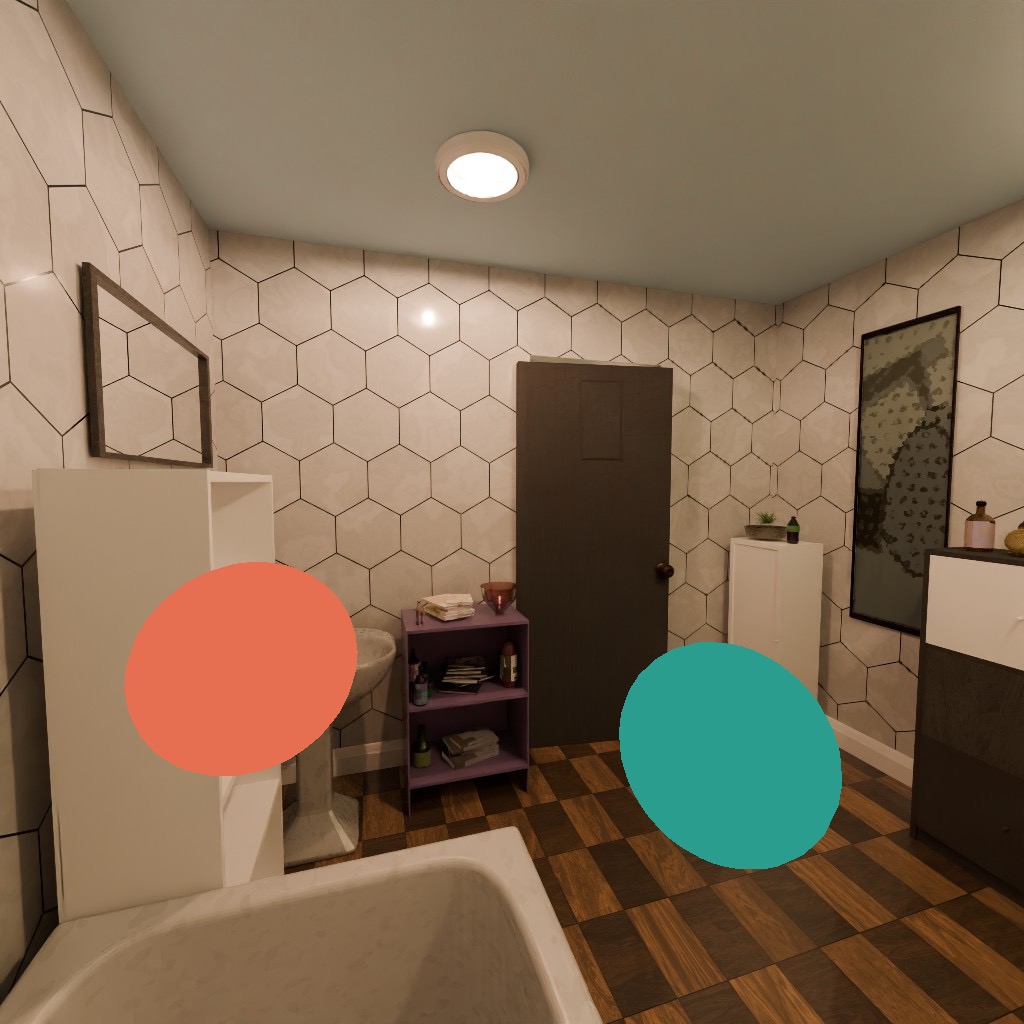}}}&
         \color{color2}\rule[4pt]{1.2pt}{1.7cm} 
    &
    \includegraphics[width=\tmplength]{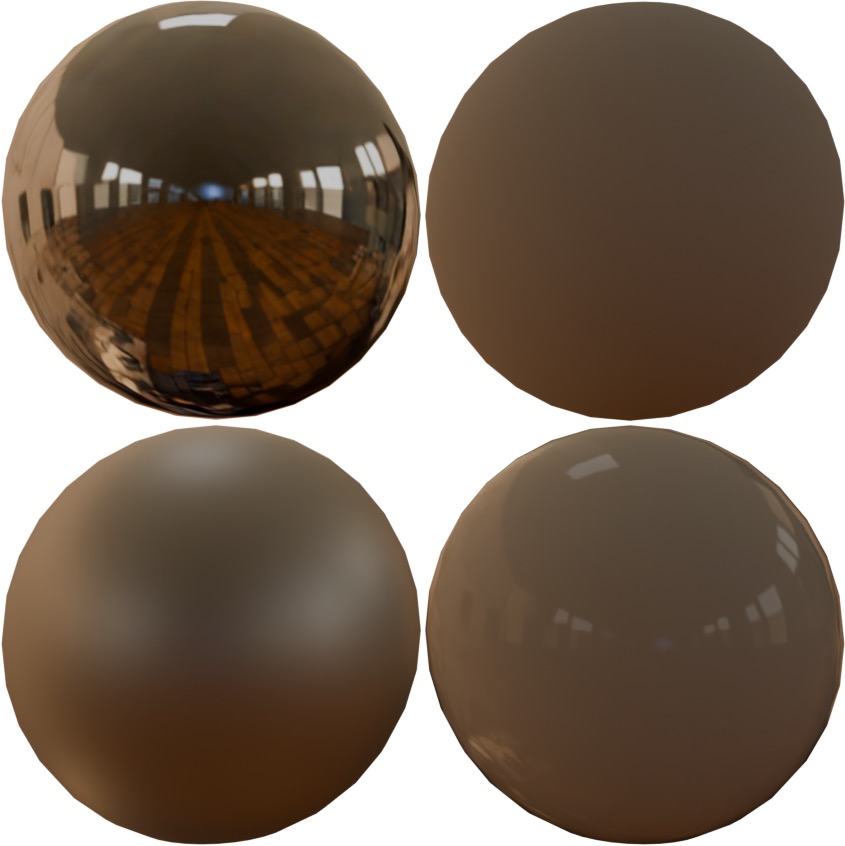}&
    \includegraphics[width=\tmplength]{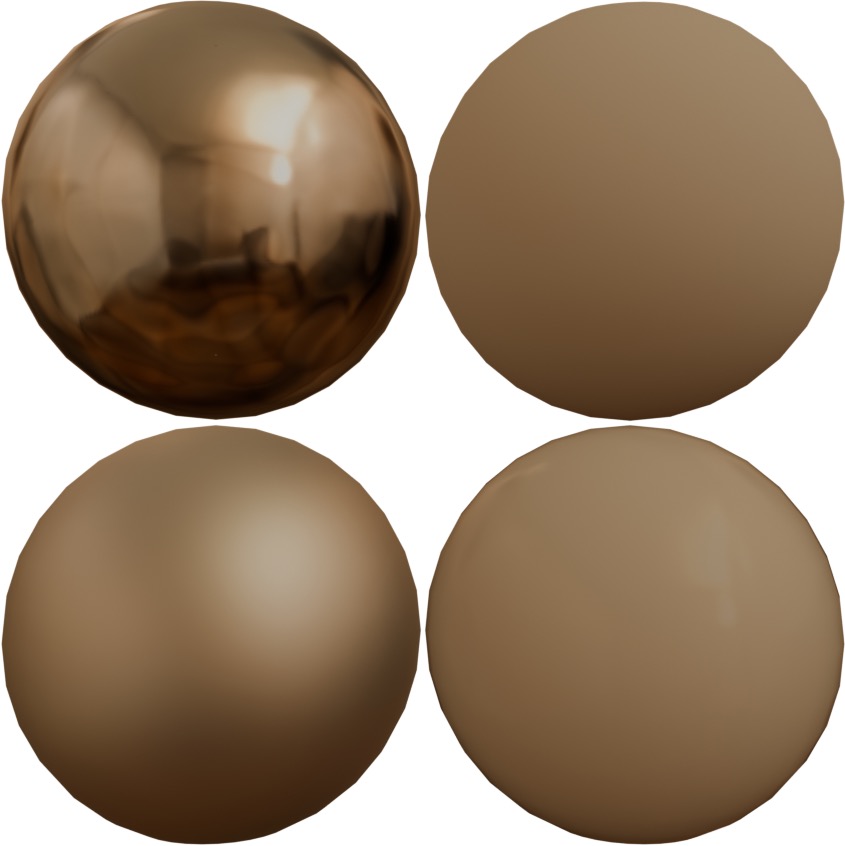}&
    \includegraphics[width=\tmplength]{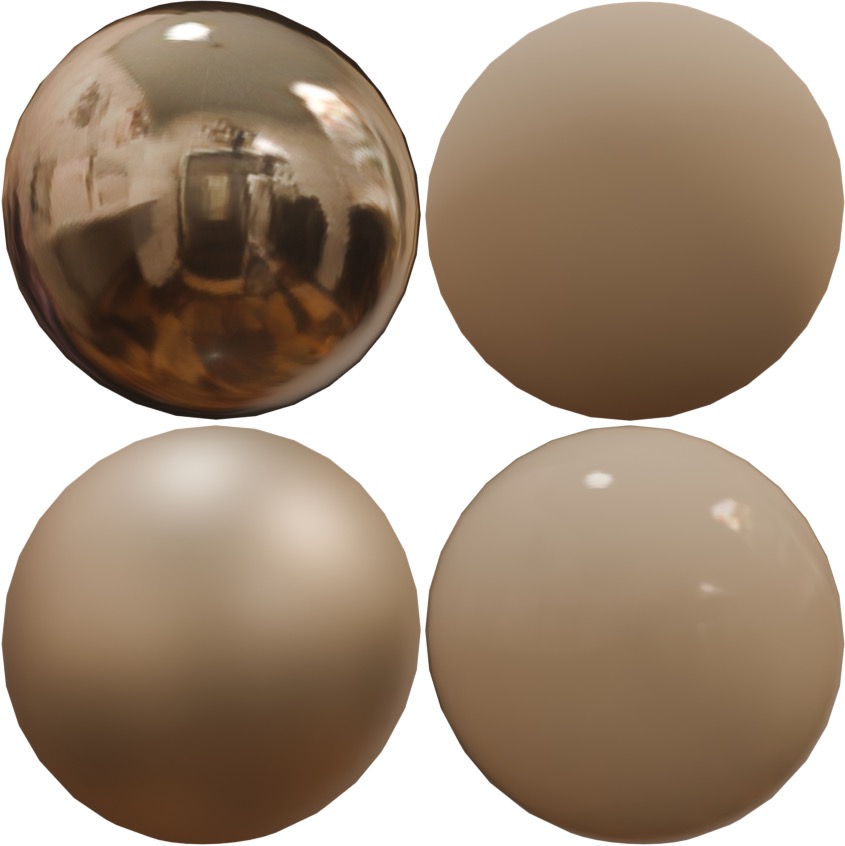}&
    \includegraphics[width=\tmplength]{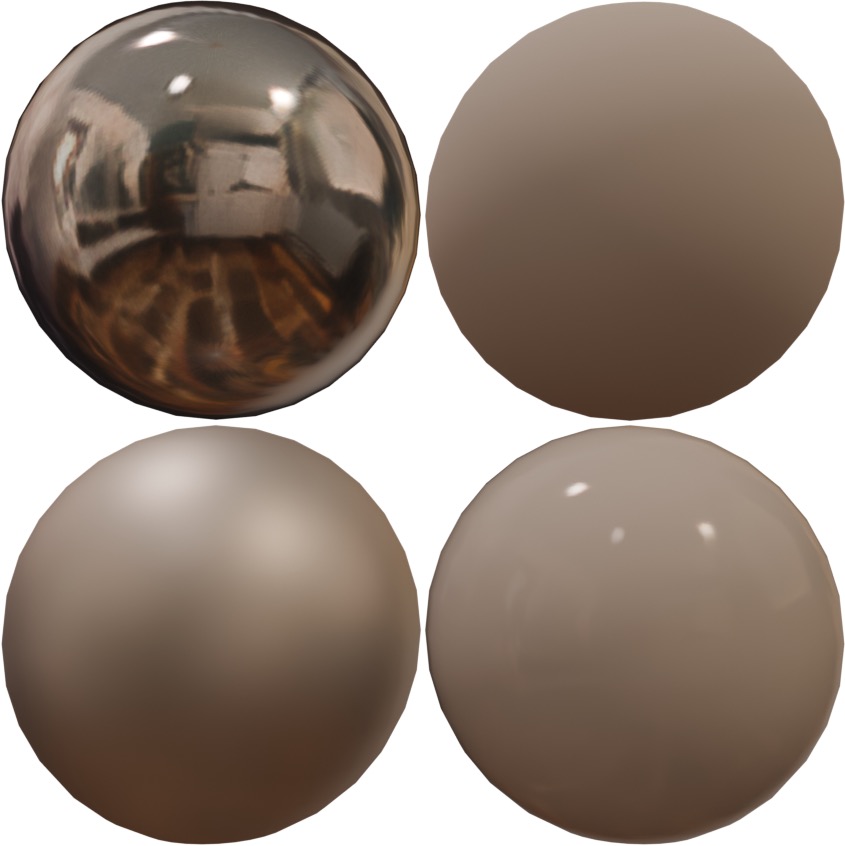}&
    \includegraphics[width=\tmplength]{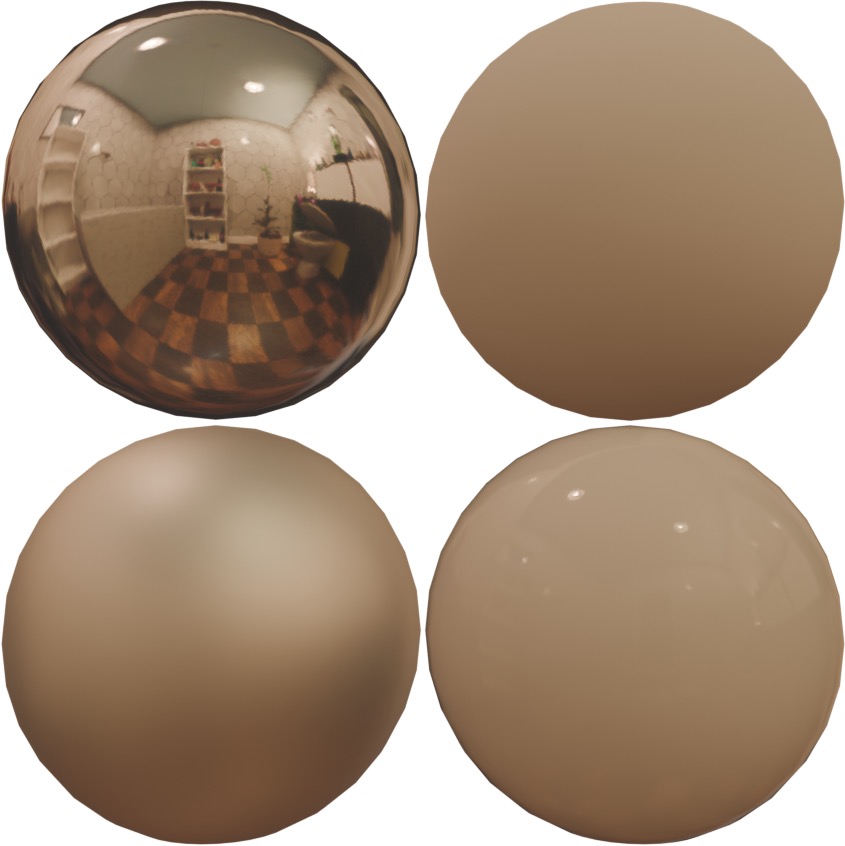}\\
    &
         \color{color5}\rule[4pt]{1.2pt}{1.7cm}
    &
    \includegraphics[width=\tmplength]{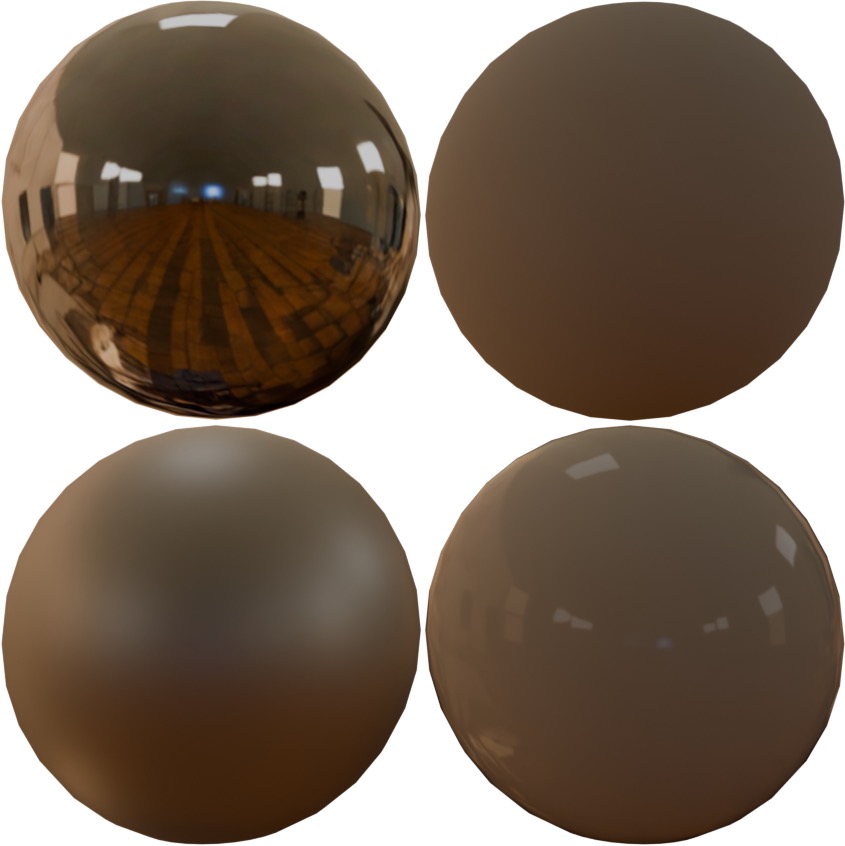}&
    \includegraphics[width=\tmplength]{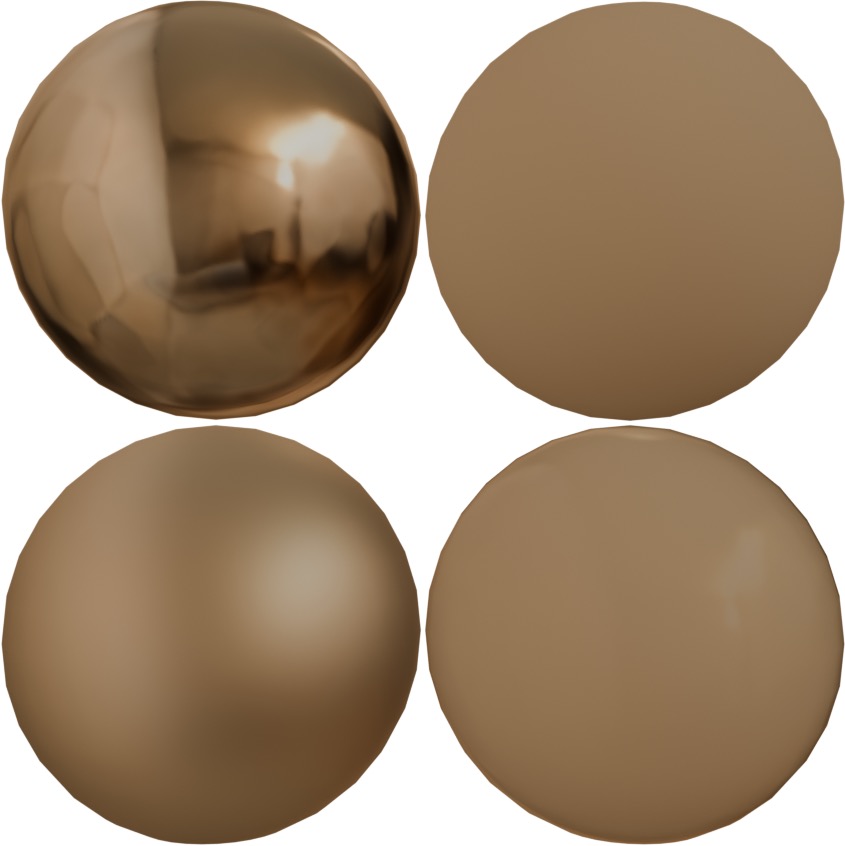}&
    \includegraphics[width=\tmplength]{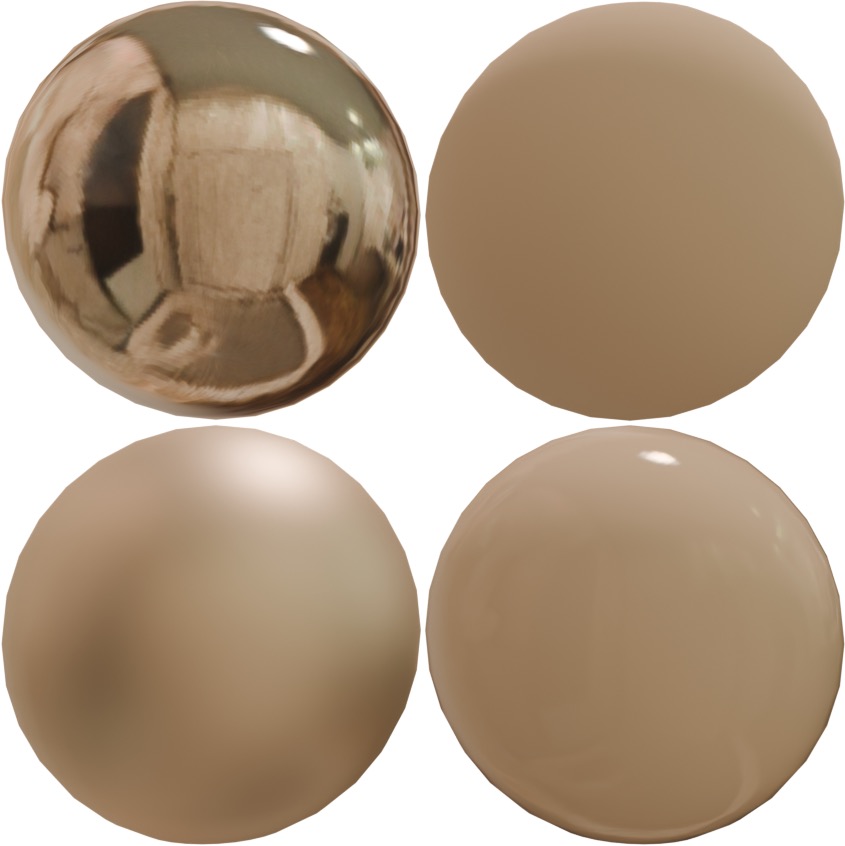}&
    \includegraphics[width=\tmplength]{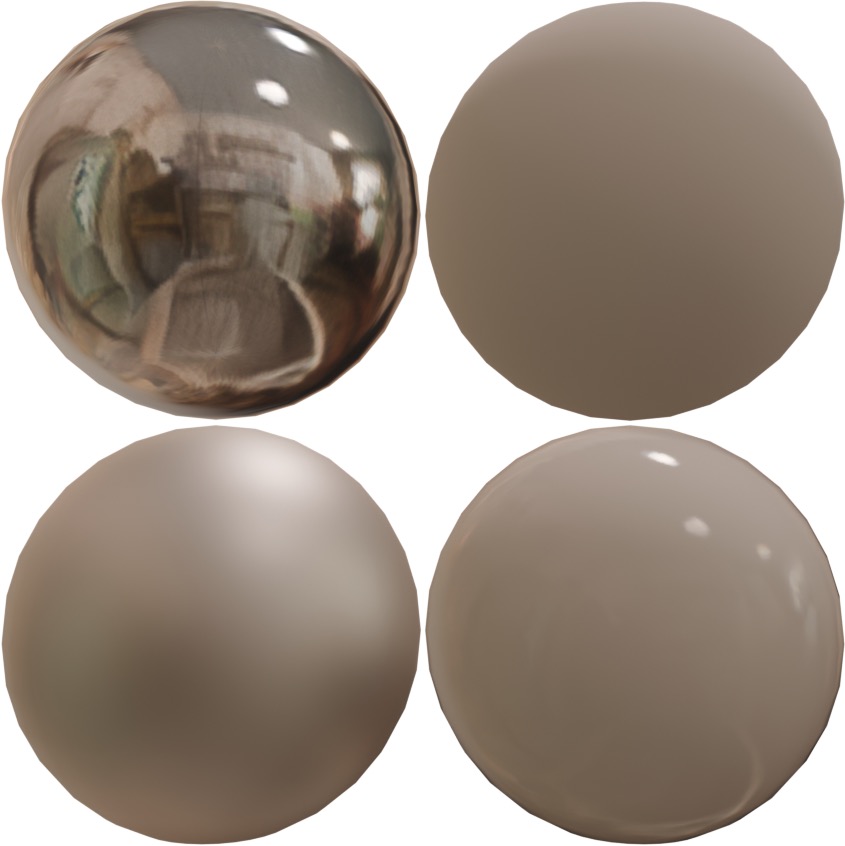}&
    \includegraphics[width=\tmplength]{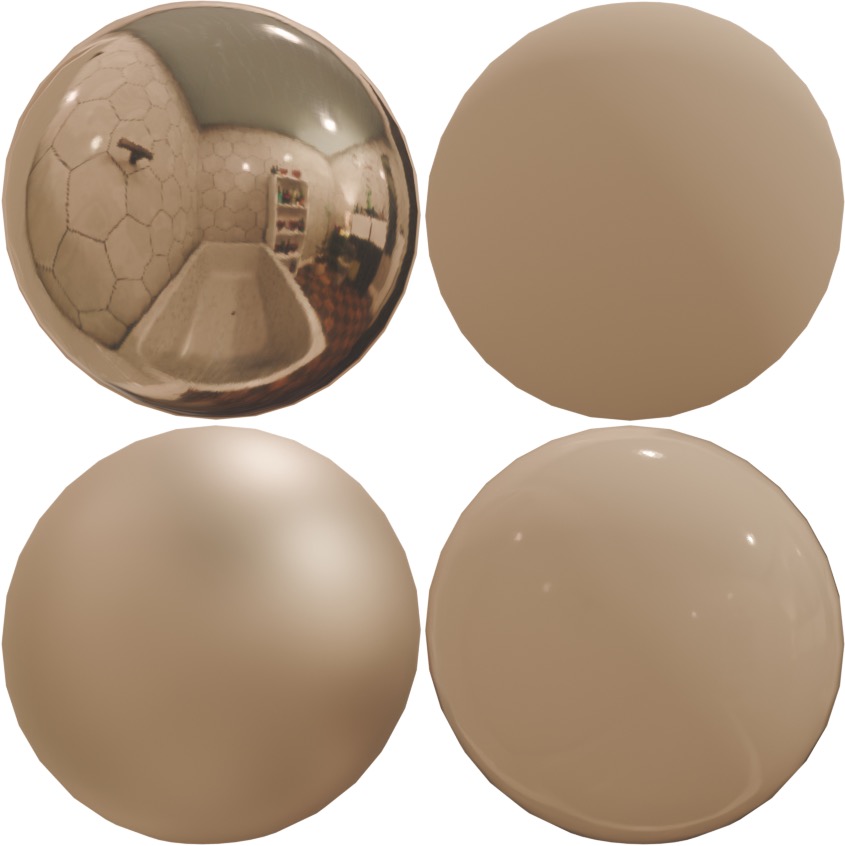}\\
    \end{tabular}
    \caption{Sample predictions from the Infinigen test set~\cite{infinigen2024indoors} for, from left to right: DiffusionLight~\cite{Phongthawee2023DiffusionLight}, 4D Lighting~\cite{4DLighting}, and the image and video versions of the proposed \themethod. We visualize predictions by rendering the same four test spheres used for the quantitative metrics (see \cref{tab:quant_single_img}): mirror (top left), diffuse (top right), matte (bottom left) and glossy (bottom right).} 
    \label{fig:single_im_qualitative}
\end{figure*}

\section{Experiments}
\label{sec:experiments}

\subsection{Implementation details}

We train two models: an image diffusion model and a video diffusion model. We fine-tune the full image Flux.1 Schnell model~\cite{flux2024} for 150k steps using 12896 images at $512 \times 512$ resolution. For the video model, we use Wan2.2 5B~\cite{wan2025} model and fine-tune for 250k steps using 30,096 sequences of 21 frames also at $512 \times 512$ resolution. For both versions, a series of color, exposure and degradation augmentations is employed. The training takes 50 hours for the image model and 188 hours for the video model on 8 A100E GPUs.

\begin{table*}
\centering
\footnotesize
\setlength{\tabcolsep}{1pt}
\begin{tabular}{llcccccccccccccc}
\midrule
& & \multicolumn{2}{c}{RMSE$_\downarrow$}  &  \multicolumn{2}{c}{SI-RMSE$_\downarrow$}  & \multicolumn{2}{c}{SSIM$_\uparrow$}  &   \multicolumn{2}{c}{Ang. Err.$_\downarrow$}  &  \multicolumn{2}{c}{T-LPIPS$_\downarrow$}  &  \multicolumn{2}{c}{T-LPIPS-Diff$_\downarrow$}  &  \multicolumn{2}{c}{Warped Err$_\downarrow$} \\
\midrule
 Dataset & Method   & Mirr & Diff   &   Mirr & Diff   & Mirr & Diff  & Mirr & Diff  & Mirr & Diff  & Mirr & Diff  & Mirr & Diff  \\
 \midrule
 \multirow{4}{*}{\shortstack[l]{Dynamic\\object}} 
 & 4D Lighting   & \num{0.39} & \num{0.29}  & \num{1.18}   & \num{0.20} & \num{0.70}   & \num{0.90}  & \num{7.1}  & \num{4.7} & \best{\num{0.0048}}  & \best{\num{0.0004}} &  \second{\num{0.0418}}  & \best{\num{0.0009}} &  \best{\num{0.0439}}  & \best{\num{0.0079}} \\
 & \themethod (image) & \best{\num{0.28}}  & \best{\num{0.15}} & \best{\num{0.43}}   & \best{\num{0.10}}  & \second{\num{0.77}}   & \best{\num{0.97}}  & \best{\num{3.0}}  & \best{\num{1.4}} & \num{0.1340}  & \num{0.0054}  &  \num{0.0886}  & \num{0.0045} &  \num{0.1887}  & \num{0.0330} \\
 & \themethod (video)  & \second{\num{0.30}}  & \second{\num{0.18}} & \second{\num{0.45}}   & \second{\num{0.12}} & \best{\num{0.78}}   & \best{\num{0.97}}  & \second{\num{4.5}}  & \second{\num{3.0}} & \second{\num{0.0242}}  & \second{\num{0.0014}}  &  \best{\num{0.0227}}  & \second{\num{0.0013}} &  \second{\num{0.0589}}  & \second{\num{0.0134}} \\
 \midrule
  \multirow{4}{*}{\shortstack[l]{Dynamic\\camera}} 
 & 4D Lighting   & \num{0.39}  & \num{0.37}  & \num{0.88}   & \num{0.17} & \num{0.71}   & \num{0.90}  & \num{6.5}  & \num{3.7}  & \best{\num{0.0057}}  & \best{\num{0.0007}} &  \second{\num{0.0279}}  & \best{\num{0.0007}}  &  \best{\num{0.0354}}  & \second{\num{0.0124}}\\
 & \themethod (image) & \best{\num{0.30}}  & \best{\num{0.16}} & \best{\num{0.44}}   & \best{\num{0.10}}  & \second{\num{0.76}}   & \best{\num{0.97}}  & \best{\num{3.3}}  & \best{\num{1.4}}  & \num{0.1051}  & \num{0.0057} &  \num{0.0715}  & \num{0.0053} &  \num{0.1500}  & \num{0.0394} \\
 & \themethod (video)  & \best{\num{0.30}}  & \second{\num{0.23}} & \second{\num{0.47}}   & \second{\num{0.12}} & \best{\num{0.77}}   & \best{\num{0.97}}  & \second{\num{4.4}}  & \second{\num{2.5}} & \second{\num{0.0220}}  & \second{\num{0.0010}} &  \best{\num{0.0148}}  & \second{\num{0.0011}}  &  \second{\num{0.0506}}  & \best{\num{0.0122}} \\
 \midrule
  \multirow{4}{*}{\shortstack[l]{Dynamic\\lighting}}
 & 4D Lighting  & \num{0.39}  & \num{0.33} & \num{1.37}   & \num{0.69}  & \num{0.68}   & \num{0.88}  & \num{12.6}  & \num{10.0} & \second{\num{0.0030}}  & \best{\num{0.0006}}  &  \second{\num{0.0067}}  & \best{\num{0.0005}}  &  \second{\num{0.0158}}  & \second{\num{0.0093}} \\
 & \themethod (image) & \best{\num{0.28}}  & \best{\num{0.16}} & \best{\num{0.44}}   & \best{\num{0.11}}  & \best{\num{0.78}}   & \best{\num{0.97}}  & \best{\num{3.6}}  & \best{\num{1.8}} & \num{0.0162}  & \num{0.0018}  &  \num{0.0085}  & \num{0.0012} &  \num{0.0336}  & \num{0.0193} \\
 & \themethod (video)  & \second{\num{0.34}}  & \second{\num{0.22}} & \second{\num{0.49}}   & \second{\num{0.14}} & \second{\num{0.76}}   & \second{\num{0.96}}  & \second{\num{4.7}}  & \second{\num{2.9}} & \best{\num{0.0027}}  & \second{\num{0.0008}}  &  \best{\num{0.0065}}  & \best{\num{0.0005}} &  \best{\num{0.0108}}  & \best{\num{0.0076}} \\
 \midrule
  \multirow{4}{*}{Combination} 
 & 4D Lighting    & \num{0.38}  & \num{0.31}  & \num{0.98}   & \num{0.19} & \num{0.70}   & \num{0.91}  & \num{7.7}  & \num{3.9} & \best{\num{0.0071}}  & \best{\num{0.0010}}  &  \second{\num{0.0496}}  & \best{\num{0.0013}}  &  \best{\num{0.0558}}  & \best{\num{0.0150}}\\
 & \themethod (image) & \best{\num{0.29}}  & \best{\num{0.16}} & \best{\num{0.46}}   & \best{\num{0.11}}  & \best{\num{0.77}}   & \best{\num{0.97}}  & \best{\num{3.7}}  & \best{\num{1.9}} & \num{0.1354}  & \num{0.0075}  &  \num{0.0787}  & \num{0.0060}  &  \num{0.1941}  & \num{0.0463}\\
 & \themethod (video)  & \second{\num{0.33}}  & \second{\num{0.23}} & \second{\num{0.48}}   & \second{\num{0.15}} & \best{\num{0.77}}   & \second{\num{0.95}}  & \second{\num{4.6}}  & \second{\num{2.7}} & \second{\num{0.0370}}  & \second{\num{0.0031}}  &  \best{\num{0.0208}}  & \second{\num{0.0030}}  &  \second{\num{0.0780}}  & \second{\num{0.0250}} \\
\end{tabular}
\caption{Quantitative evaluation of lighting estimation on dynamic scenes. We compare \themethod with ``4D Lighting''~\cite{4DLighting}. ``Mirr'' (mirror) and ``Diff'' (diffuse) refer to the different test spheres (see \cref{sec:metrics_datasets}). Due to space limits, Glossy and Matte metrics are omitted and available in the supplementary materials. Results are color coded by \best{best}, \second{second} best.}
\label{tab:quant_temporal_img}
\end{table*}

\subsection{Evaluation metrics, datasets and baselines}
\label{sec:metrics_datasets}

As is typical for lighting estimation methods, we compute metrics on spheres relit with the predicted HDRI maps.
This allows comparing both the specular appearance and physical accuracy of the HDRI on different materials.
Concretely, we render spheres of different materials with the predicted HDR illumination: a perfect mirror, perfect diffuse, semi-rough metallic (herein called matte), and perfectly glossy. 
We evaluate our method for single image predictions on a synthetic and a real dataset.
The synthetic dataset consists of 28 scenes from Infinigen Indoor~\cite{infinigen2024indoors}, in which 4 light probes are randomly scattered in space.
The Laval Indoor Spatially Varying HDR dataset~\cite{garon2019fast} is a real dataset of physical probes placed in a scene and captured in HDR.
Because mirror light probes are not perfectly reflective, the authors of the original dataset graciously shared the reflectivity of the sphere used (74\%), which we used to adjust the HDR probes and treated as ground truth.
\change{To benchmark on outdoor scenes, we generate a non-spatially varying dataset consisting of 26 randomly rotated crop sequences from the Laval Outdoor HDR Dataset~\cite{holdgeoffroy-cvpr-19}, from which 3 frames are selected for metrics.}
For video predictions, we take 5 of the Blender demo files~\cite{BlenderDemoFiles} and augment them by animating the cameras, moving probes and modifying the light sources.
We compare our method against DiffusionLight~\cite{Phongthawee2023DiffusionLight} and 4D Lighting \cite{4DLighting} using their public implementations.

We report RMSE to inform the intensity of the predicted illuminance, SI-RMSE and SSIM as indicators of the structure of the predictions, and RGB angular error to assess the color reconstruction.
For temporal results, as is typical of video lighting tasks~\cite{mei2025lux}, we report T-LPIPS, LPIPS on neighboring frames.
However, to account for the motion of the ground truth scene, we also report T-LPIPS-Diff, the absolute difference between T-LPIPS of the prediction and the ground truth.
In addition, we report the warped error, computed by warping the current frame of the prediction by the optical flow predicted from an off-the-shelf module~\cite{RAFT2020} and comparing with the RMSE of the next frame.

\subsection{Single image results}

Quantitative results for single image predictions are provided in \cref{tab:quant_single_img}.
On both synthetic and real data, our image model performs better than the two baselines, with our video model trailing close behind.
We attribute the discrepancy between both versions to the capacity of the respective models (12B vs 5B parameters for image vs video) and to the optimization for both quality and time consistency of the video model.
The metrics from 4D Lighting for Infinigen differ from the numbers reported in \cite{4DLighting} since we sample a new set of scenes and spheres.
The metrics confirm that \themethod better predicts total illuminance (from the RMSE), accurate geometry (from SI-RMSE and SSIM), and better colors (from angular error).
DiffusionLight does not adapt to the 3D locations of the spheres, as reflected in the lower scores.
In real scenes, \themethod achieves better or equal (for video) results than 4D Lighting, showing strong generalization to real images.
\change{Our method also performs strongly on outdoor scenes as shown in metrics from the Laval Outdoor dataset.}

Visual samples from the predictions, shown in \cref{fig:single_im_qualitative}, visually demonstrate the higher quality reflections in comparison to 4D Lighting and the better HDR predictions, particularly when looking at the glossy highlights.

\begin{figure}
    \centering
    \footnotesize
    \setlength{\mywidth}{\linewidth}
    \includegraphics[width=\mywidth]{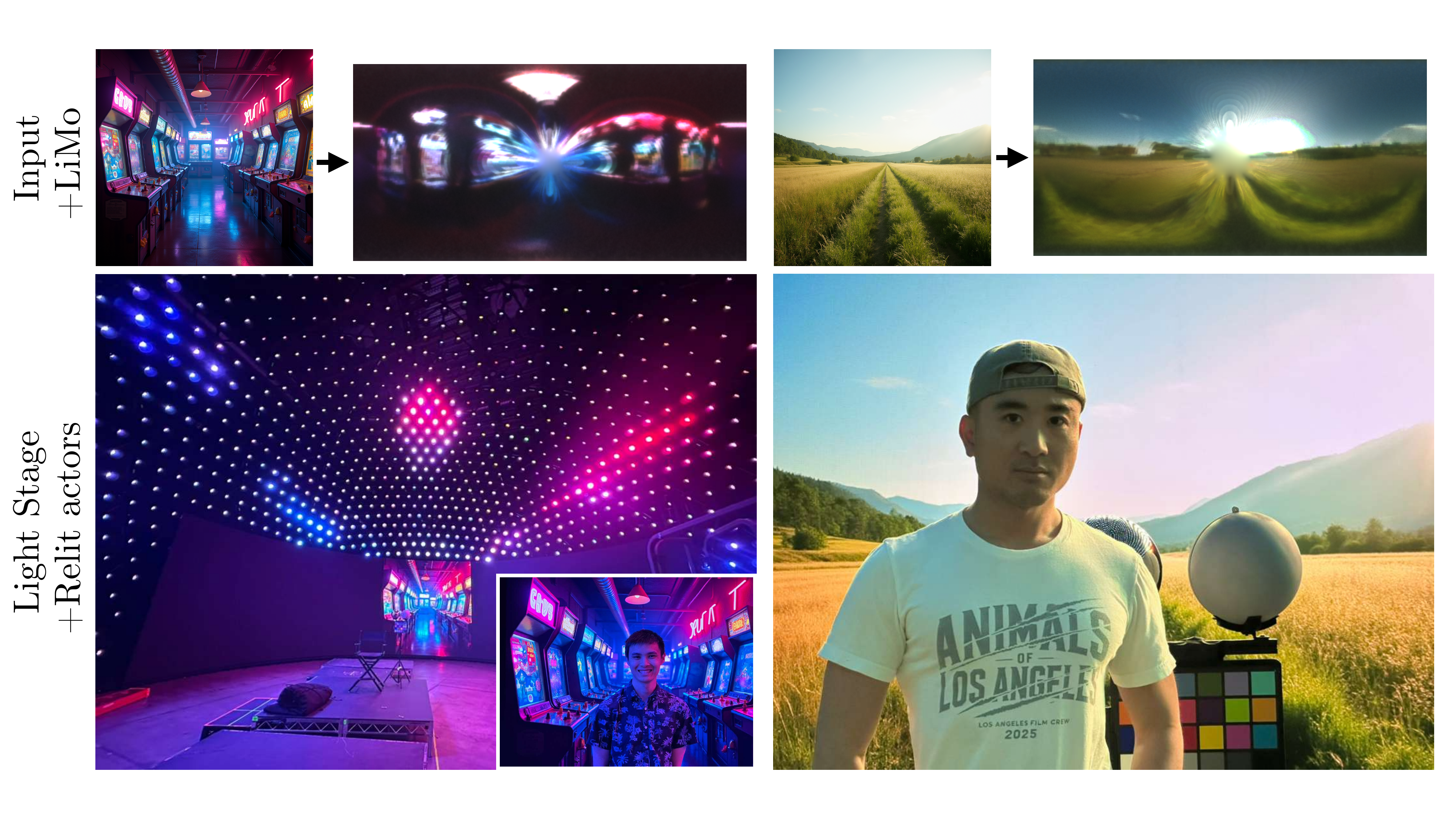}
    \caption{ \themethod is readily usable for virtual production. A background image is used as input to our method (top) and the prediction is displayed on a Light Stage, a dome of emitting Lighting, (bottom left) to relight actors (bottom middle and right). }
    \label{fig:virtual_prod}
\end{figure}

Additionally, we show in \cref{fig:virtual_prod} how \themethod integrates directly into existing image-based lighting workflows for virtual production, using predicted HDRis to relight actors in a given background environment.

\subsection{Video results}

\begin{figure*}
   \centering
   \footnotesize
   \setlength{\tabcolsep}{0.5pt}
   \setlength{\tmplength}{0.11\linewidth}

\begin{tabular}{>{\centering\arraybackslash}m{1.1cm} m{1.94cm}m{1.94cm}m{1.94cm}m{1.94cm} p{0.1cm} m{1.94cm}m{1.94cm}m{1.94cm}m{1.94cm}}
\multirow{1}{*}{Scene} &
\includegraphics[width=\tmplength]{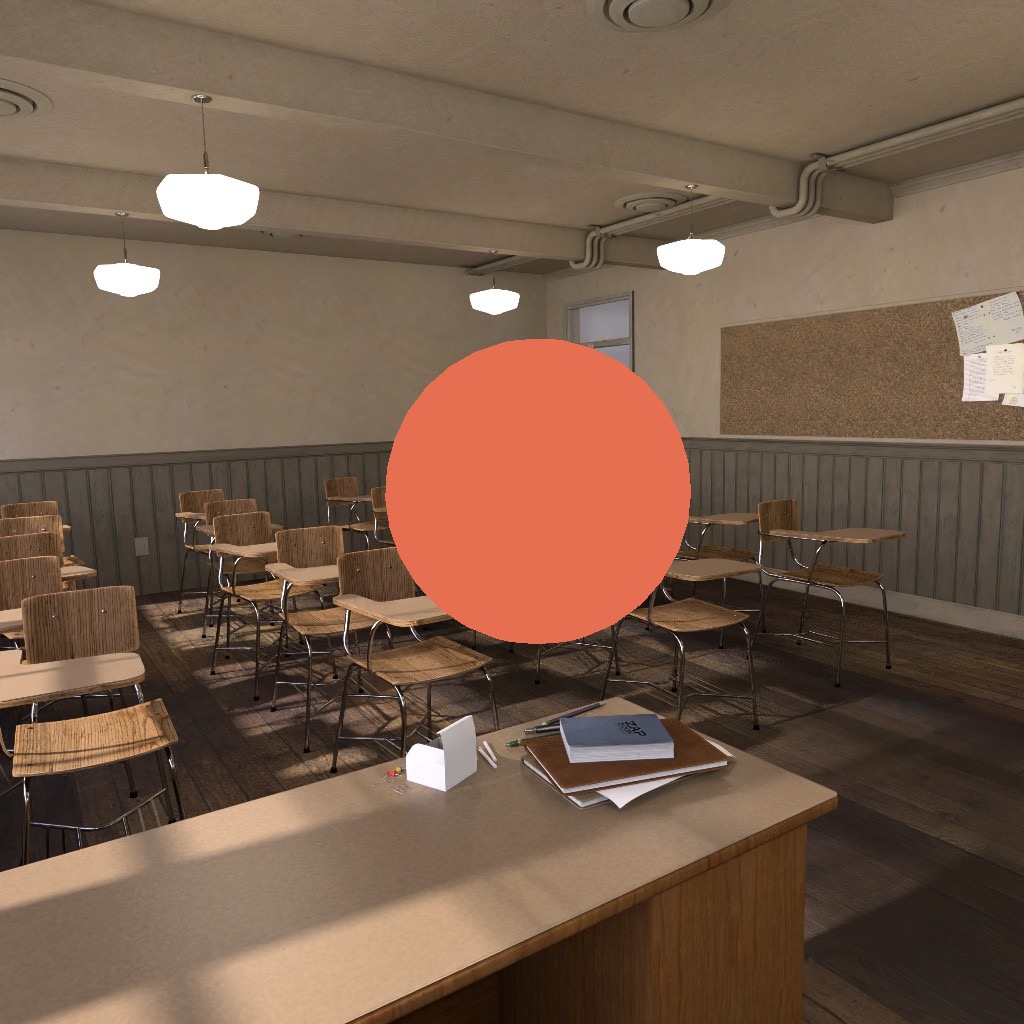} &
\includegraphics[width=\tmplength]{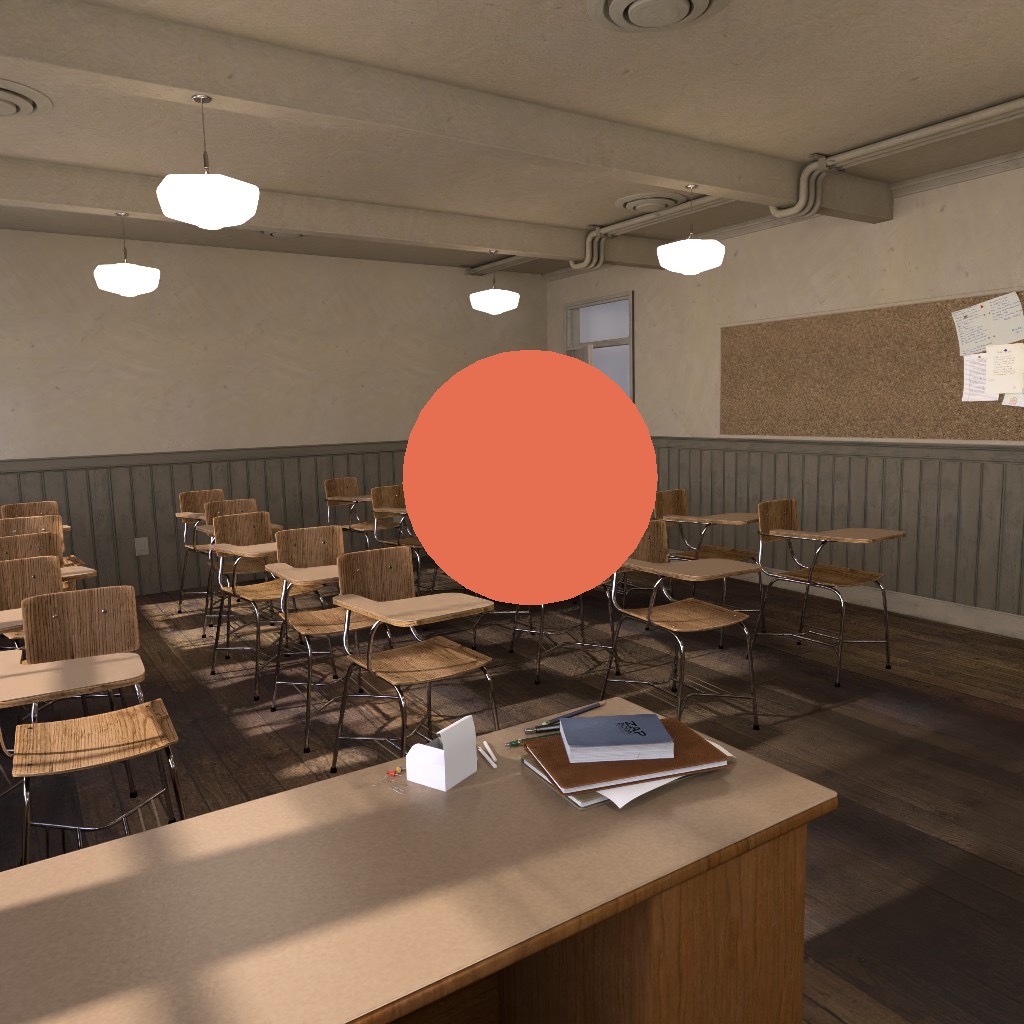} &
\includegraphics[width=\tmplength]{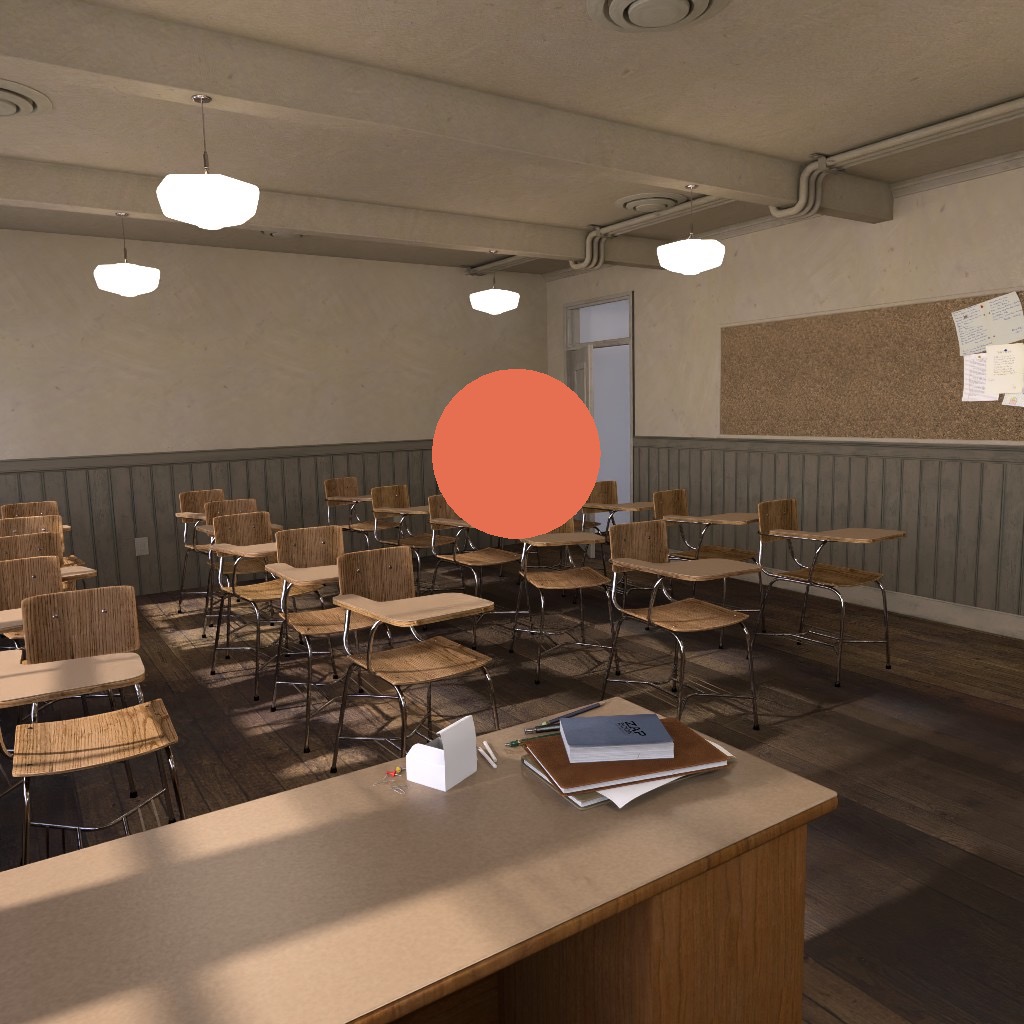} &
\includegraphics[width=\tmplength]{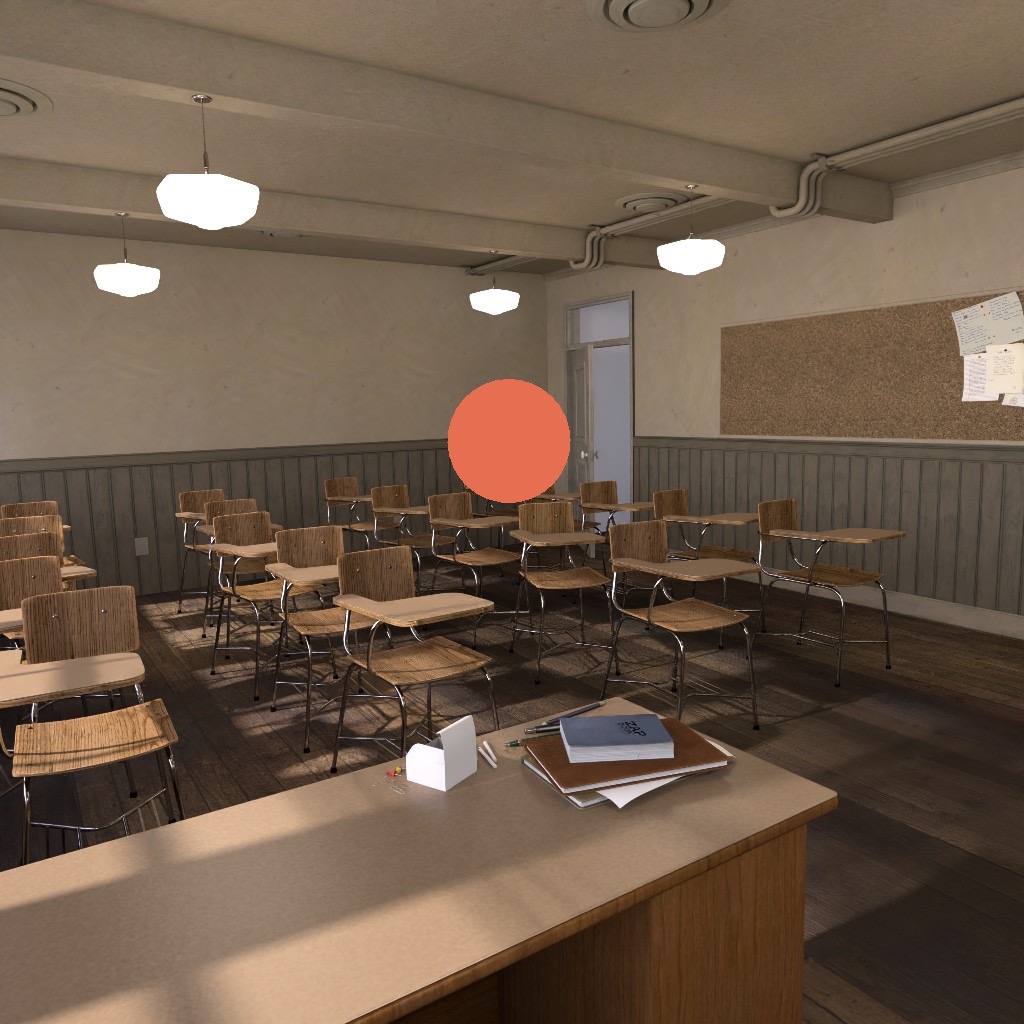} &&
\includegraphics[width=\tmplength]{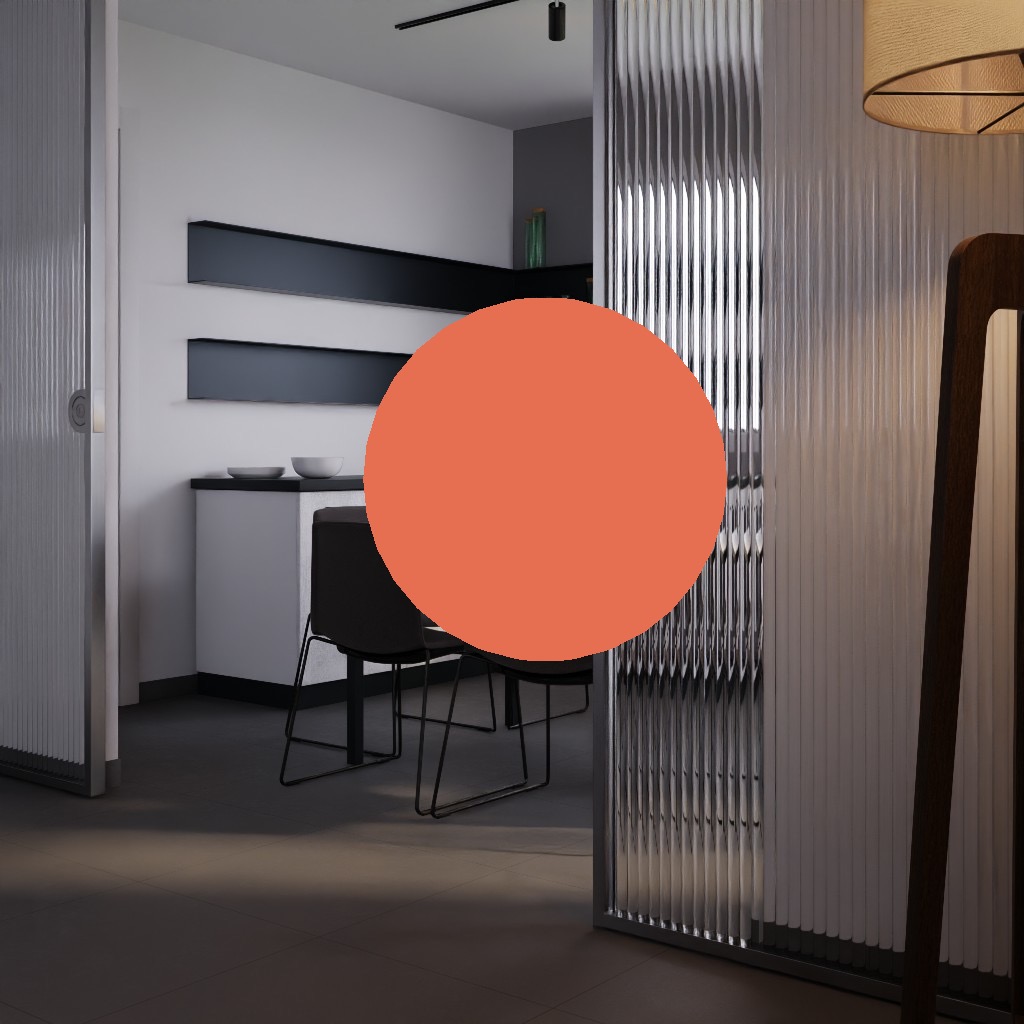} &
\includegraphics[width=\tmplength]{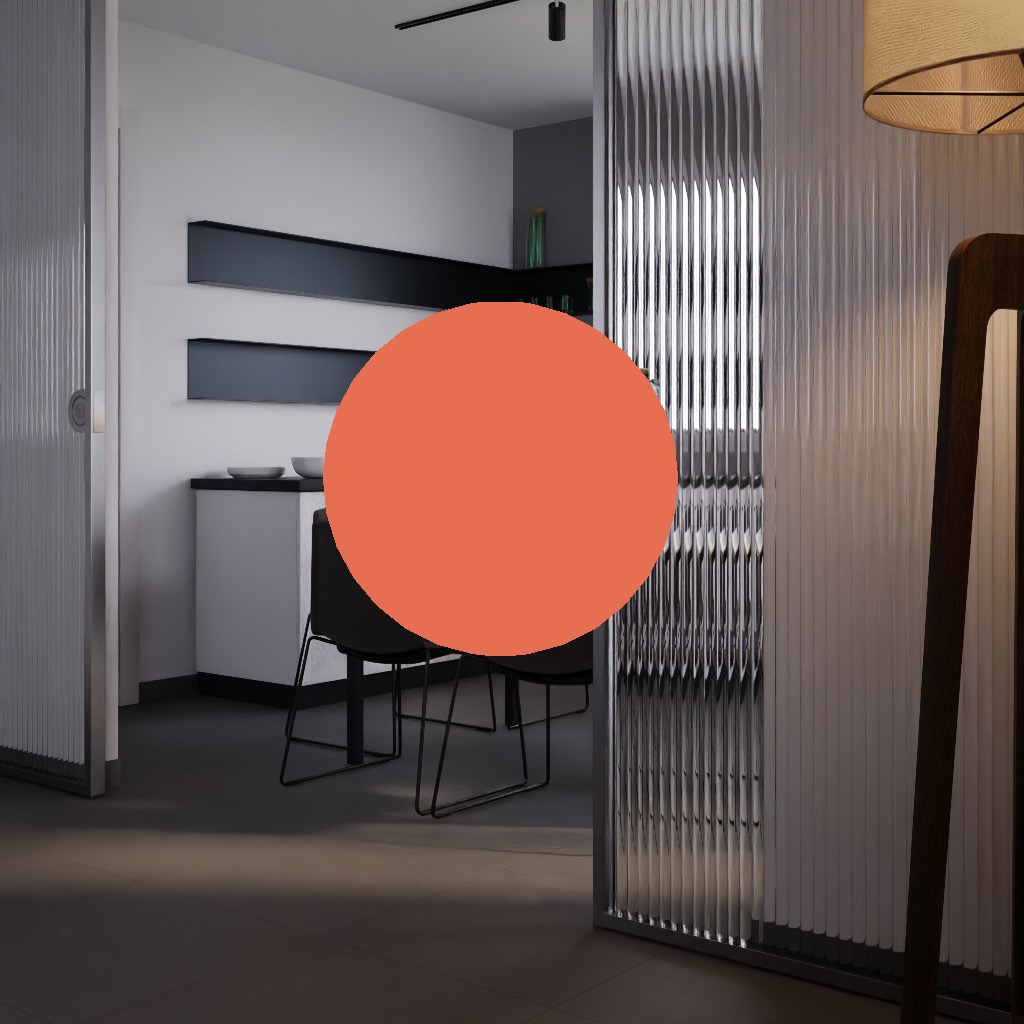} &
\includegraphics[width=\tmplength]{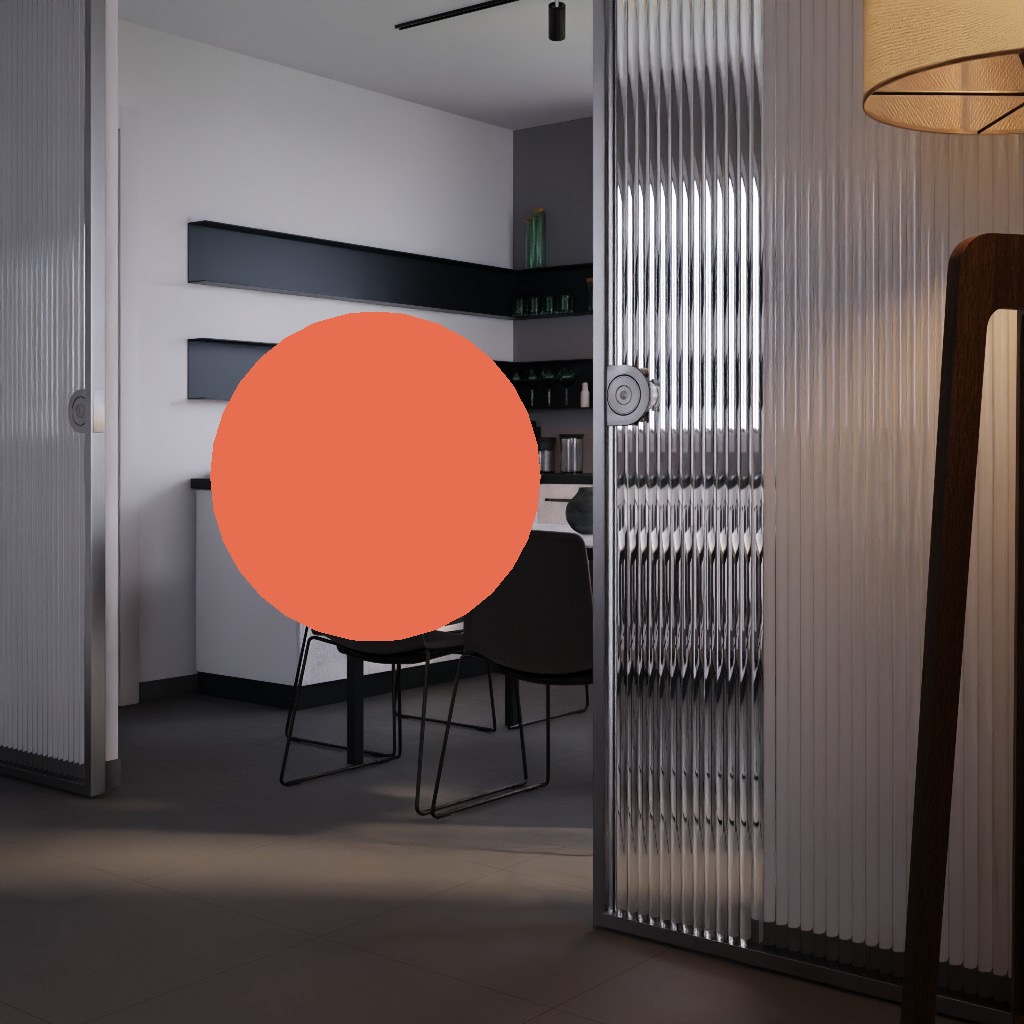} &
\includegraphics[width=\tmplength]{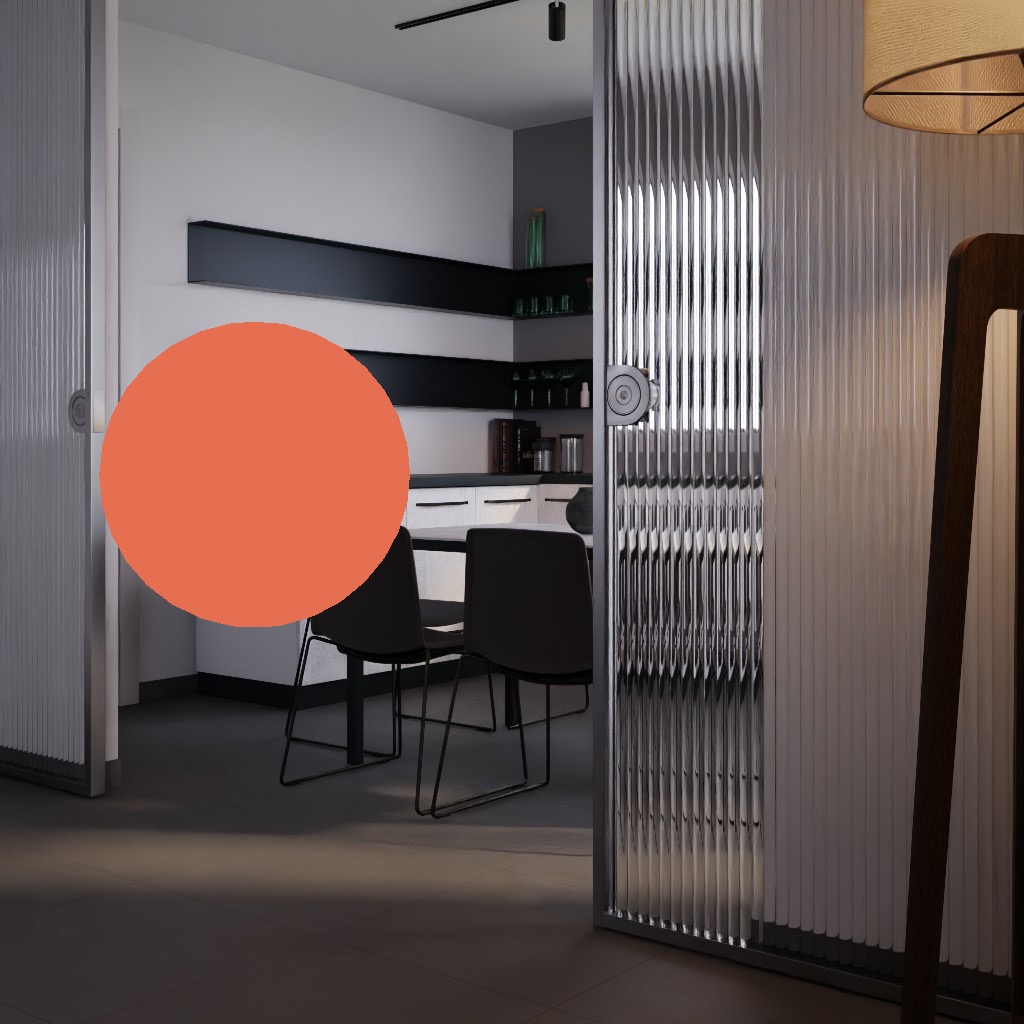}
\\

4D \mbox{Lighting} &
\includegraphics[width=\tmplength]{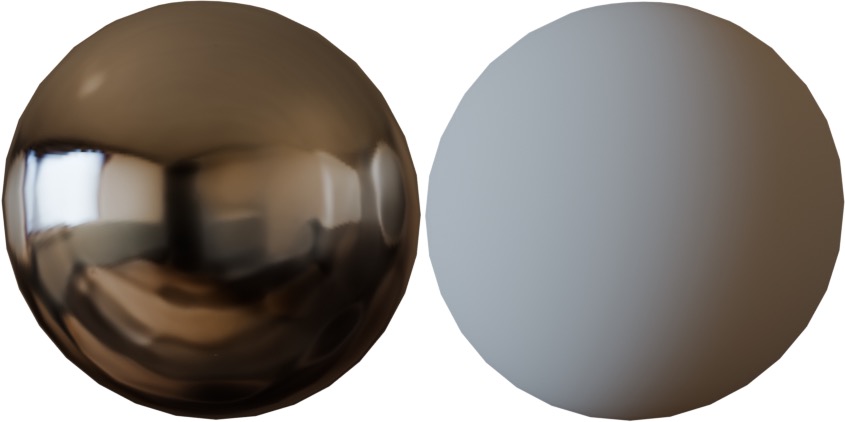} &
\includegraphics[width=\tmplength]{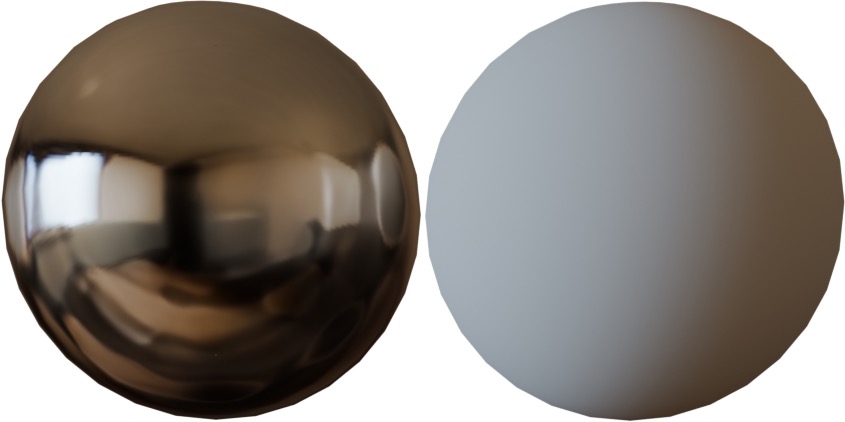} &
\includegraphics[width=\tmplength]{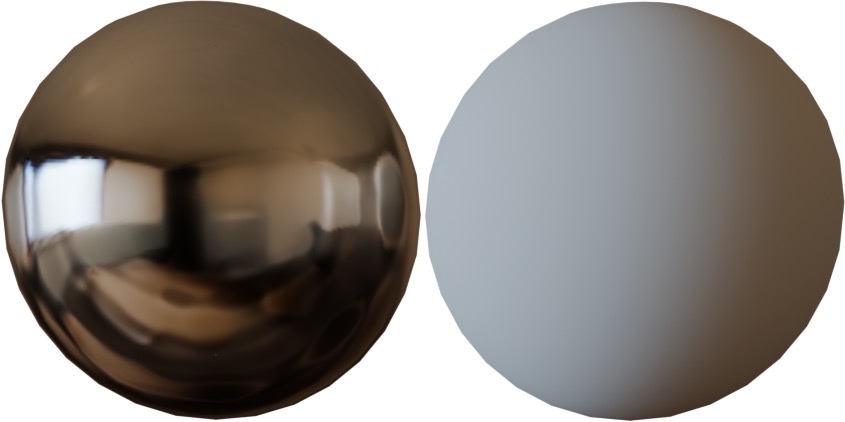} &
\includegraphics[width=\tmplength]{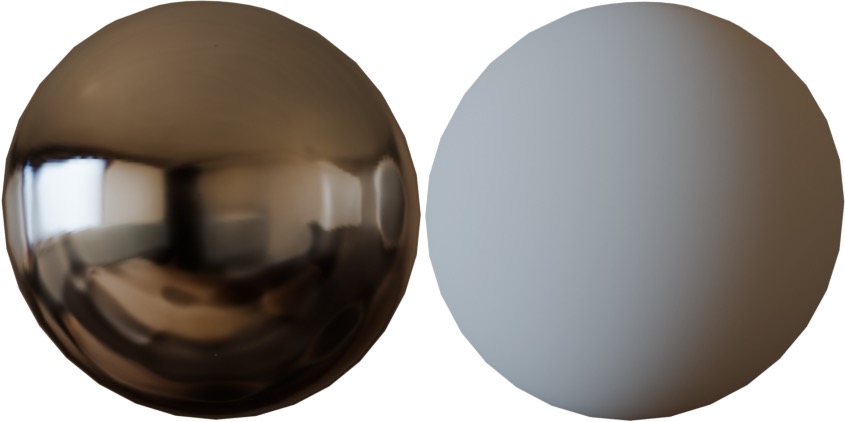} &&
\includegraphics[width=\tmplength]{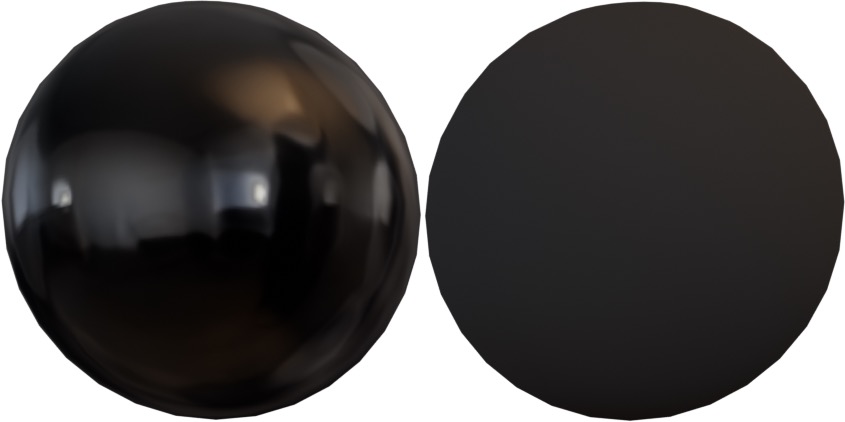} &
\includegraphics[width=\tmplength]{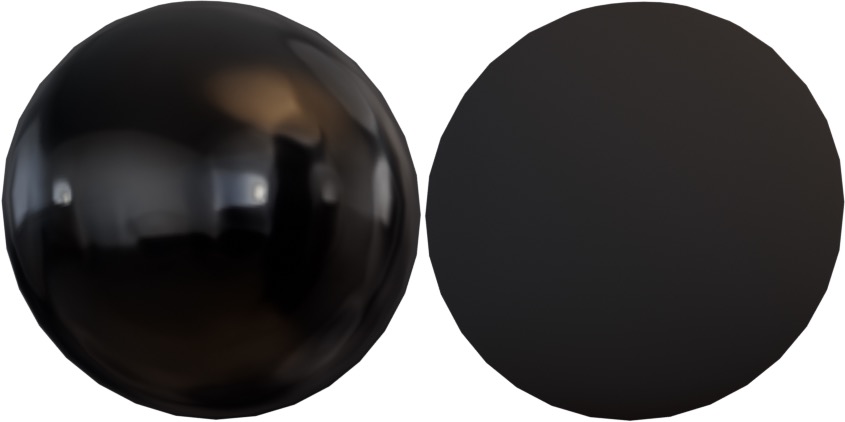} &
\includegraphics[width=\tmplength]{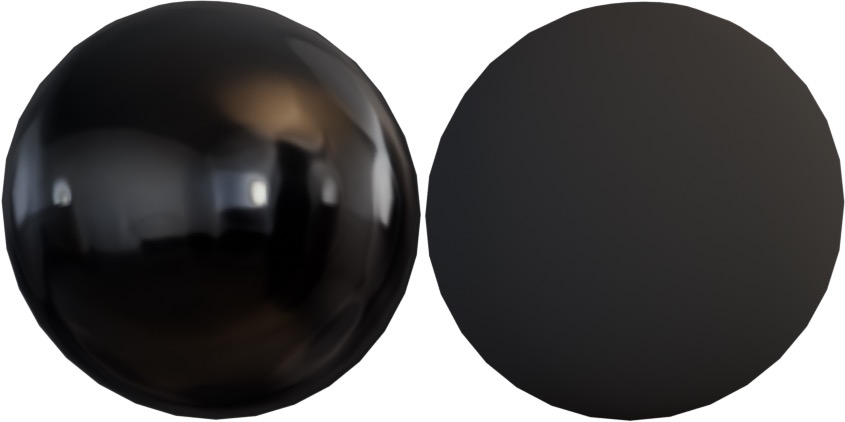} &
\includegraphics[width=\tmplength]{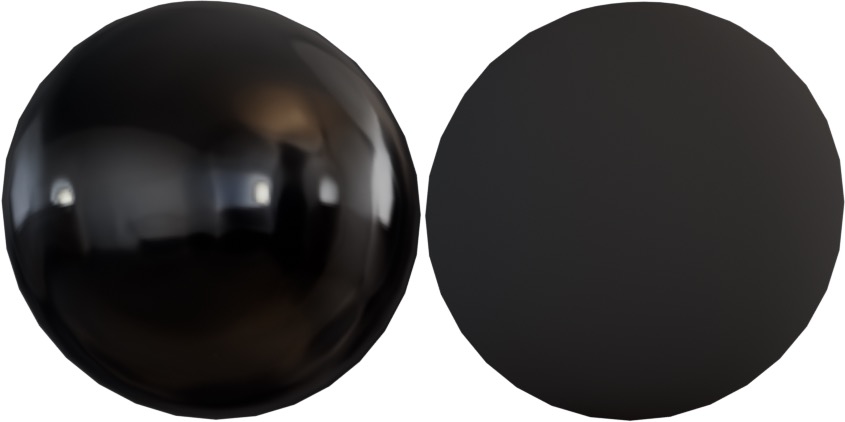}
\\

\themethod \mbox{(image)} &
\includegraphics[width=\tmplength]{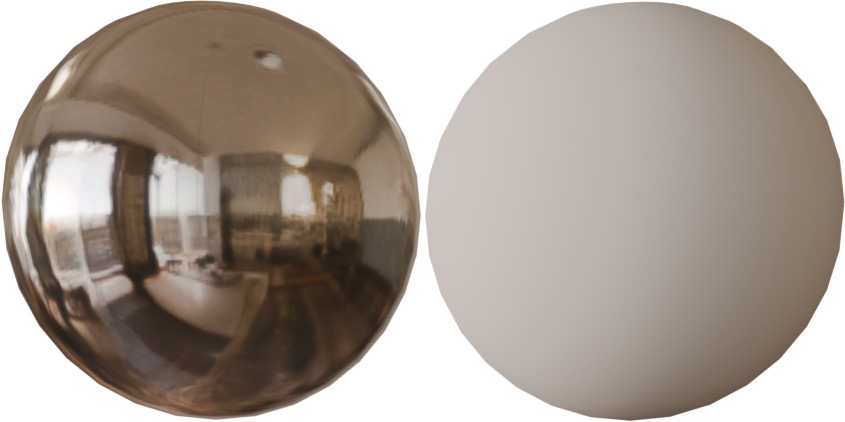} &
\includegraphics[width=\tmplength]{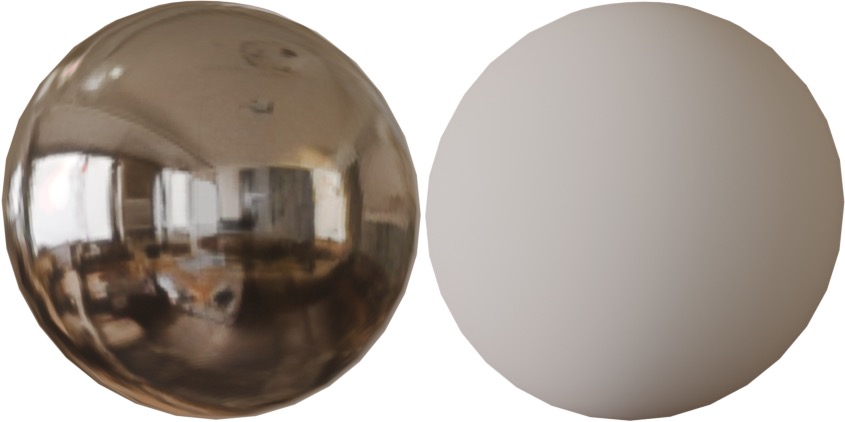} &
\includegraphics[width=\tmplength]{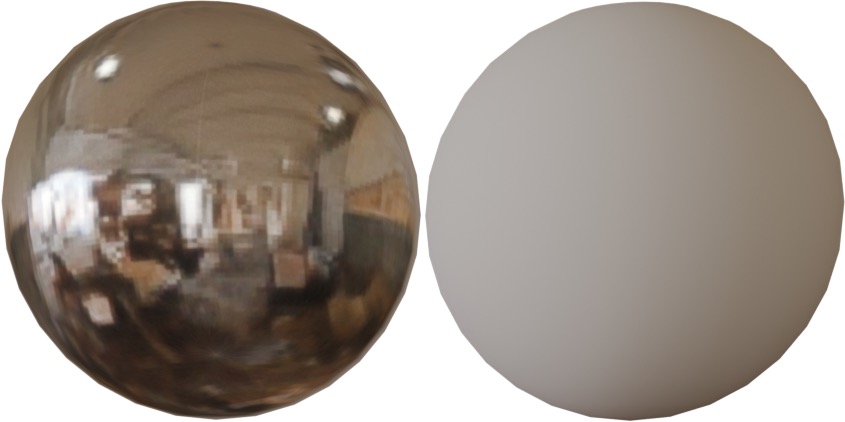} &
\includegraphics[width=\tmplength]{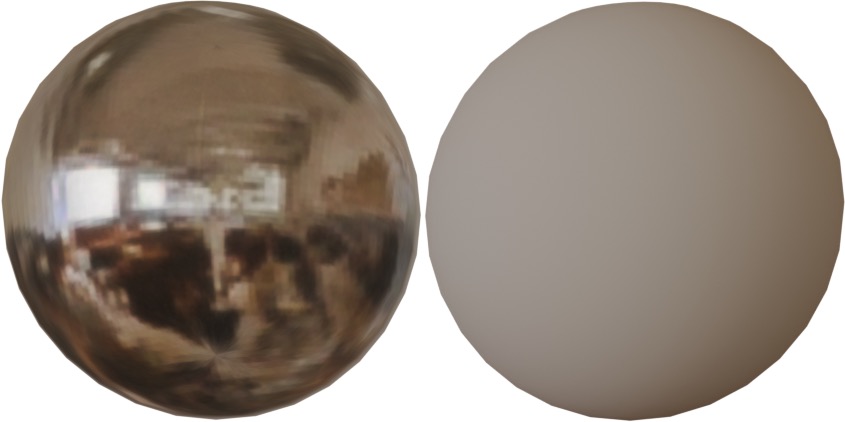} &&
\includegraphics[width=\tmplength]{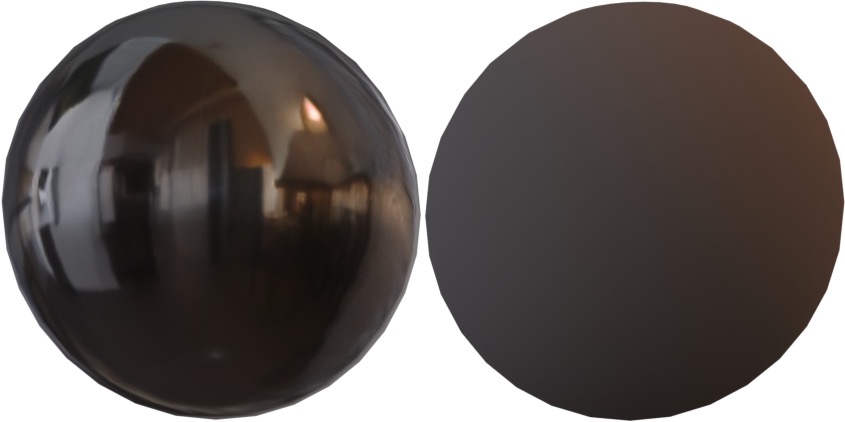} &
\includegraphics[width=\tmplength]{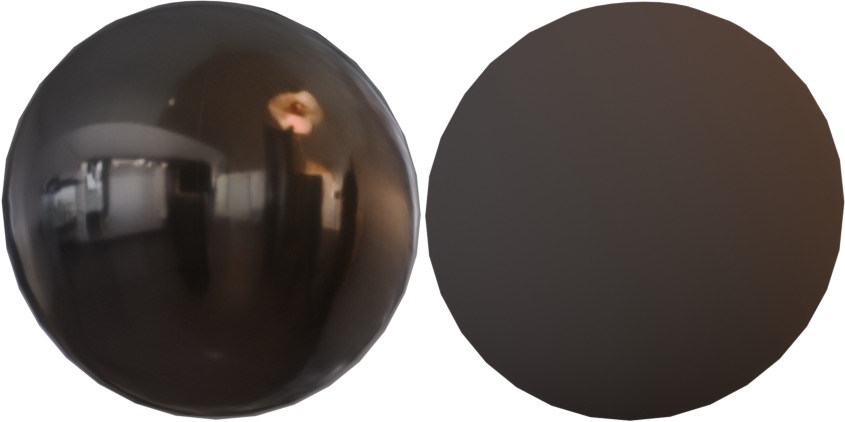} &
\includegraphics[width=\tmplength]{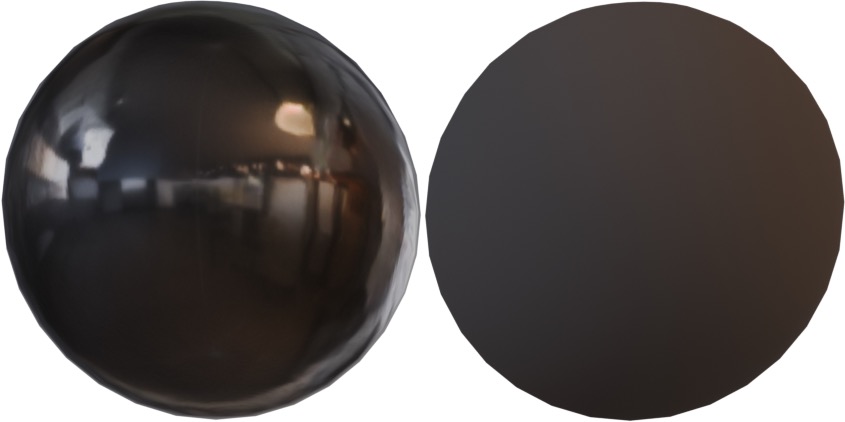} &
\includegraphics[width=\tmplength]{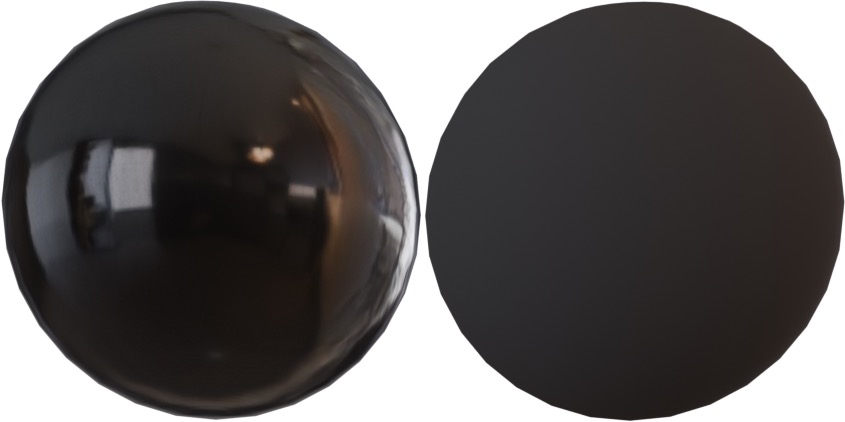}
\\

\themethod (video) &
\includegraphics[width=\tmplength]{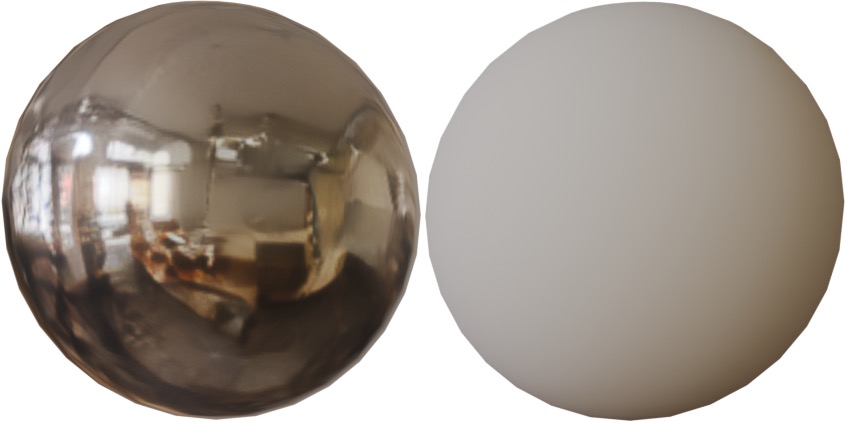} &
\includegraphics[width=\tmplength]{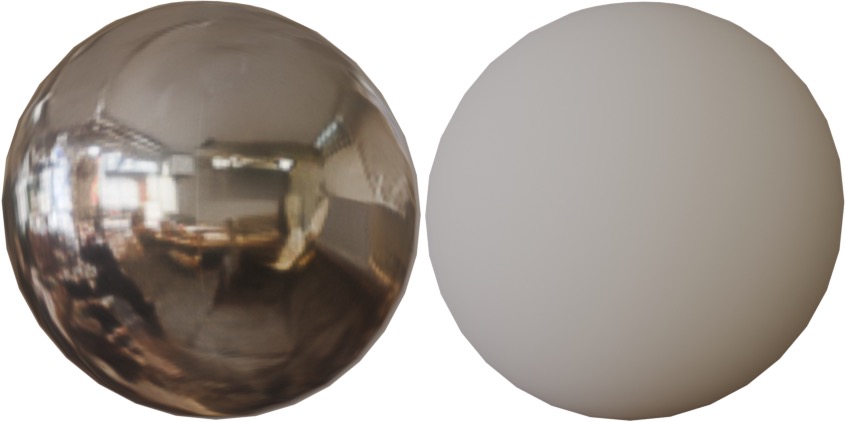} &
\includegraphics[width=\tmplength]{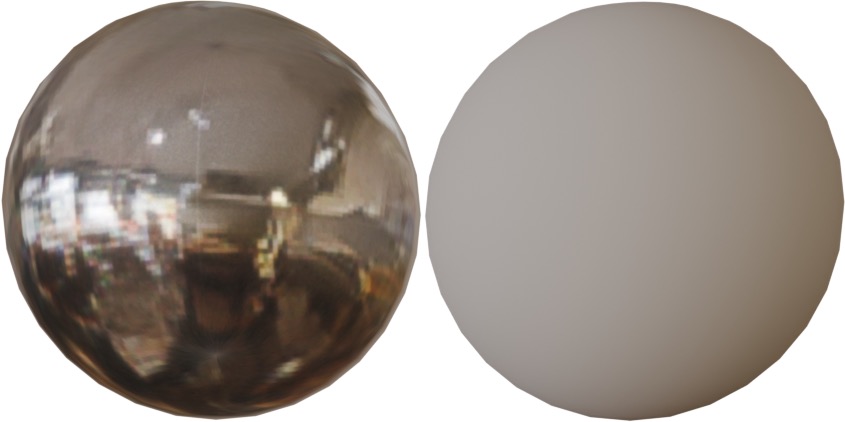} &
\includegraphics[width=\tmplength]{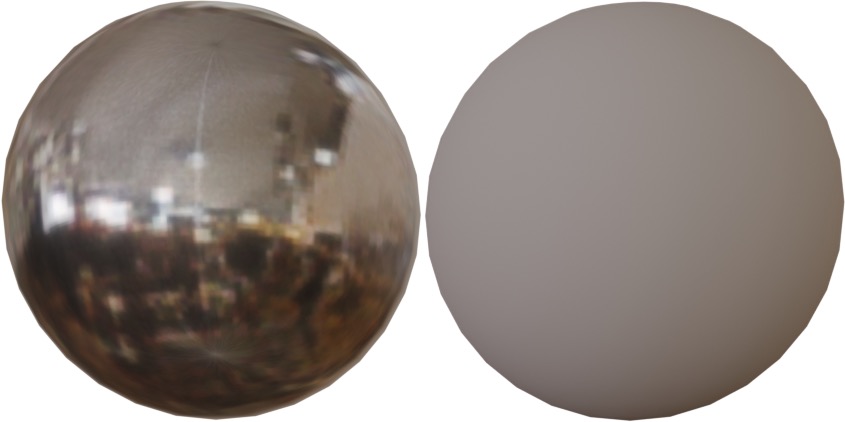} &&
\includegraphics[width=\tmplength]{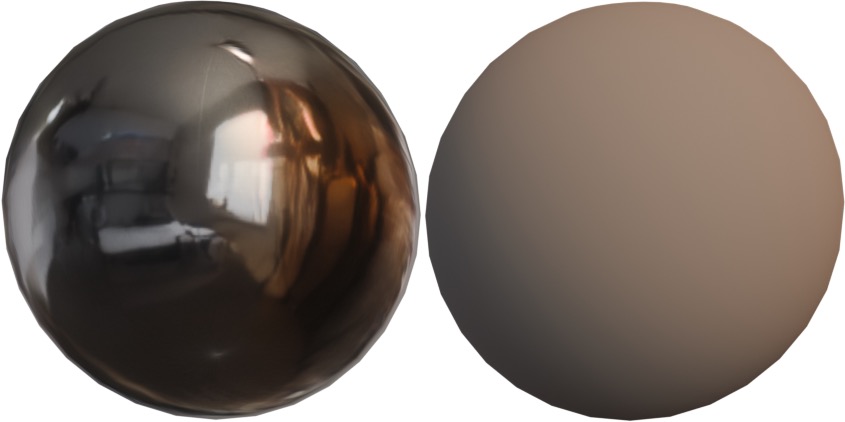} &
\includegraphics[width=\tmplength]{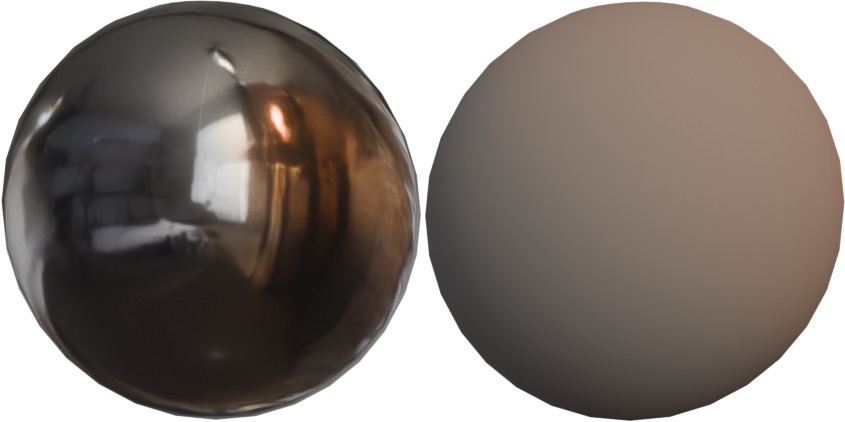} &
\includegraphics[width=\tmplength]{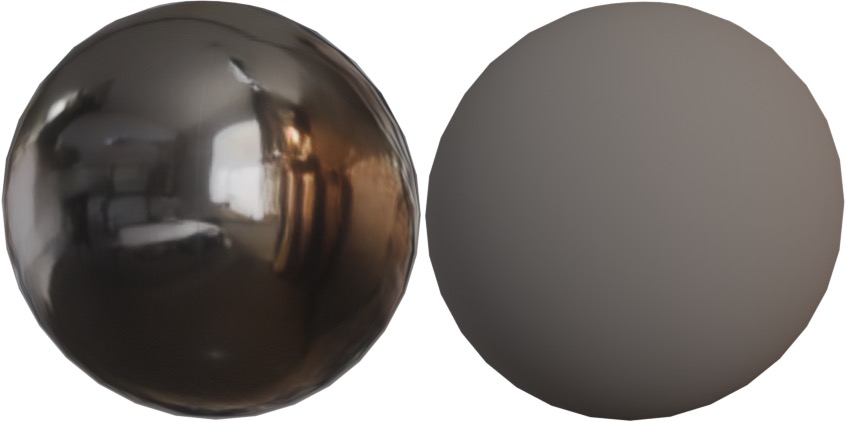} &
\includegraphics[width=\tmplength]{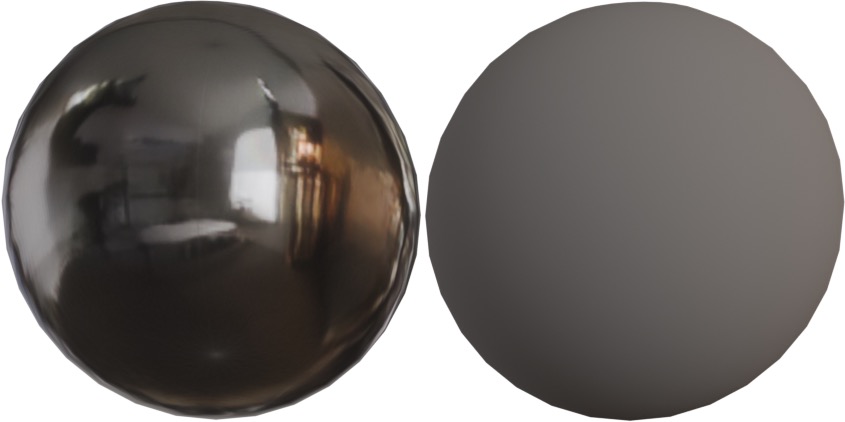}
\\

GT &
\includegraphics[width=\tmplength]{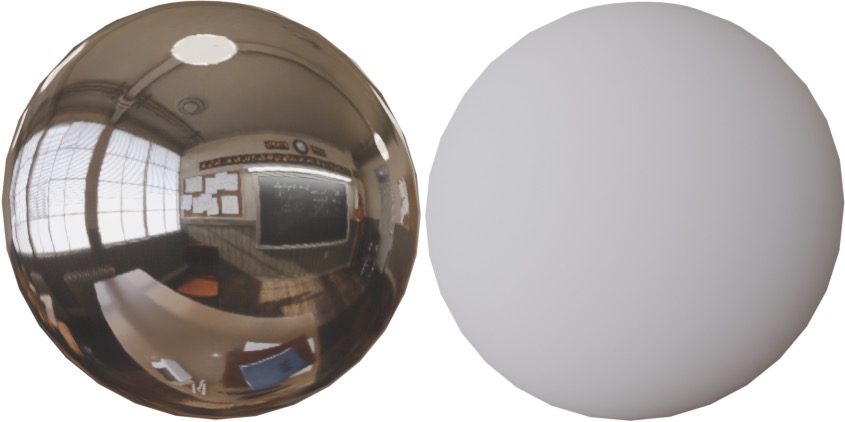} &
\includegraphics[width=\tmplength]{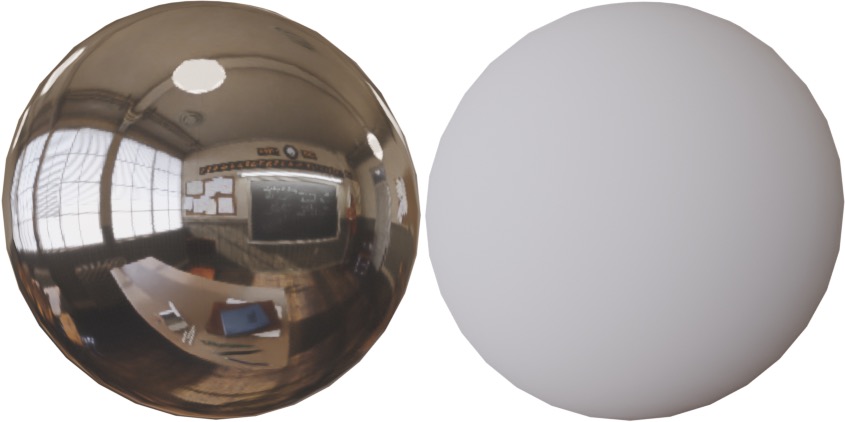} &
\includegraphics[width=\tmplength]{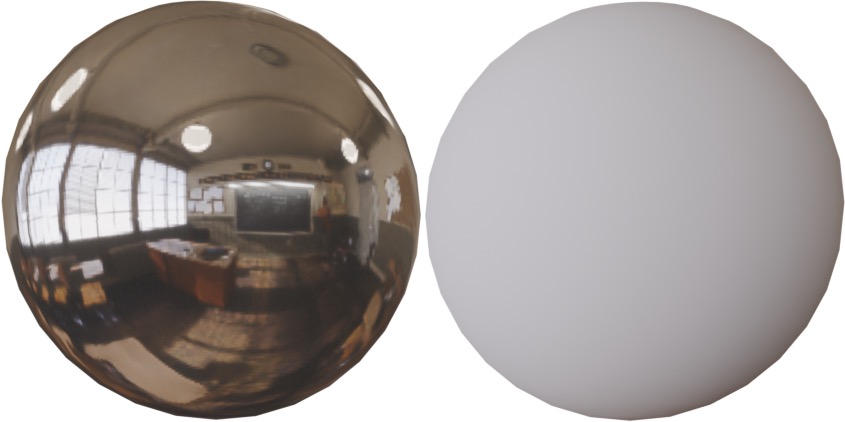} &
\includegraphics[width=\tmplength]{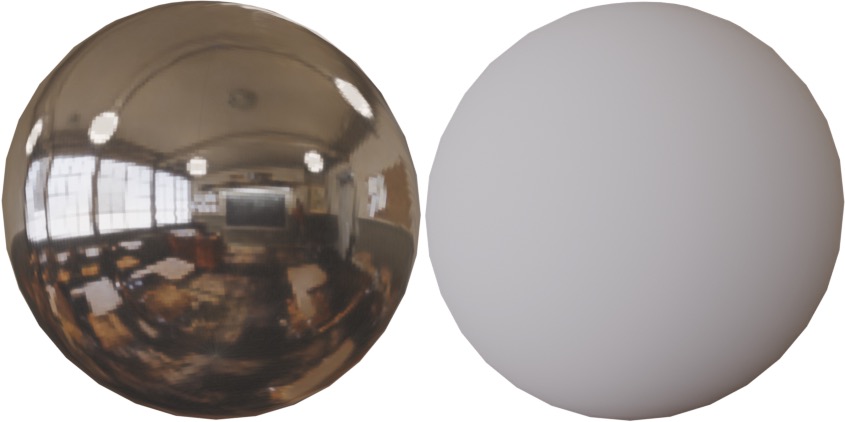} &&
\includegraphics[width=\tmplength]{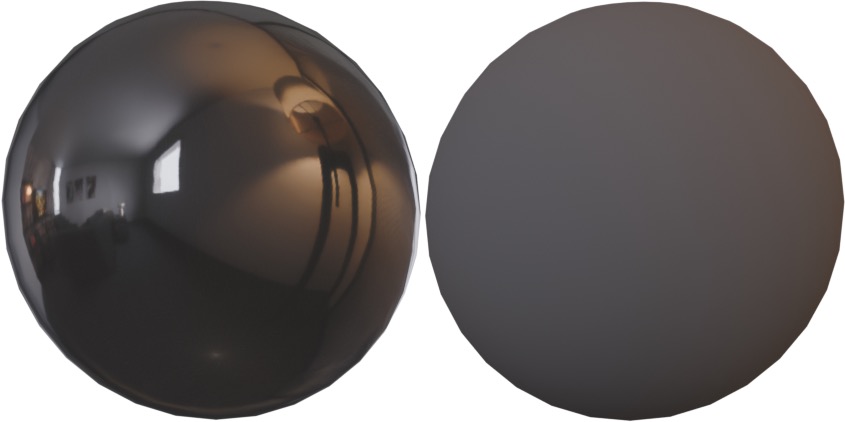} &
\includegraphics[width=\tmplength]{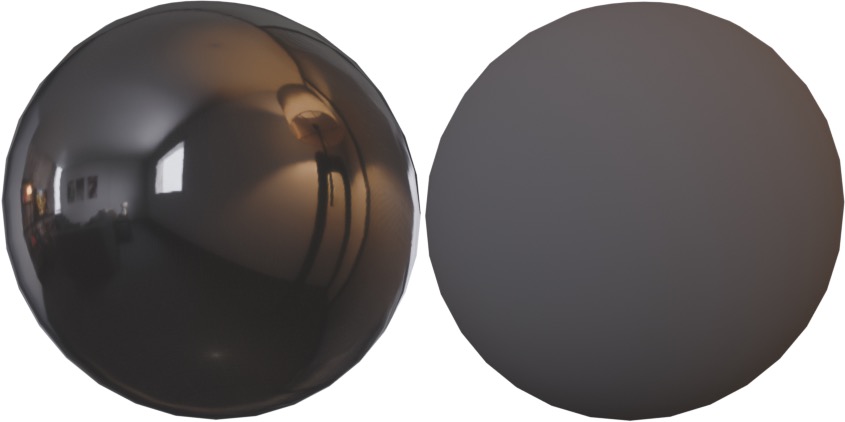} &
\includegraphics[width=\tmplength]{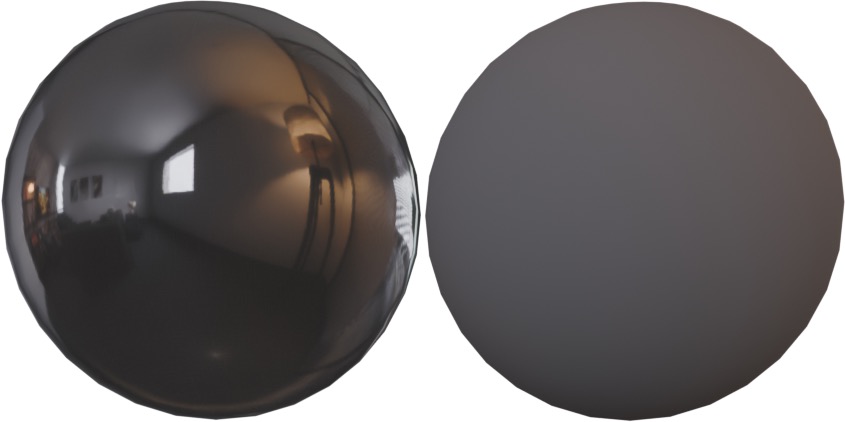} &
\includegraphics[width=\tmplength]{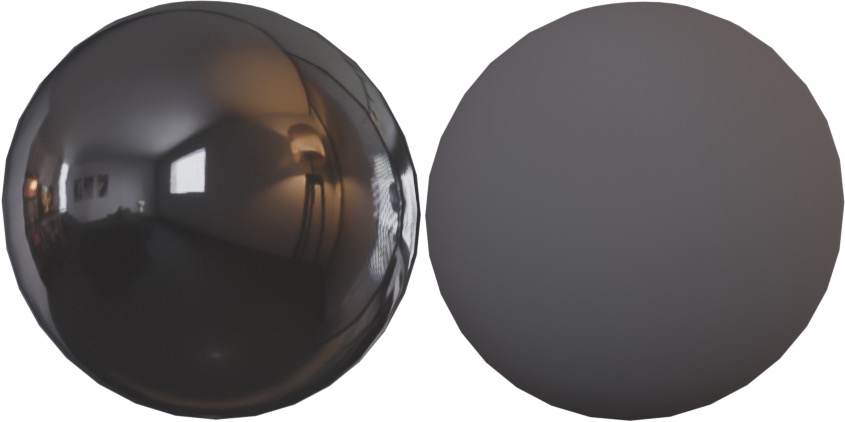}
\\

\end{tabular}

\caption{Example qualitative prediction results on our proposed video test set. Observe how our predictions are more detailed and more closely match the ground truth than the previous work 4D Lighting~\cite{4DLighting} as the sphere moves around the scene.}
\label{fig:sequence_qualitative}
\end{figure*}

To evaluate the temporal results of our method, we devise four test cases: dynamic object, dynamic camera, dynamic lighting and a combination of the above.
Our novel test set, based on 5 augmented Blender demo files, is used for evaluation.
The scores presented in \cref{tab:quant_temporal_img} tell a similar story for the metrics per-frame, where our image model outperforms 4D Lighting, with our video model close second.
However, for the three temporal metrics, the video model beats the image model.
T-LPIPS, a typical measurement of lighting consistency, is missleading as a certain amount of motion is expected.
To compensate, the T-LPIPS-Diff metric compares the T-LPIPS of the prediction to that of the ground truth.
Here we see that in every mirror render, 4D Lighting does not vary as much as it should, and our video model is equal or close behind for diffuse renderings. Although we observe lower warped L2 error metrics with 4D Lighting for some experiments, we attribute them to over-smoothing from the MLP formulation. This can be seen in our lower temporal metrics for the lighting change scenario where abrupt discontinuities are required.
Moreover, to demonstrate the capabilities of \themethod, \cref{fig:sequence_qualitative} shows samples from the test dataset.
Of note is the inability of 4D Lighting to vary the lighting appropriately as the sphere is pushed farther into the scene, whereas ours is more realistic.
More in-the-wild results can be found in \cref{fig:teaser} and the supp. materials.

\begin{table*}
\centering
\footnotesize
\setlength{\tabcolsep}{1pt}
\begin{tabular}{lccccccccccccccccccc}
\midrule
& \multicolumn{4}{c}{RMSE$_\downarrow$}  & 
\multicolumn{4}{c}{Si-RMSE$_\downarrow$}  & \multicolumn{4}{c}{SSIM$_\uparrow$}   &  \multicolumn{4}{c}{Ang Err$_\downarrow$} \\
\midrule
 Method      & Mirr & Diff & Gloss & Mat  & Mirr & Diff & Gloss & Mat  & Mirr & Diff & Gloss & Mat  & Mirr & Diff & Gloss & Mat \\
 \midrule
 w/o Diffuse, Geo &   \num{0.262} & \third{\num{0.210}} & \third{\num{0.219}} & \third{\num{0.209}}& \num{0.442} & \third{\num{0.131}} & \third{\num{0.151}} & \third{\num{0.200}} & \num{0.770}  & \num{0.942} & \third{\num{0.926}} & \third{\num{0.938}} & \num{5.15} & \num{3.60} & \num{3.56} & \num{3.37}  \\
 w/o Diffuse   & \second{\num{0.253}} & \second{\num{0.207}} & \second{\num{0.215}} & \second{\num{0.204}} & \second{\num{0.431}} & \second{\num{0.127}} & \second{\num{0.145}} & \second{\num{0.189}} & \second{\num{0.776}}  & \second{\num{0.943}} & \second{\num{0.929}} & \second{\num{0.939}} & \third{\num{4.95}} & \third{\num{3.39}} & \third{\num{3.35}} & \third{\num{3.19}}  \\
 w/o Geo  &  \third{\num{0.259}} & \num{0.229} & \num{0.230} & \num{0.213}& \second{\num{0.431}} & \num{0.137} & \num{0.162} &\num{0.211} & \third{\num{0.772}}  & \third{\num{0.936}} & \num{0.923} & \num{0.935} & \second{\num{4.77}} & \second{\num{3.13}} & \second{\num{3.04}} & \second{\num{3.07}}  \\
 \themethod (full)  &  \best{\num{0.247}} & \best{\num{0.160}} & \best{\num{0.164}} & \best{\num{0.169}} &  \best{\num{0.403}} & \best{\num{0.107}} & \best{\num{0.129}} & \best{\num{0.176}} & \best{\num{0.783}}  & \best{\num{0.951}} & \best{\num{0.939}} & \best{\num{0.946}} & \best{\num{4.35}} & \best{\num{2.25}} & \best{\num{2.20}} & \best{\num{2.42}}  \\
\end{tabular}
\caption{Ablation of the use of the added geometric maps $I_{\text{dir}}$ and $I_{\text{dist}}$ for predictions (see \cref{sec:conditions}) and diffuse sphere for HDRI optimization (see \cref{sec:general_approach}) on Infinigen with our image model. ``Mirr'' (mirror), ``Diff'' (diffuse), ``Gloss'' (glossy) and ``Mat'' (matte) refer to the different test spheres (see \cref{sec:metrics_datasets}).  Results are color coded by \best{best}, \second{second} and \third{third} best.}
\label{tab:ablation}
\end{table*}

\subsection{Ablations}
\label{sec:ablation}
To justify our design choices, we ablate the use of the diffuse sphere and of our novel geometric maps. All ablations are performed with the \themethod (image) model on the Infinigen test set. Metrics are reported in \cref{tab:ablation}, where we observe that our full model performs better in every scenario.

\myparagraph{Predicting the diffuse sphere}
The metrics validate the effectiveness of using the predicted Diffuse sphere in conjunction with the mirror sphere for the optimization of the final HDRI. Notably, the color prediction, as informed from the angular error, is worse when the diffuse sphere is omitted.

\begin{figure}
    \centering
    \footnotesize
    \setlength{\tabcolsep}{2pt}
    \setlength{\fboxsep}{0pt}
    \setlength{\mywidth}{0.3\linewidth}
    \begin{tabular}{cccc}
    & Depth maps & w/o geo maps & w/ geo maps \\*[0.5em]
    \rot{ \hspace{25pt} Near} &
    \includegraphics[width=\mywidth]{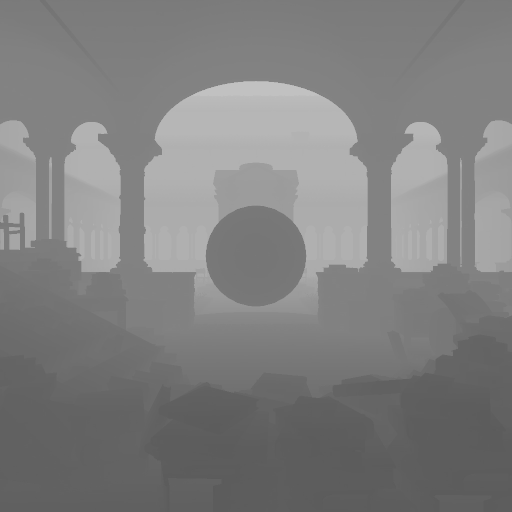} &
    \begin{overpic}[width=\mywidth]{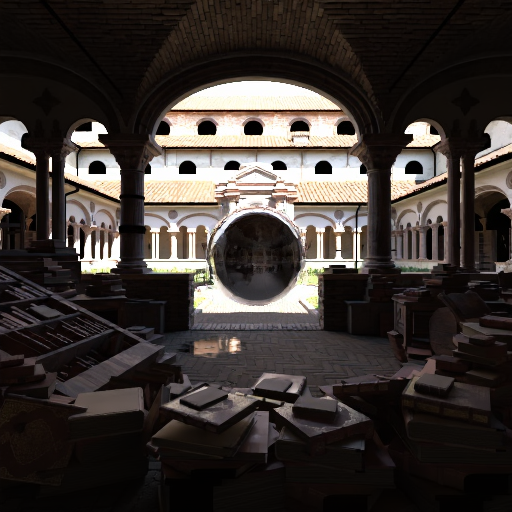}\put(65,65){\fbox{\includegraphics[width=0.4\mywidth, trim=200px 200px 200px 200px, clip]{figures/ablation_depth/depth_close.png}}}\end{overpic} &
    \begin{overpic}[width=\mywidth]{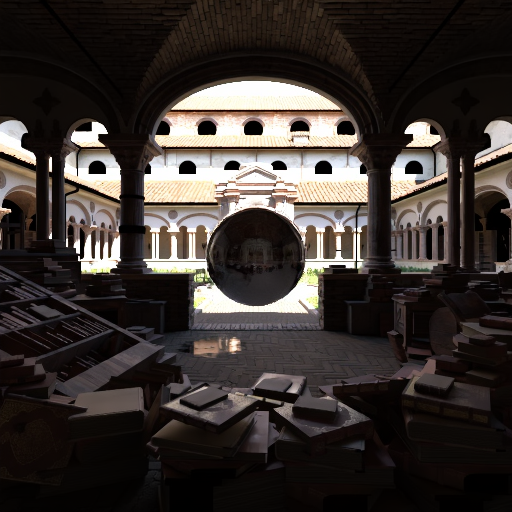}\put(65,65){\fbox{\includegraphics[width=0.4\mywidth, trim=200px 200px 200px 200px, clip]{figures/ablation_depth/geo_close.png}}}\end{overpic} \\*[0.25em]
    \rot{ \hspace{25pt} Far} &
    \includegraphics[width=\mywidth]{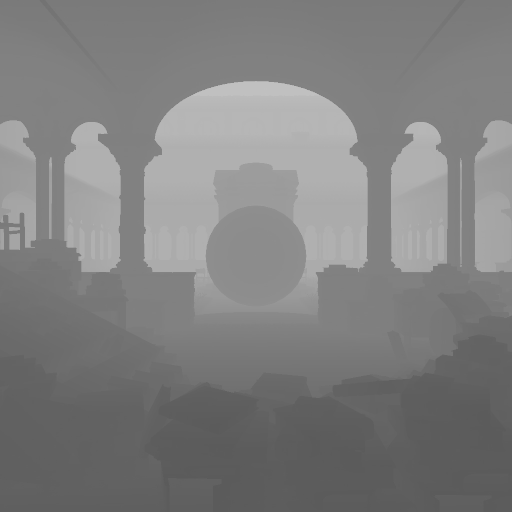} &
    \begin{overpic}[width=\mywidth]{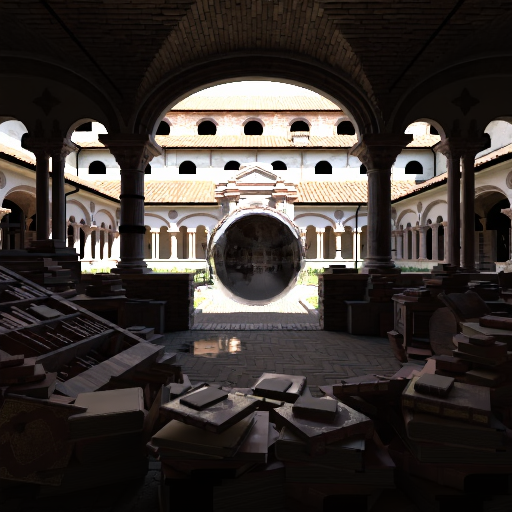}\put(65,65){\fbox{\includegraphics[width=0.4\mywidth, trim=200px 200px 200px 200px, clip]{figures/ablation_depth/depth_far.png}}}\end{overpic} &
    \begin{overpic}[width=\mywidth]{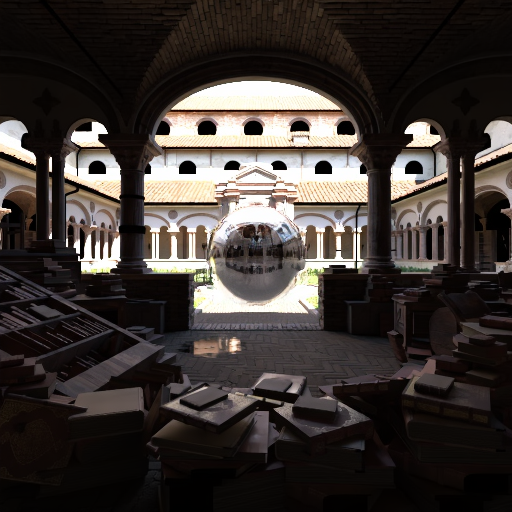}\put(65,65){\fbox{\includegraphics[width=0.4\mywidth, trim=200px 200px 200px 200px, clip]{figures/ablation_depth/geo_far.png}}}\end{overpic} 
    \end{tabular}
    \caption{The effect of our proposed geometric maps $\{I_{\text{dir}}, I_{\text{dist}}\}$. We insert a sphere near (top) and far (bottom). Observe how the addition of geometric maps (right) helps the network in reasoning about light occlusions, which are not captured otherwise (middle).}
    \label{fig:effect_vector}
\end{figure}

\myparagraph{Geometric maps}
In \cref{sec:conditions} we observed that depth maps of the spatial position are insufficient for the network to properly inpaint a correctly-placed sphere. This observation is illustrated in \cref{fig:effect_vector}, where a sphere is kept the same image space size, but pushed farther into the scene, from shadow to direct sunlight. Depth conditioning is insufficient: the two inpainted spheres are nearly identical. Our novel geometric maps help the network interpret the position of the sphere in relation to the scene's elements and correctly inpaints a shadowed and lit sphere respectively. Those results are validated by the metrics, where removing the geometric maps results in worse performance than removing the diffuse sphere. Interestingly, removing both the geometric maps and the diffuse sphere results in slightly better scores than simply removing the geometric maps, justified by the fact that the diffuse sphere helps predict the correct dynamic range only, when the geometric information is correctly understood.

\section{Conclusion}

We introduced \themethod, a method for spatiotemporal scene lighting estimation.
We demonstrated that combining geometrically grounded conditioning, the priors from a pre-trained diffusion model, and multiple predictions of diffuse and mirror spheres leads to state-of-the-art image and video lighting estimation.
Our method provides a high level of physical accuracy and very good spatial understanding and temporal stability.  
Nonetheless, \themethod has some limitations.
\change{First, because the spheres in the dataset are rendered as 3D objects with non-zero size, some training samples violate the directional-lighting assumption, making the HDRI optimization ill-posed, such as when a shadow is cast directly on the sphere.}
Second, \themethod was not trained to leverage certain lighting cues such as human faces, present in many videos.
Future work could account for these by adopting a truly point-based lighting representation and leveraging face datasets with known lighting.

\myparagraph{Acknowledgements}
This work was partially done during C. Bolduc’s internship at Eyeline Labs, and partially supported by an NSERC Canada Graduate Research Scholarship to C. Bolduc.

{
    \small
    \bibliographystyle{ieeenat_fullname}
    \bibliography{main}
}

\clearpage
\setcounter{page}{1}
\maketitlesupplementary

\section{Data generation details}
We use Blender, paired with BlenderKit assets to procedurally generate indoor and outdoor renders. 
For indoor scenes, we use the full indoor scenes provided by BlenderKit, generating more cameras based on the original ones.
Since scenes are not always completely modeled, we leverage existing camera assuming they point toward points of interest.
We randomly sample a direction from the original camera frustrum to obtain a target lookat point using ray-casting.
Then we sample 3D points in the scene bounding box and check the visibility of the selected lookat point from them.
If the point is visible from the sampled position, we consider the location to be a valid camera location and create a new camera pointed at the lookat.

For outdoor scenes, we select a central model from the building, vehicle or nature categories of BlenderKit.
We add a ground plane, a random ground material, and add surrounding buildings, objects and vegetation using particle systems.
We use HDRis from Polyhaven for lighting.
We then derive cameras pointed at the central object from which it is visible.
In total we use around 500 indoor scenes, reusing them for different motion for a total of 4400 scenes, and generate 1200 outdoor scenes. For each scene we render 4 viewpoints. 

\begin{figure}[!t]
    \centering
    \setlength{\tabcolsep}{1pt}
    \setlength{\mywidth}{0.30\linewidth}
    \begin{tabular}{cccc}
    \rot{\tiny \hspace{15pt}Dynamic spheres} &
    \includegraphics[width=\mywidth]{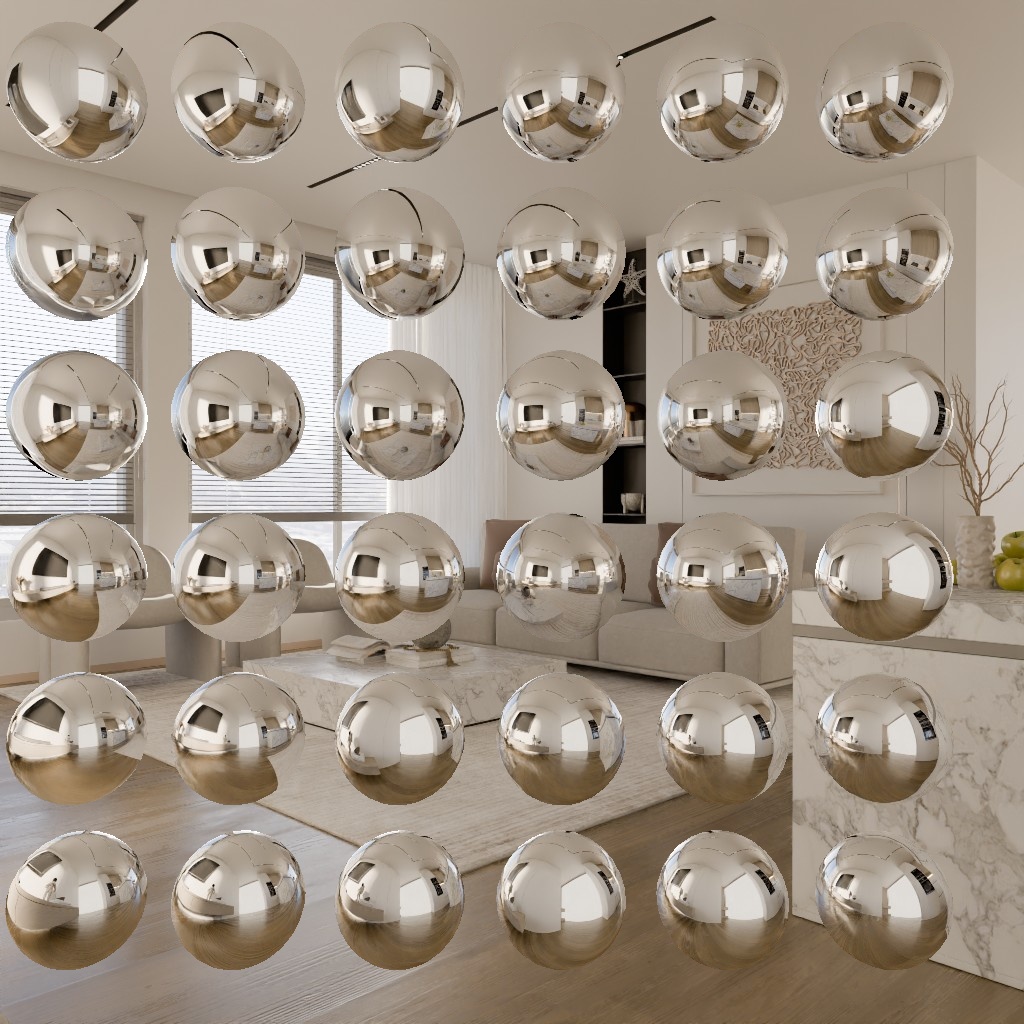} &
    \includegraphics[width=\mywidth]{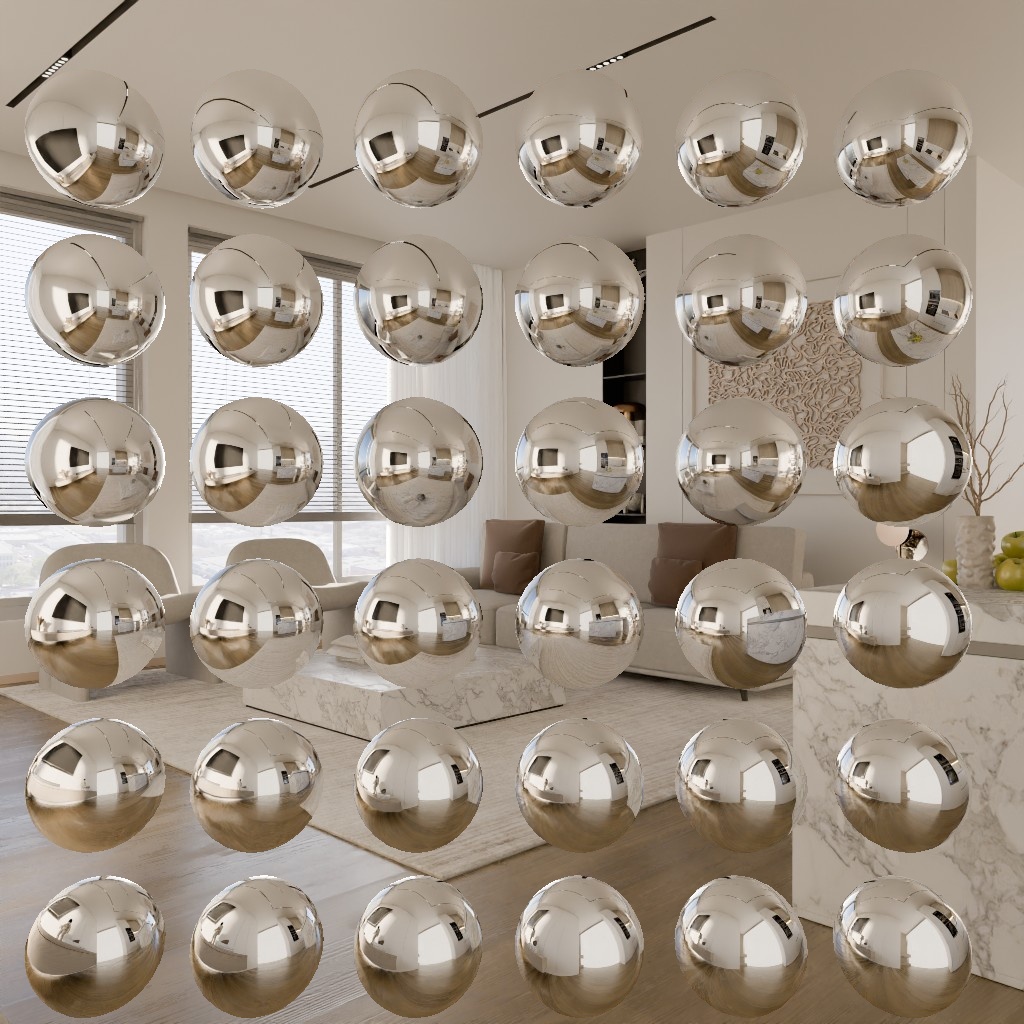} & \includegraphics[width=\mywidth]{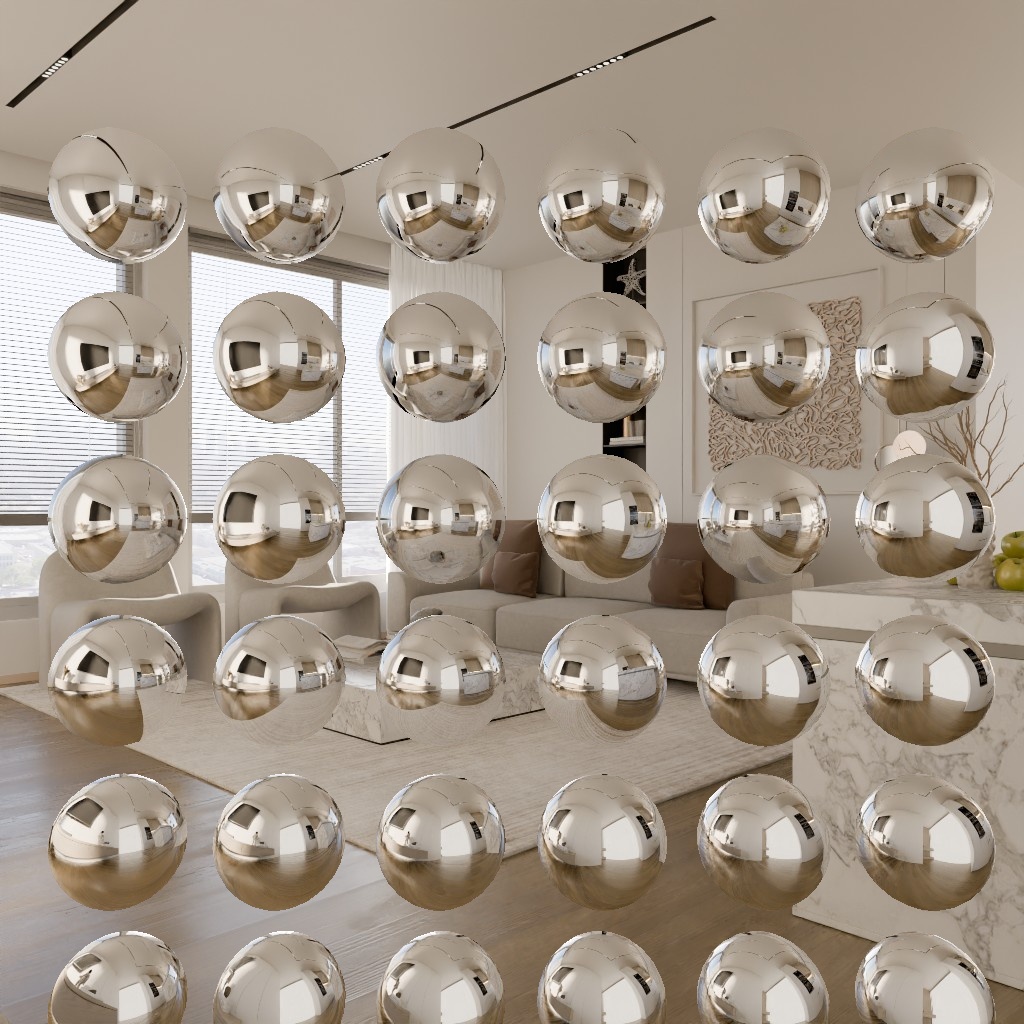}  \\*[-0.25em]
    \rot{\tiny \hspace{15pt}Dynamic camera} &
    \includegraphics[width=\mywidth]{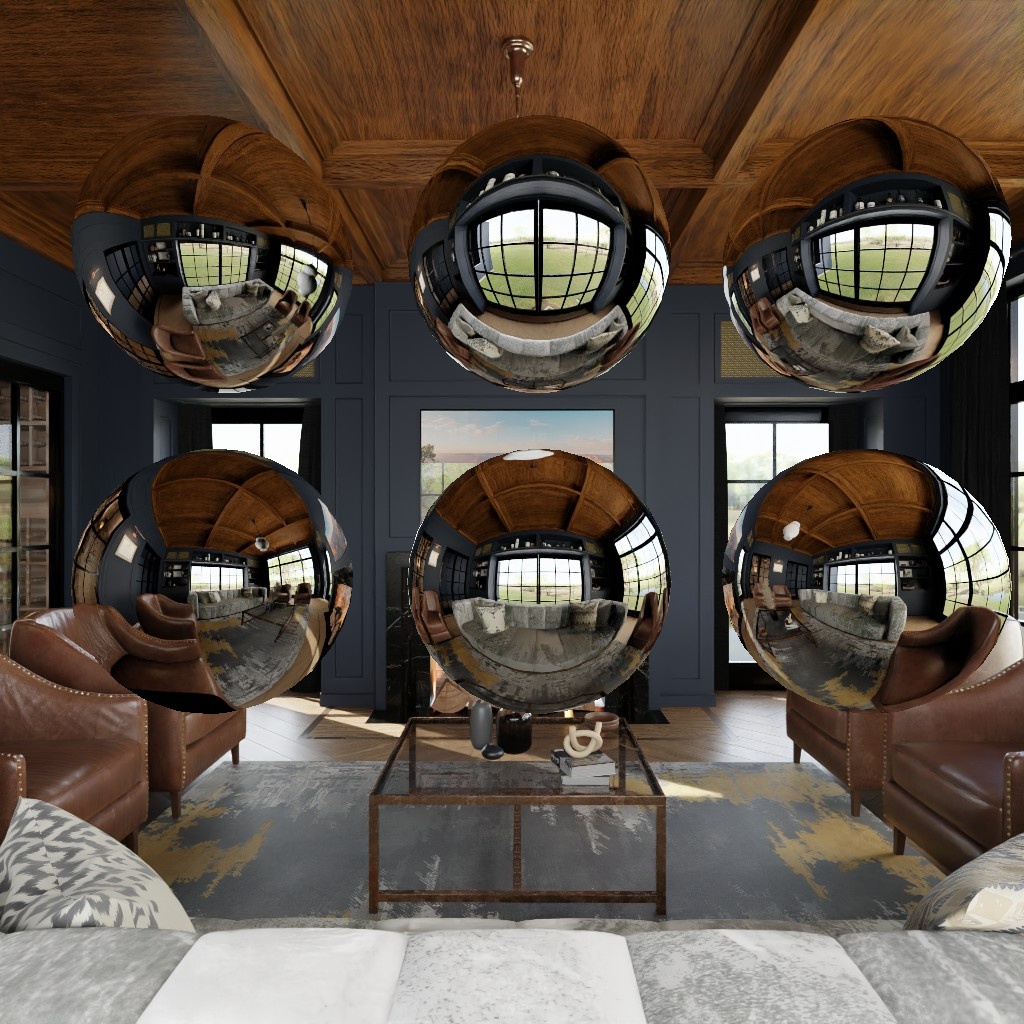} &
    \includegraphics[width=\mywidth]{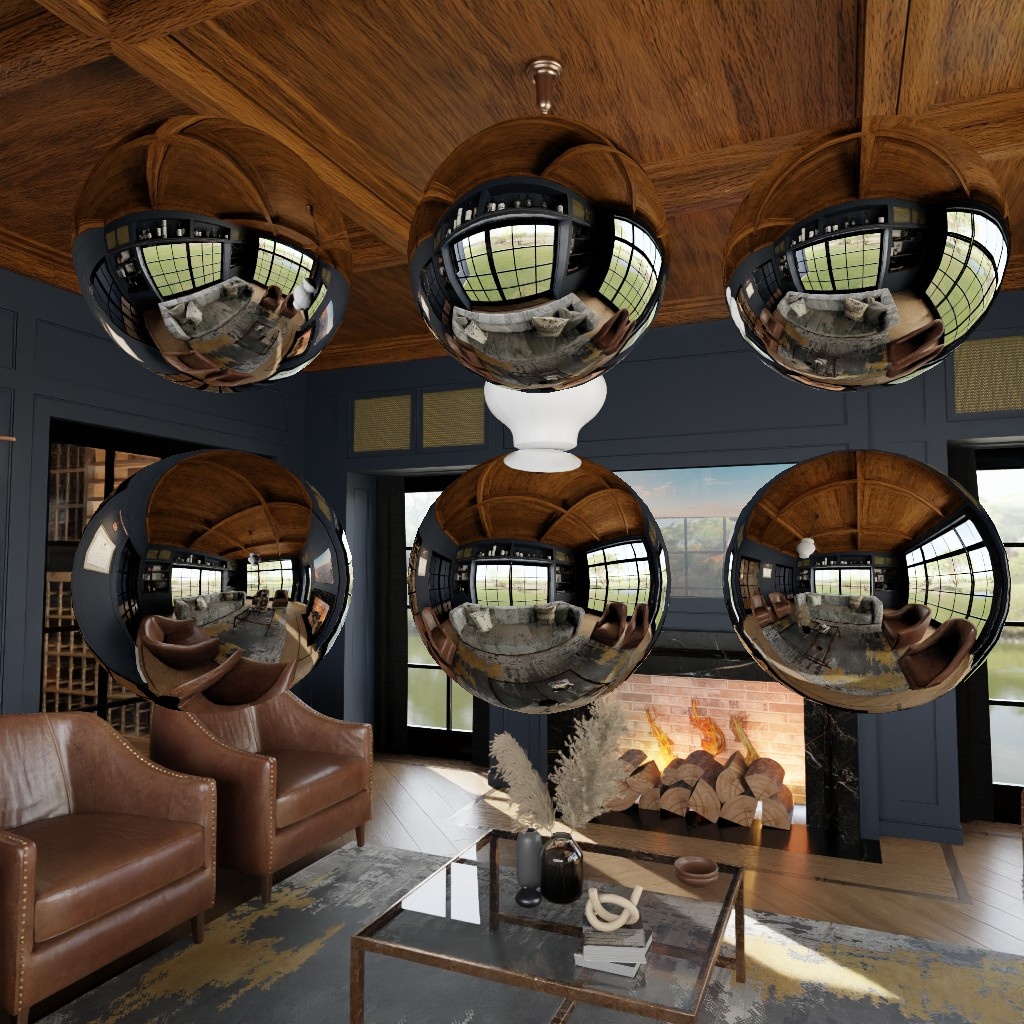} & \includegraphics[width=\mywidth]{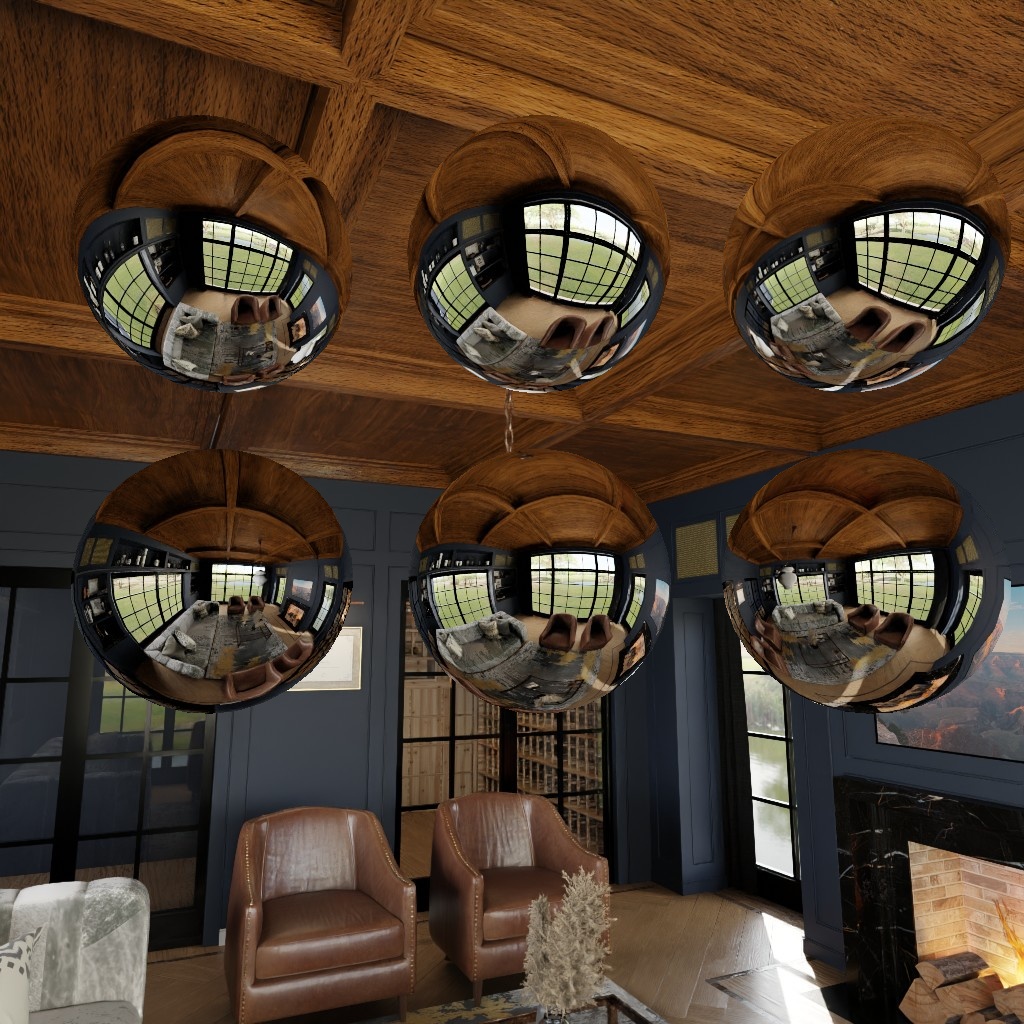}  \\*[-0.25em]
     \rot{\tiny \hspace{15pt}Dynamic lighting} &
    \includegraphics[width=\mywidth]{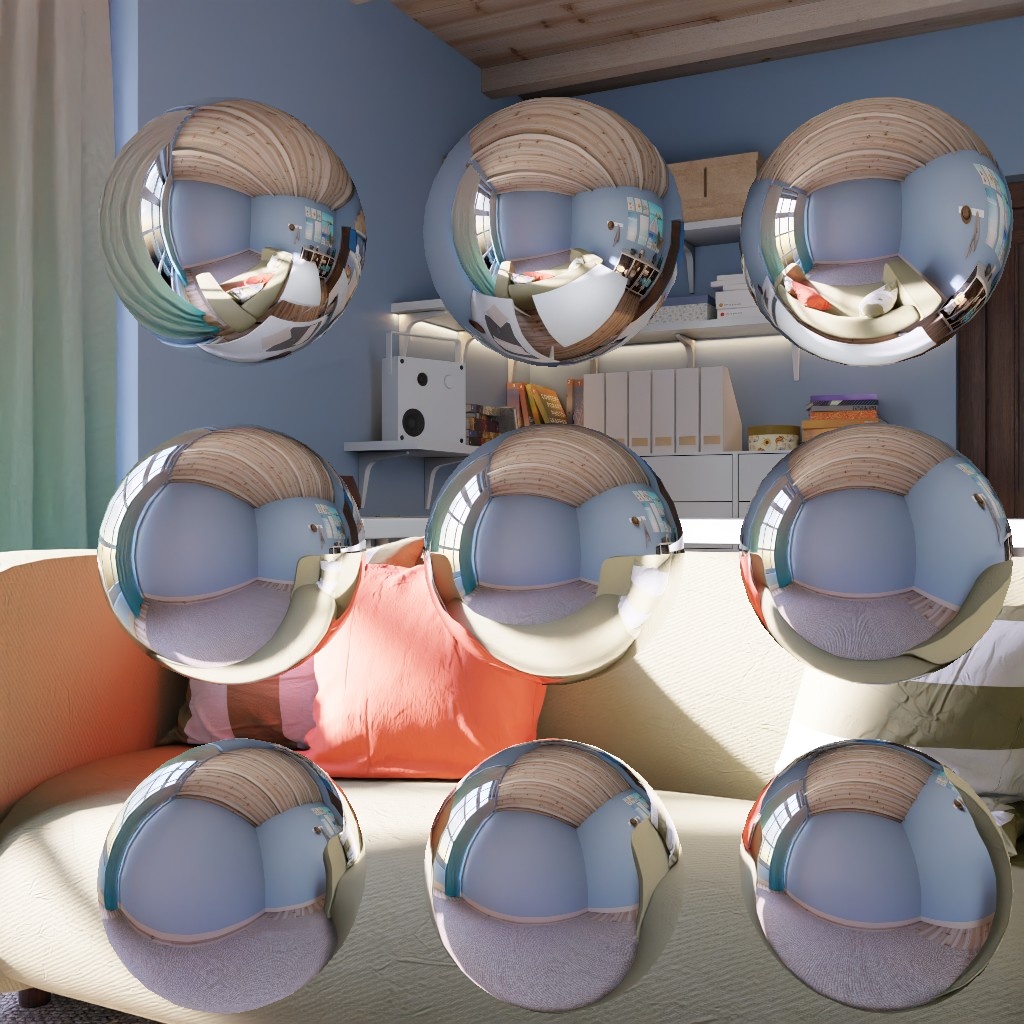} &
    \includegraphics[width=\mywidth]{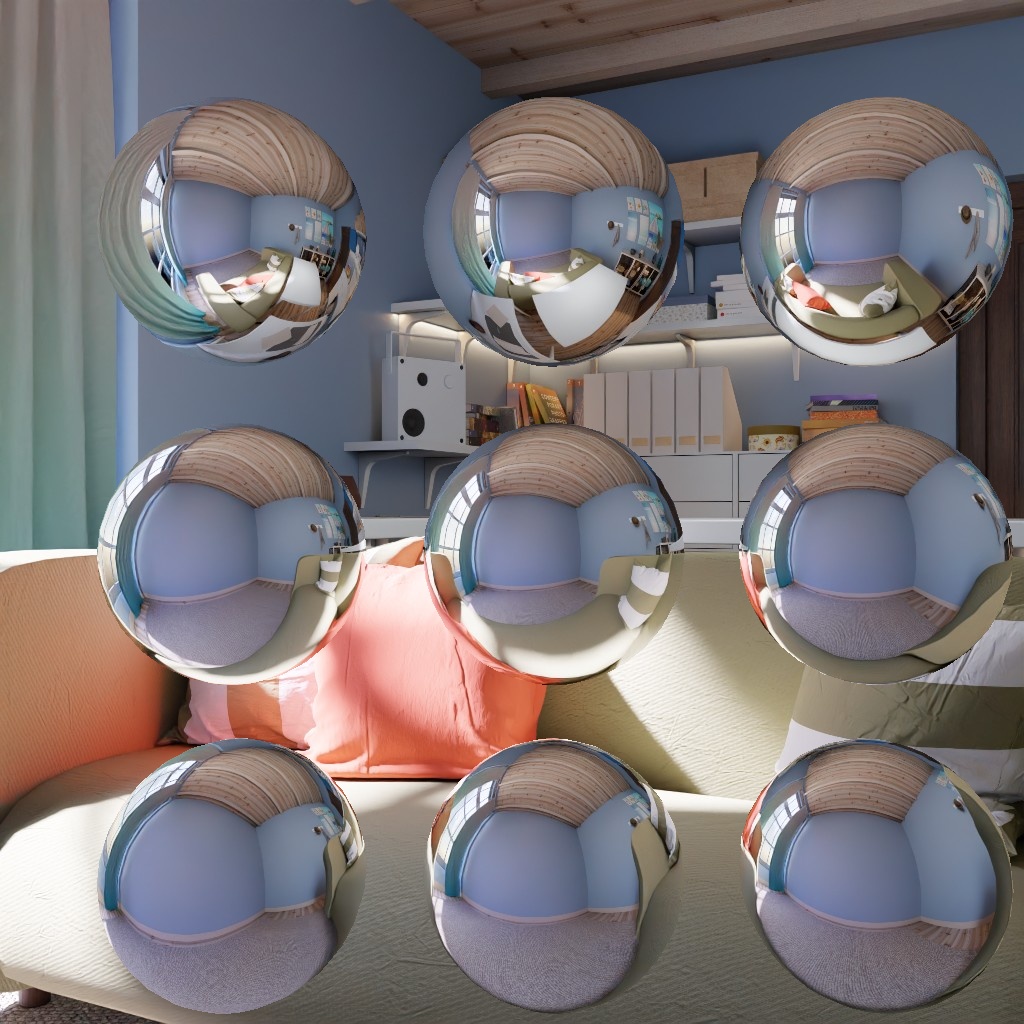} & \includegraphics[width=\mywidth]{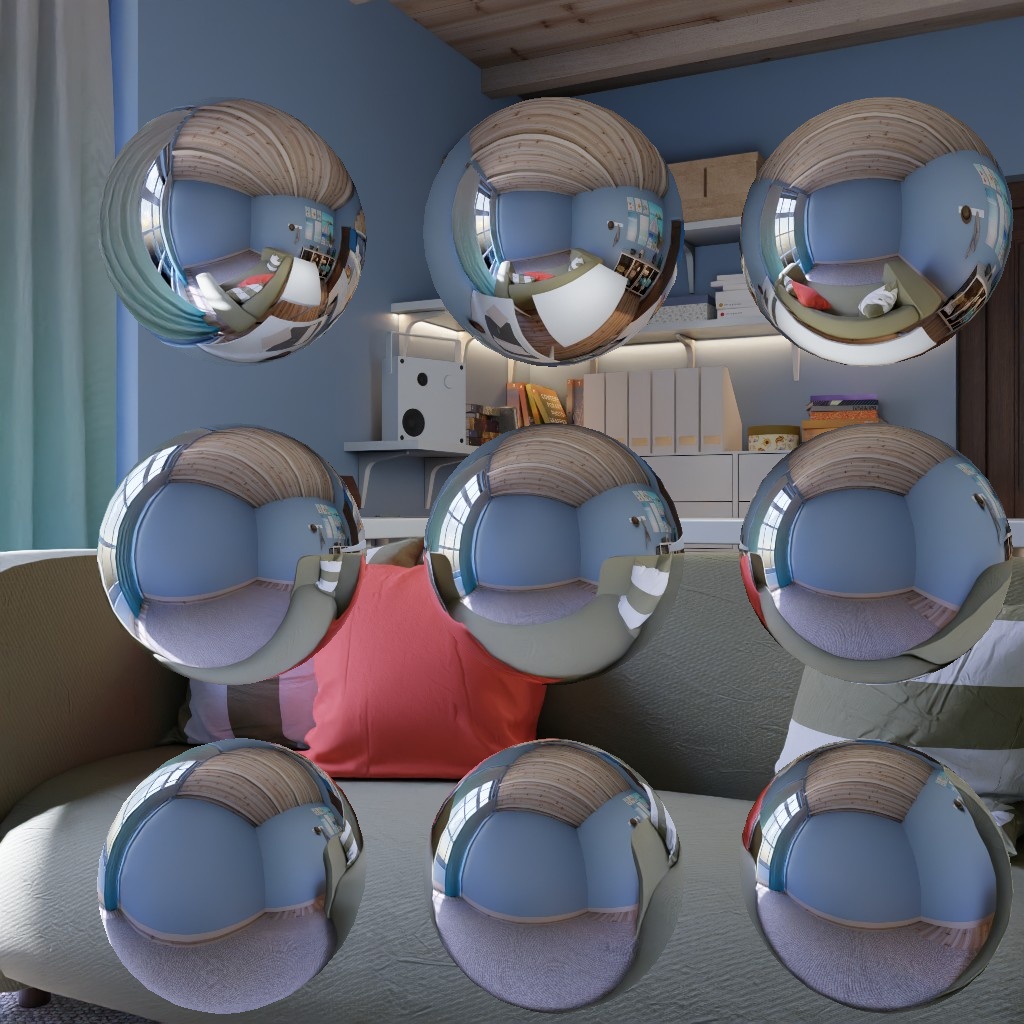} \\
    \end{tabular}
    \caption{Sample frames from the training dataset, illustrating the three animation scenarios. Note that diffuse spheres and empty scenes are also rendered (not shown). While multiple spheres are rendered simultaneously for efficiency, at training and inference the network regresses only one sphere at a time.}
    \label{fig:data_example}
\end{figure}

\section{HDRI map optimization details}
The predicted images from the network $\hat{I}$ are cropped around the inpainted spheres. The same is done with the sphere mask, normals and position maps. The equirectangular HDRI is a Laplacian pyramid at a fixed resolution of 512x256 with 8 levels. We employ circular padding to leverage the cyclic nature of equirectangular maps. For faster convergence and better conditioning, the HDRI is defined in $\log_2$ space. We optimize with Adam using a learning rate of $5\mathrm{e}{-3}$ for 1000 iterations per frame, for a total of 21 000 steps. 

At every step, we randomly select a frame $t$, sphere type $m$ (mirror or diffuse) and EV $e$ from the predictions. The Laplacian pyramid is recomposed and is transformed back to linear space to obtain the HDRI map $L_t$. Then, the renderer $\mathcal{R}$ is used to produce the image of corresponding sphere (mirror or diffuse). This rendered image is exposed according to the randomly selected EV and converted to sRGB color space to match the network's prediction's colors:
\begin{equation}
    \hat{I}_t = sRGB(2^{e}\mathcal{R}((L_t, m))).
\end{equation}

The loss function to optimize the HDRI representation is defined as:

\begin{equation}
\ell = M_{\text{sat}}(\ell_2(\hat{I}_t, I_t) + \frac{\lambda}{2} (\ell_1(\hat{I}_t, \hat{I}_{t-1}) + \ell_1(\hat{I}_t, \hat{I}_{t+1}))),
\end{equation}

with $\lambda=0.1$ in all our experiments. The $\ell_2$ loss enforces the rendered image to closely match the predicted image, and the two following $\ell_1$ losses insure that the rendered image be similar to the neighboring frames, allowing for temporal smoothing.
To prevent the saturated part of the image from lowering the overall intensity of the optimized HDRI, we define a saturation mask
\begin{equation}
    M_{\text{sat}} =
    \begin{cases}
    0, & \hat{I}_t > \tau \ \text{and}\ I_t > \tau, \\
    1, & \text{otherwise}.
    \end{cases}
\end{equation}

The renderer $\mathcal{R}$ is a two modes differentiable Monte Carlo renderer for perfect mirror and perfect diffuse materials. The perfect reflection is implementing the reflection equation
\begin{equation}
    \mathbf{v}_i - 2(\mathbf{v}_i \cdot \mathbf{n}_i)\,\mathbf{n}_i,
\end{equation}
with $\mathbf{v}_i$ computed from the sphere's position map.

For the diffuse rendering, we first compute the luminance of the HDRI to use as importance weight:
\begin{equation}
    L = 0.2126  R + 0.7152  G + 0.0722  B
\end{equation}

The importance map for each pixel of the HDRI map is computed as a multi-importance weighting of cosine and luminance:
\begin{equation}
    w_i = (n_i \cdot r_i)\, L_i \, sin(r_i),
\end{equation}
where $r_i$ is the ray direction corresponding to pixel $i$ of the HDRI map. It is then normalized:
\begin{equation}
\mathbf{w} = \frac{w}{\sum_i w_i}.
\end{equation}

The corresponding probability distribution function is computed by dividing the normalized importance map by the solid angle of the equirectangular map 
\begin{equation}
    PDF = \frac{\mathbf{w}}{\partial{\omega}}.
\end{equation}
Samples $s$ are drawn from the importance map $\mathbf{w}$ and the final rendered colors $R_i$ is
\begin{equation}
    R_i = \frac{1}{S}\sum_{s \in S}{\frac{L_s (n_i \cdot r_i)}{PDF}}.
\end{equation}

We use 64 samples with sub-pixel sampling in all our experiments.

\section{Ablation of field of view stability}
\label{fov_ablation}
\begin{figure}[t]
    \centering
    \includegraphics[width=\linewidth]{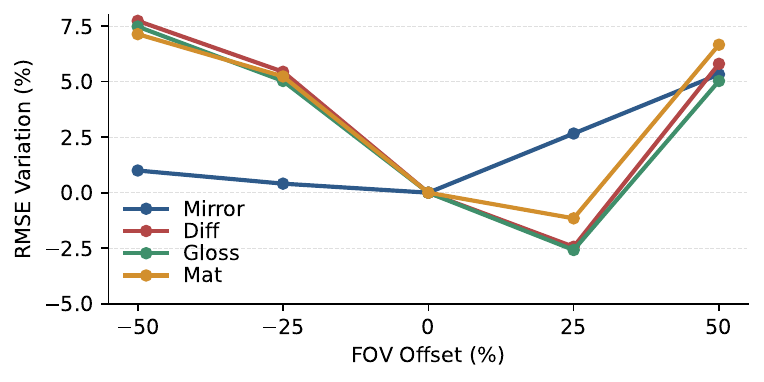}
    \caption{\change{Effect of varying field of view on RMSE on the Infinigen dataset.}}
    \label{fig:fov_ablation}
\end{figure}

\change{As \themethod assumes known FOV, which is estimated from an off-the-shelf network in practice~\cite{wang2025moge2}, we ablate the effect of perturbing the estimation in \cref{fig:fov_ablation}. We observe that \themethod is quite robust to varying FOVs. Interestingly, we observe a slight improvement (2\%) in RMSE when the FOV is over-estimated by 25\%. When the FOV is severely over- or under-estimated (by up to 50\%), the performance hit is at most 8\%.}

\change{\section{Ablation on Laval Indoor SV}
Ablations of our modules for the Laval Indoor SV Dataset~\cite{garon2019fast} are presented in \cref{tab:ablation_laval}. Interestingly, while still improving colors (Angular error), the use of the diffuse prediction for the HDRi optimization results in less accurate overall intensity (RMSE). We hypothesize that this is because we predict spheres larger than those in the ground truth in order to increase the prediction resolution, biasing the prediction.}

\section{Additional results}
\begin{table*}
\centering
\footnotesize
\setlength{\tabcolsep}{1pt}
\begin{tabular}{llcccccccccccccc}
\midrule
& & \multicolumn{2}{c}{RMSE$_\downarrow$}  &  \multicolumn{2}{c}{SI-RMSE$_\downarrow$}  & \multicolumn{2}{c}{SSIM$_\uparrow$}  &   \multicolumn{2}{c}{Ang. Err.$_\downarrow$}  &  \multicolumn{2}{c}{T-LPIPS$_\downarrow$}  &  \multicolumn{2}{c}{T-LPIPS-Diff$_\downarrow$}  &  \multicolumn{2}{c}{Warped Err$_\downarrow$} \\
\midrule
 Dataset & Method   & Gloss & Mat   &   Gloss & Mat   & Gloss & Mat  & Gloss & Mat  & Gloss & Mat  & Gloss & Mat  & Gloss & Mat  \\
 \midrule
 \multirow{4}{*}{\shortstack[l]{Dynamic\\object}} 
 & 4D Lighting   & \num{0.30} & \num{0.33}  & \num{0.21}   & \num{0.34} & \num{0.89}   & \num{0.88}  & \num{4.6}  & \num{4.9} & \best{\num{0.0009}}  & \best{\num{0.0006}} &  \second{\num{0.0064}}  & \best{\num{0.0016}} &  \best{\num{0.0125}}  & \best{\num{0.0099}} \\
 & \themethod (image) & \best{\num{0.15}}  & \best{\num{0.16}} & \best{\num{0.12}}   & \best{\num{0.17}}  & \best{\num{0.96}}   & \best{\num{0.97}}  & \best{\num{1.4}}  & \best{\num{1.7}} & \num{0.0224}  & \num{0.0138}  &  \num{0.0166}  & \num{0.0117} &  \num{0.0489}  & \num{0.0575} \\
 & \themethod (video)  & \second{\num{0.18}}  & \second{\num{0.21}} & \second{\num{0.13}}   & \second{\num{0.21}} & \best{\num{0.96}}   & \second{\num{0.96}}  & \second{\num{2.9}}  & \second{\num{3.0}} & \second{\num{0.0025}}  & \second{\num{0.0020}}  &  \best{\num{0.0053}}  & \second{\num{0.0017}} &  \second{\num{0.0200}}  & \second{\num{0.0156}} \\
 \midrule
  \multirow{4}{*}{\shortstack[l]{Dynamic\\camera}} 
 & 4D Lighting   & \num{0.38}  & \num{0.37}  & \num{0.18}   & \num{0.31} & \num{0.89}   & \num{0.88}  & \num{3.8}  & \num{4.3}  & \best{\num{0.0011}}  & \best{\num{0.0010}} &  \second{\num{0.0047}}  & \best{\num{0.0007}}  &  \best{\num{0.0163}}  & \second{\num{0.0142}}\\
 & \themethod (image) & \best{\num{0.16}}  & \best{\num{0.17}} & \best{\num{0.12}}   & \best{\num{0.18}}  & \best{\num{0.96}}   & \best{\num{0.97}}  & \best{\num{1.4}}  & \best{\num{1.8}}  & \num{0.0179}  & \num{0.0114} &  \num{0.0125}  & \num{0.0102} &  \num{0.0499}  & \num{0.0541} \\
 & \themethod (video)  & \second{\num{0.21}}  & \second{\num{0.23}} & \second{\num{0.13}}   & \second{\num{0.21}} & \best{\num{0.96}}   & \second{\num{0.96}}  & \second{\num{2.4}}  & \second{\num{3.0}} & \second{\num{0.0019}}  & \second{\num{0.0015}} &  \best{\num{0.0042}}  & \second{\num{0.0012}}  &  \second{\num{0.0178}}  & \best{\num{0.0150}} \\
 \midrule
  \multirow{4}{*}{\shortstack[l]{Dynamic\\lighting}}
 & 4D Lighting  & \num{0.34}  & \num{0.35} & \num{0.70}   & \num{0.80}  & \num{0.86}   & \num{0.85}  & \num{10.1}  & \num{10.1} & \best{\num{0.0008}}  & \best{\num{0.0007}}  &  \second{\num{0.0011}}  & \best{\num{0.0005}}  &  \second{\num{0.0095}}  & \second{\num{0.0096}} \\
 & \themethod (image) & \best{\num{0.17}}  & \best{\num{0.18}} & \best{\num{0.13}}   & \best{\num{0.19}}  & \best{\num{0.95}}   & \best{\num{0.96}}  & \best{\num{1.8}}  & \best{\num{2.2}} & \num{0.0032}  & \num{0.0022}  &  \num{0.0018}  & \num{0.0015} &  \num{0.0203}  & \num{0.0217} \\
 & \themethod (video)  & \second{\num{0.22}}  & \second{\num{0.25}} & \second{\num{0.15}}   & \second{\num{0.24}} & \second{\num{0.94}}   & \second{\num{0.95}}  & \second{\num{2.8}}  & \second{\num{3.2}} & \second{\num{0.0009}}  & \second{\num{0.0008}}  &  \best{\num{0.0010}}  & \best{\num{0.0005}} &  \best{\num{0.0077}}  & \best{\num{0.0077}} \\
 \midrule
  \multirow{4}{*}{Combination} 
 & 4D Lighting    & \num{0.32}  & \num{0.33}  & \num{0.20}   & \num{0.33} & \num{0.89}   & \num{0.88}  & \num{4.0}  & \num{4.3} & \best{\num{0.0017}}  & \best{\num{0.0013}}  &  \second{\num{0.0103}}  & \best{\num{0.0018}}  &  \best{\num{0.0217}}  & \best{\num{0.0167}}\\
 & \themethod (image) & \best{\num{0.16}}  & \best{\num{0.18}} & \best{\num{0.12}}   & \best{\num{0.19}}  & \best{\num{0.96}}   & \best{\num{0.97}}  & \best{\num{1.8}}  & \best{\num{2.2}} & \num{0.0249}  & \num{0.0170}  &  \num{0.0137}  & \num{0.0140}  &  \num{0.0615}  & \num{0.0723}\\
 & \themethod (video)  & \second{\num{0.23}}  & \second{\num{0.24}} & \second{\num{0.16}}   & \second{\num{0.24}} & \second{\num{0.94}}   & \second{\num{0.94}}  & \second{\num{2.7}}  & \second{\num{3.0}} & \second{\num{0.0048}}  & \second{\num{0.0042}}  &  \best{\num{0.0072}}  & \second{\num{0.0033}}  &  \second{\num{0.0337}}  & \second{\num{0.0289}} \\
\end{tabular}
\caption{Quantitative evaluation of lighting estimation on dynamic scenes for ``Gloss'' (glossy) and ``Mat'' (matte) spheres in complement to \cref{tab:quant_temporal_img}. We compare \themethod with ``4D Lighting''~\cite{4DLighting}. Results are color coded by \best{best}, \second{second} best.}
\label{tab:quant_temporal_img_plus}
\end{table*}
In complement to \cref{tab:quant_temporal_img}, \cref{tab:quant_temporal_img_plus} reports metrics on our sequences test dataset for glossy and matte spheres.

\begin{figure*}[!th]
   \centering
   \footnotesize
   \setlength{\tabcolsep}{0.5pt}
   \setlength{\tmplength}{0.11\linewidth}
   \setlength{\cbarheight}{1.9cm}
    \begin{tabular}{ccccccc}
    Scene & & DiffusionLight & 4D Lighting & \themethod (image) & \themethod (video) & GT \\
    
    \multirow[t]{2}{*}{\raisebox{-0.5\height}{\includegraphics[width=3\tmplength]{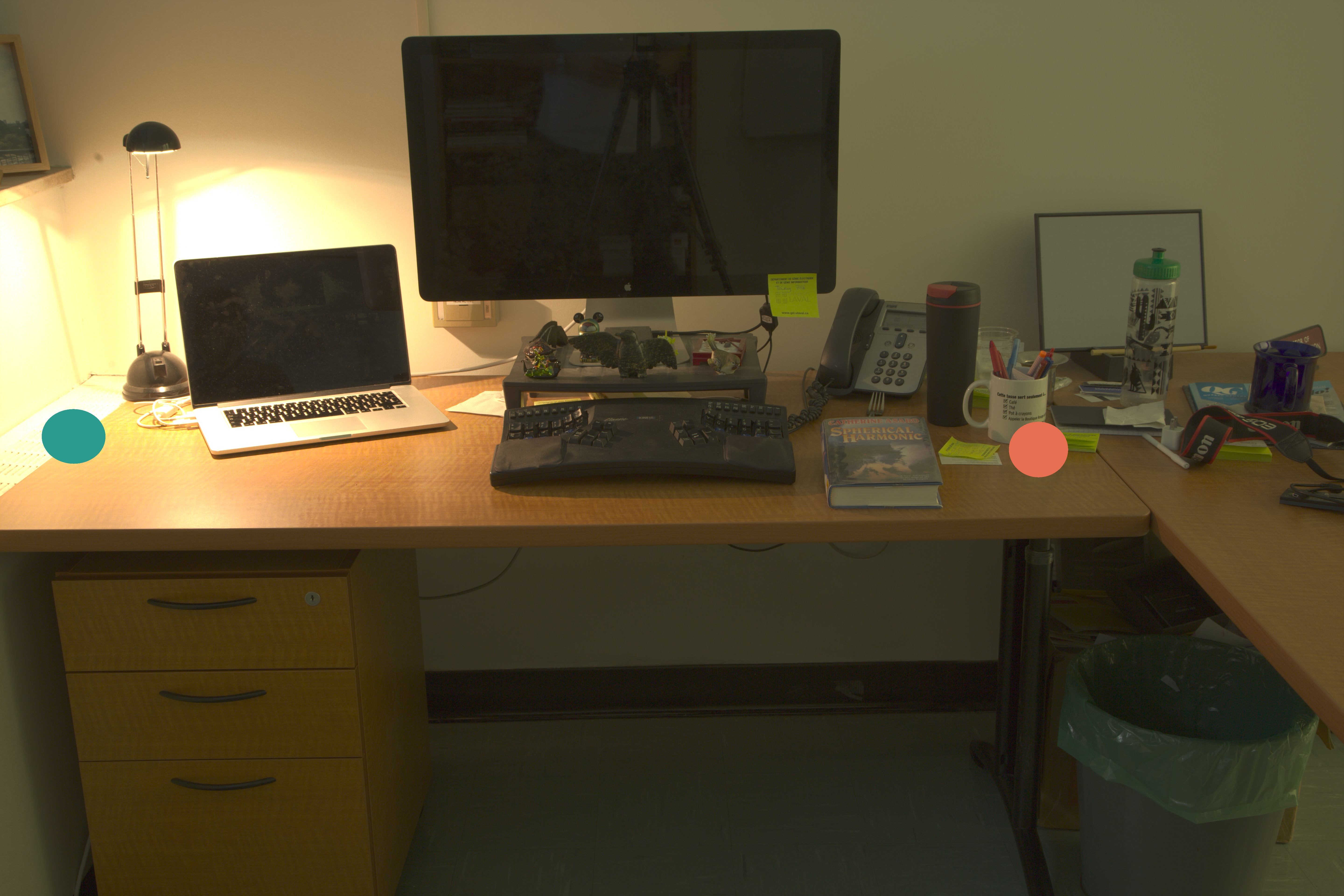}}}&
         \color{color2}\rule[4pt]{1.2pt}{1.7cm} 
    &
    \includegraphics[width=\tmplength]{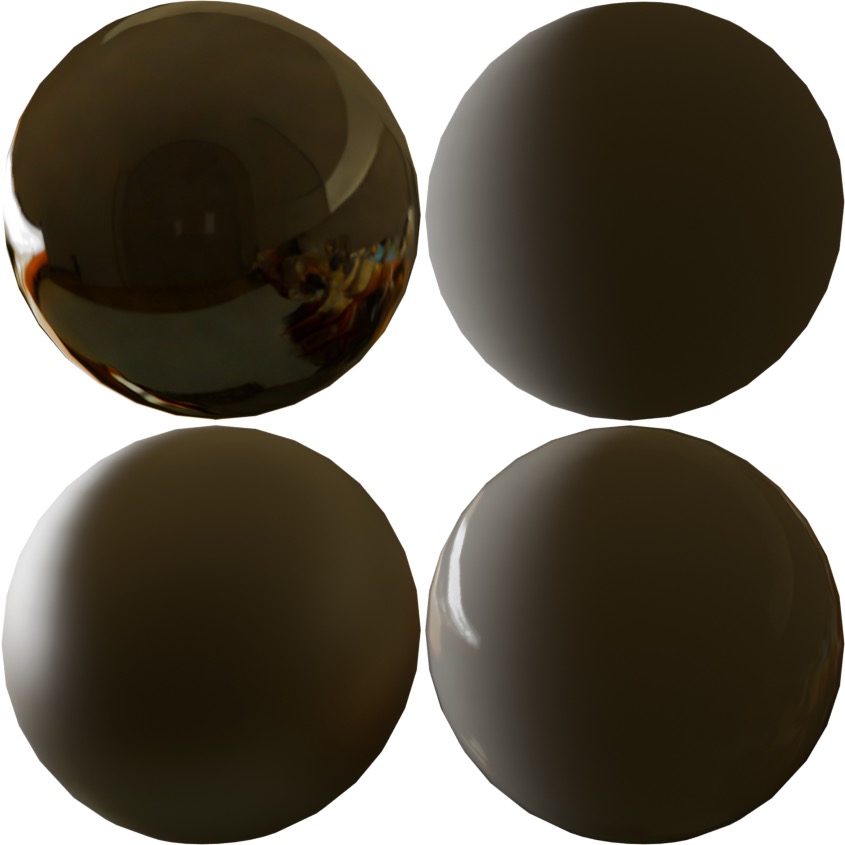}&
    \includegraphics[width=\tmplength]{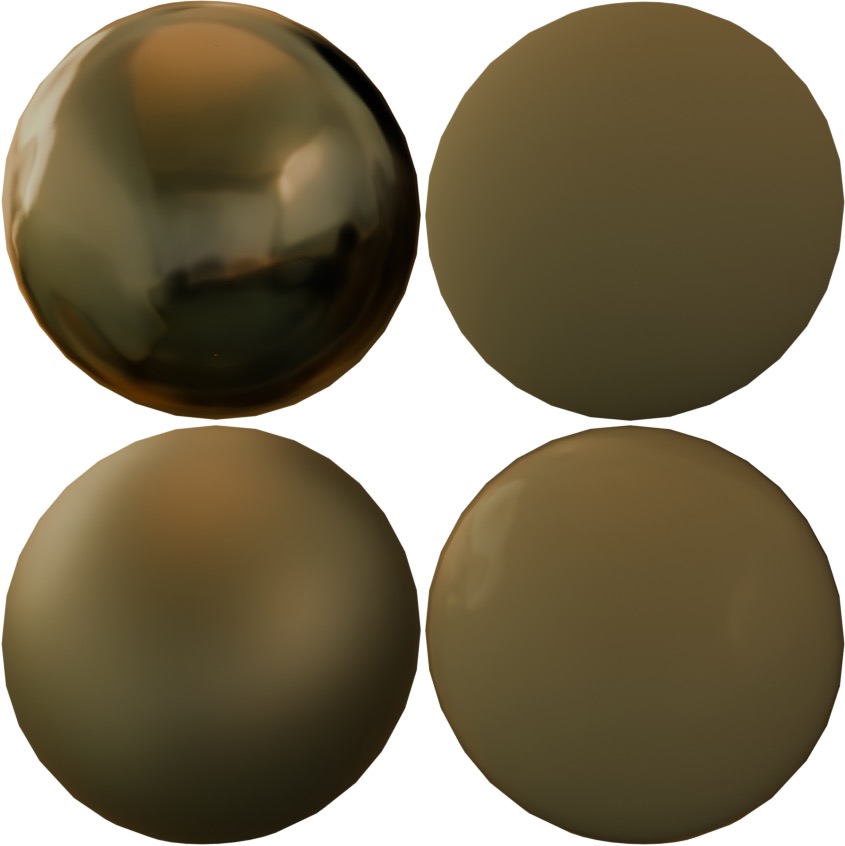}&
    \includegraphics[width=\tmplength]{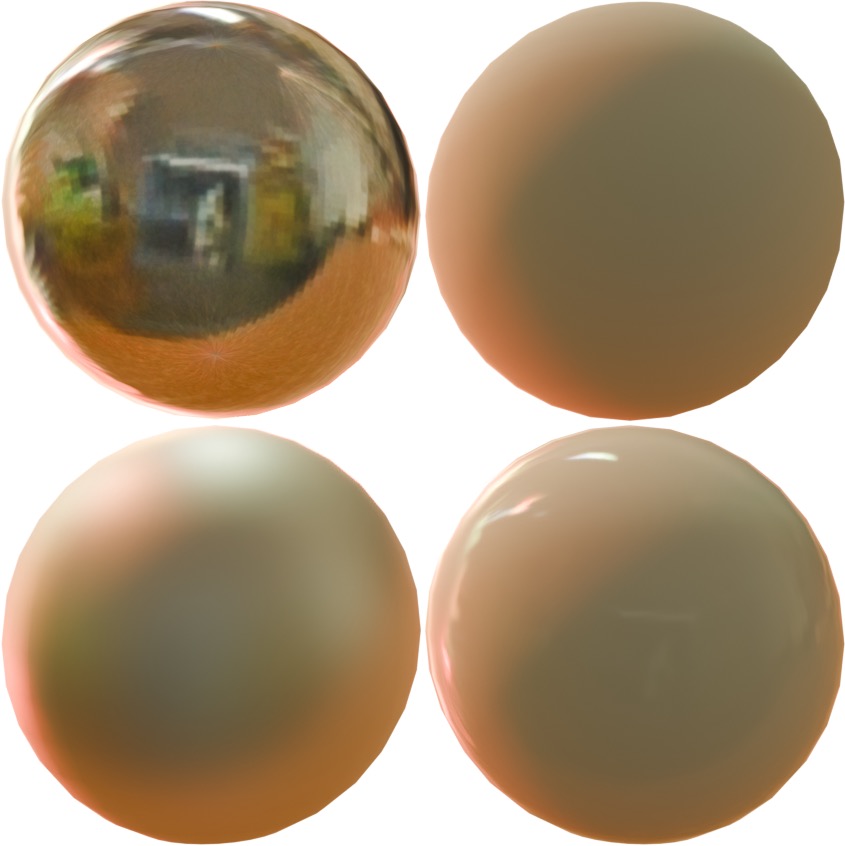}&
    \includegraphics[width=\tmplength]{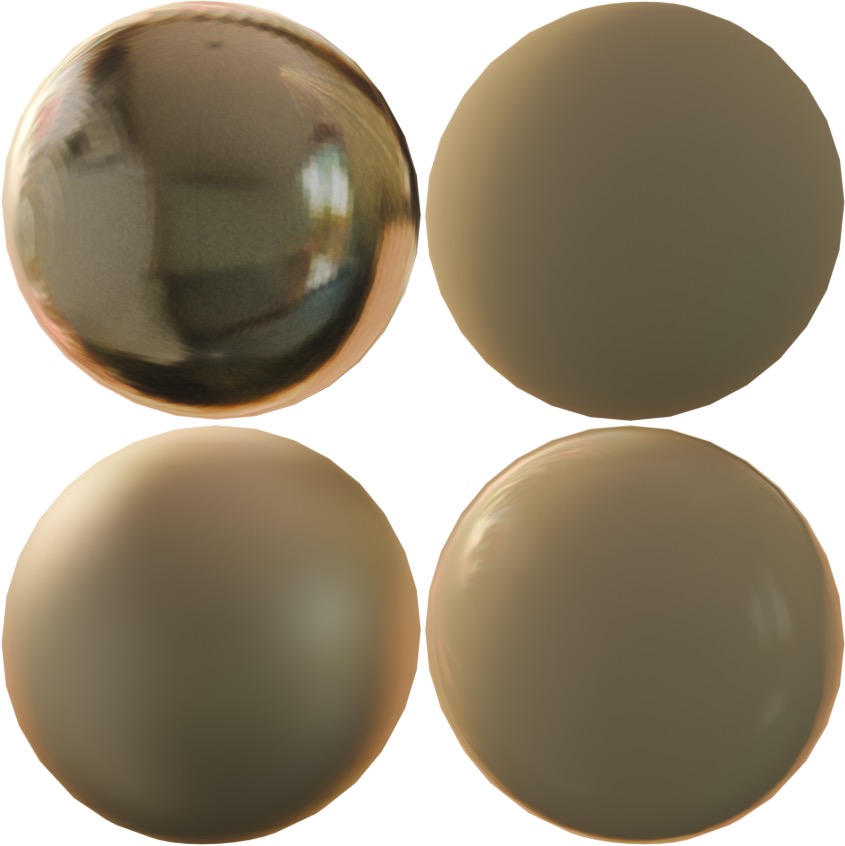}&
    \includegraphics[width=\tmplength]{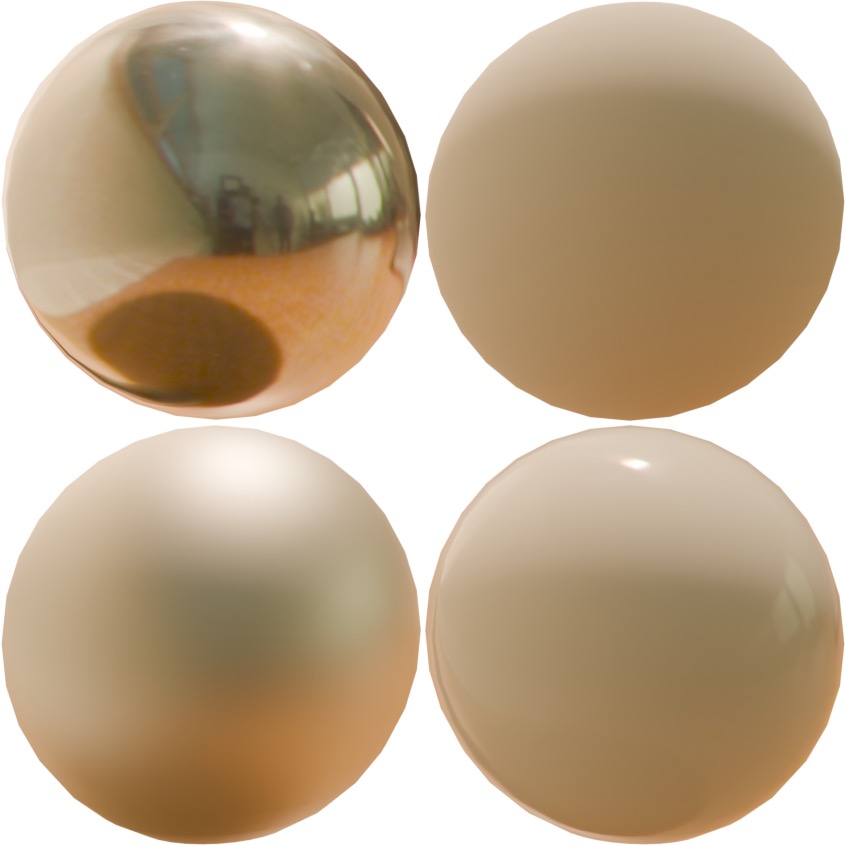}\\
    &
         \color{color5}\rule[4pt]{1.2pt}{1.7cm}
    &
    \includegraphics[width=\tmplength]{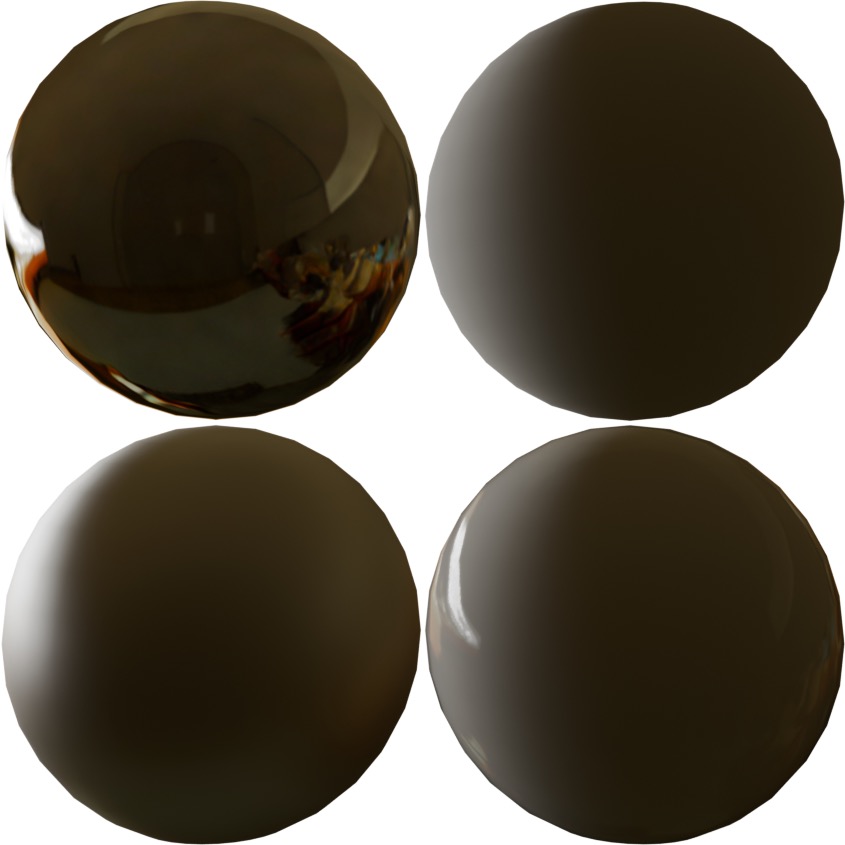}&
    \includegraphics[width=\tmplength]{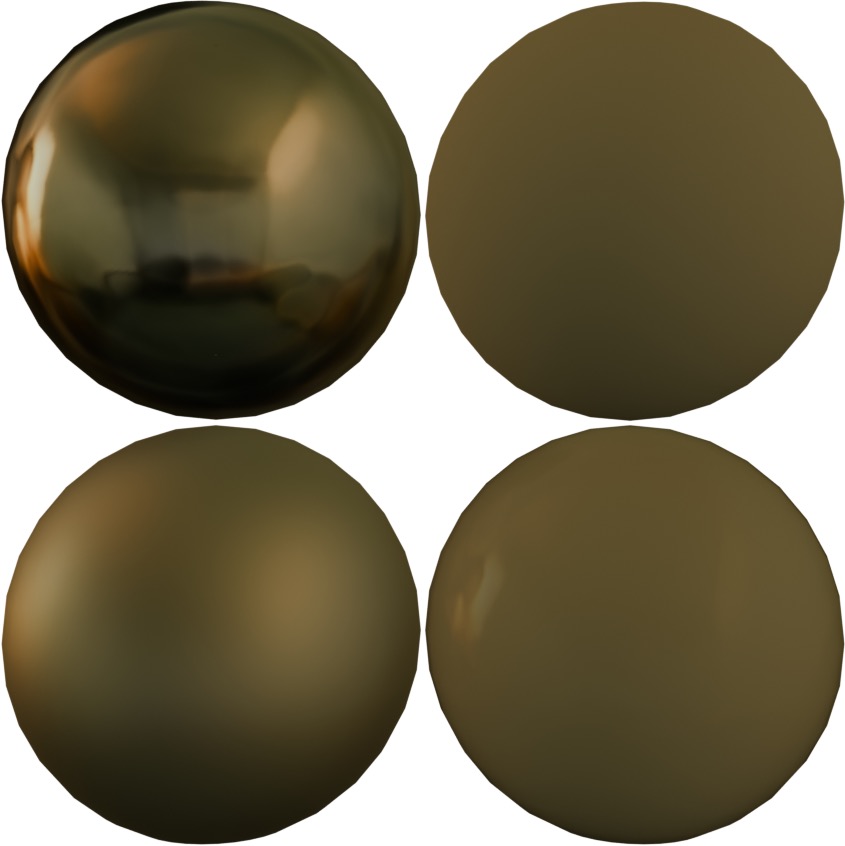}&
    \includegraphics[width=\tmplength]{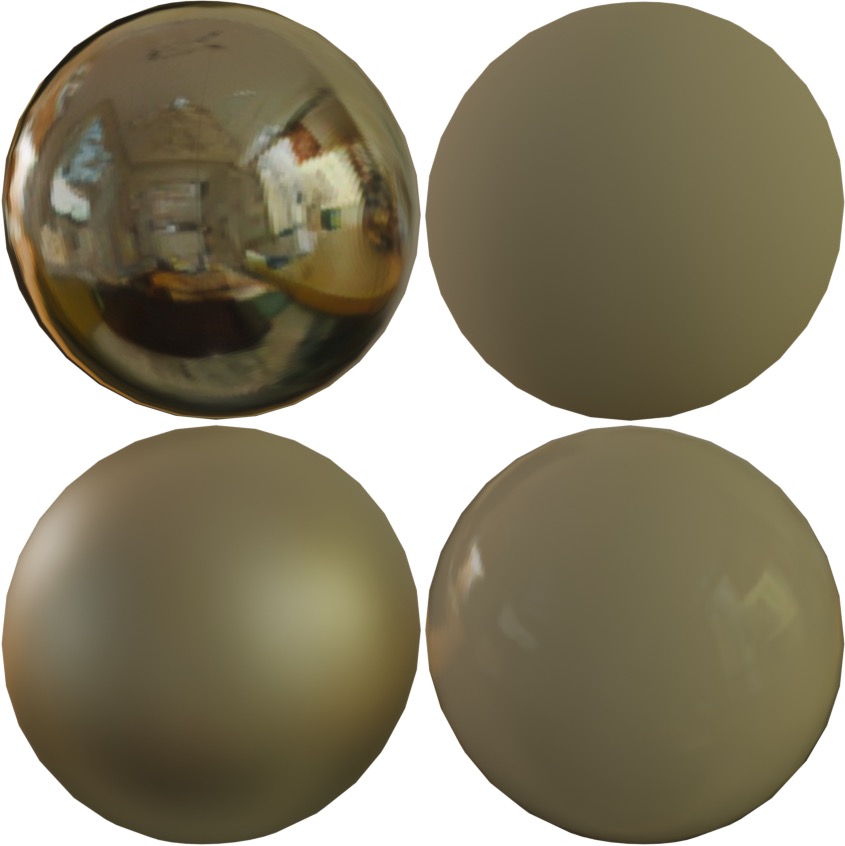}&
    \includegraphics[width=\tmplength]{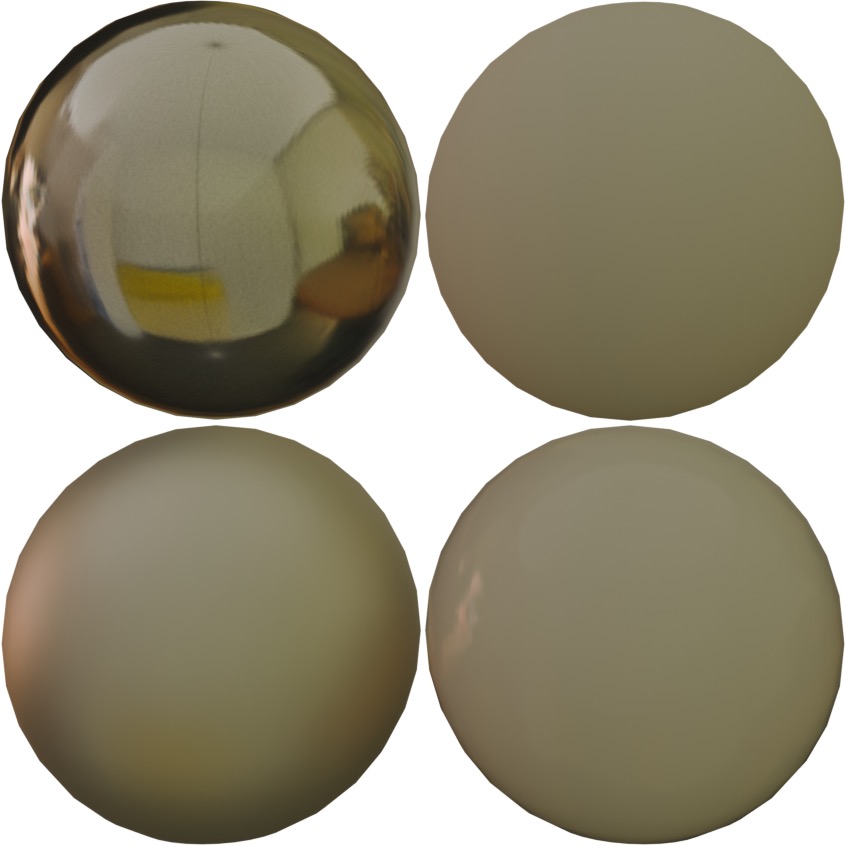}&
    \includegraphics[width=\tmplength]{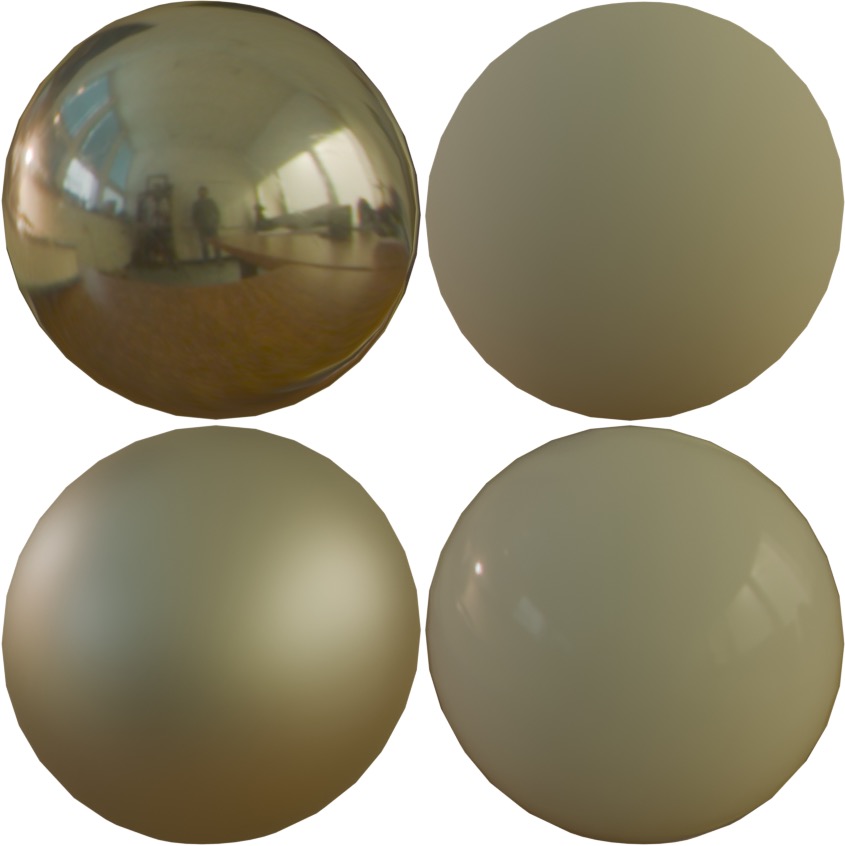}\\
    \multirow[t]{2}{*}{\raisebox{-0.5\height}{\includegraphics[width=3\tmplength]{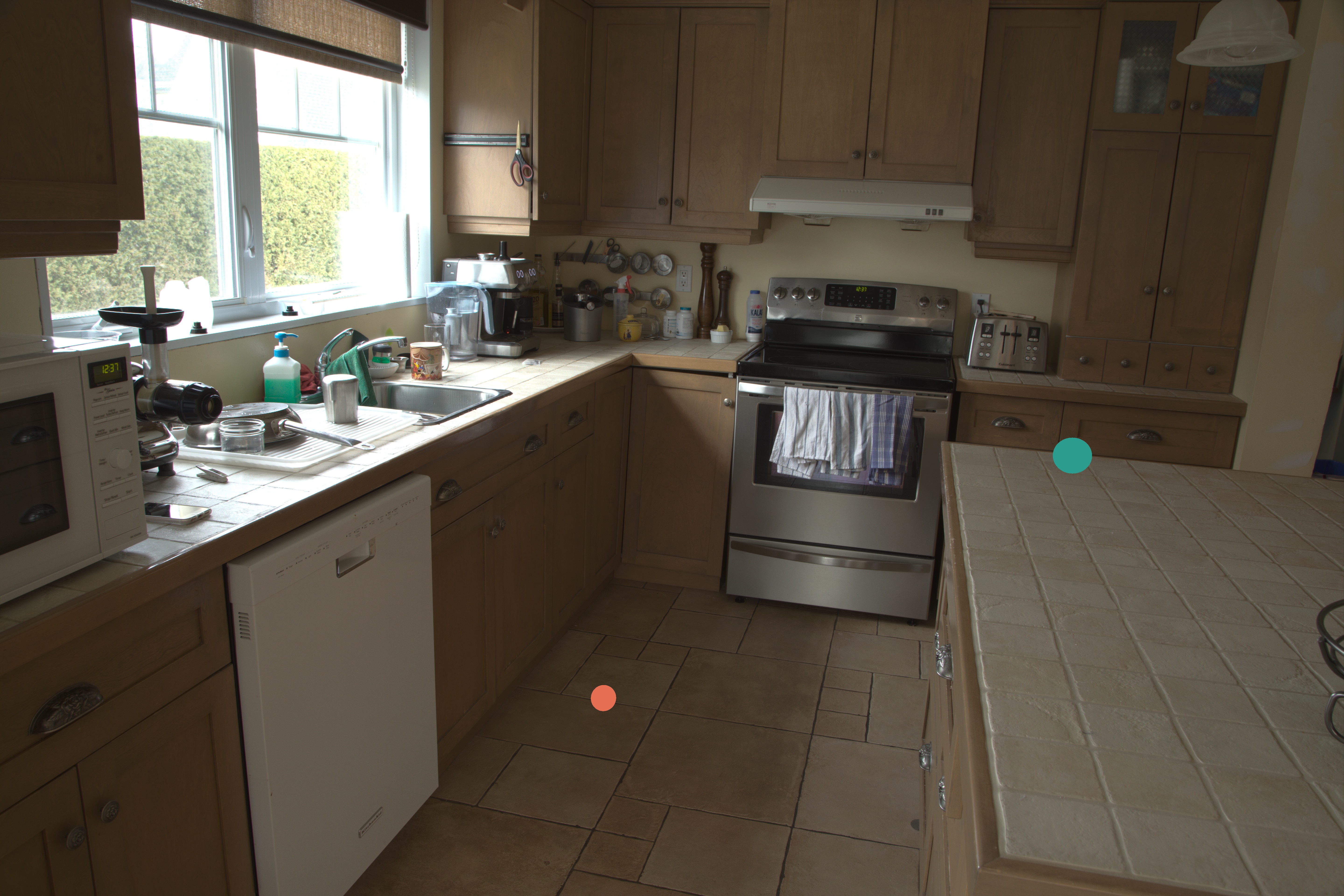}}}&
         \color{color2}\rule[4pt]{1.2pt}{1.7cm} 
    &
    \includegraphics[width=\tmplength]{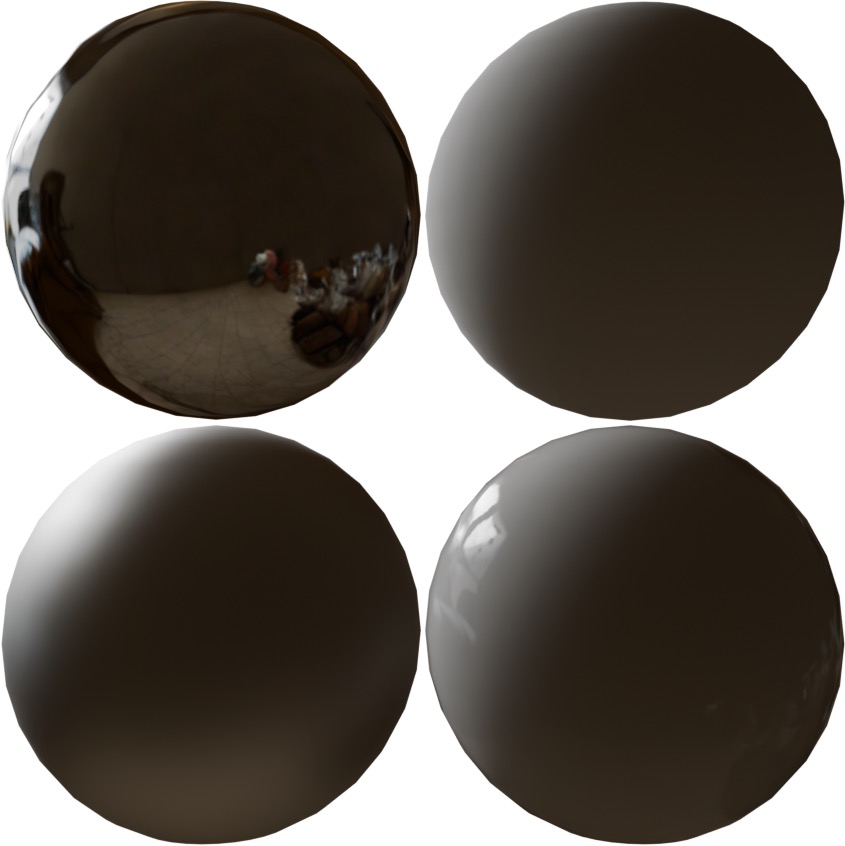}&
    \includegraphics[width=\tmplength]{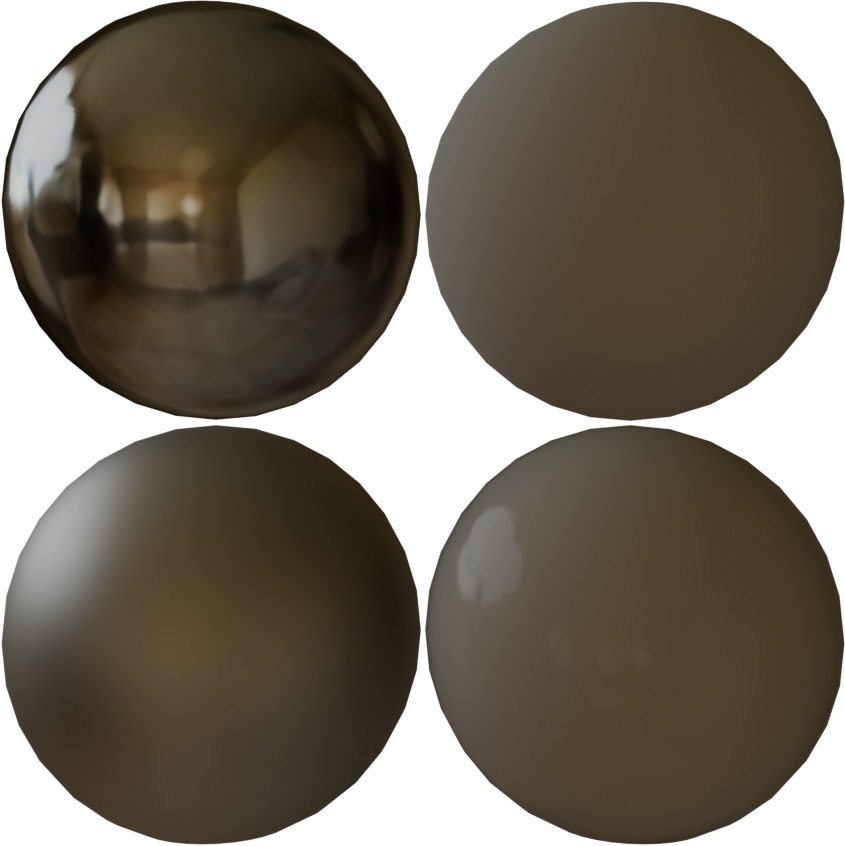}&
    \includegraphics[width=\tmplength]{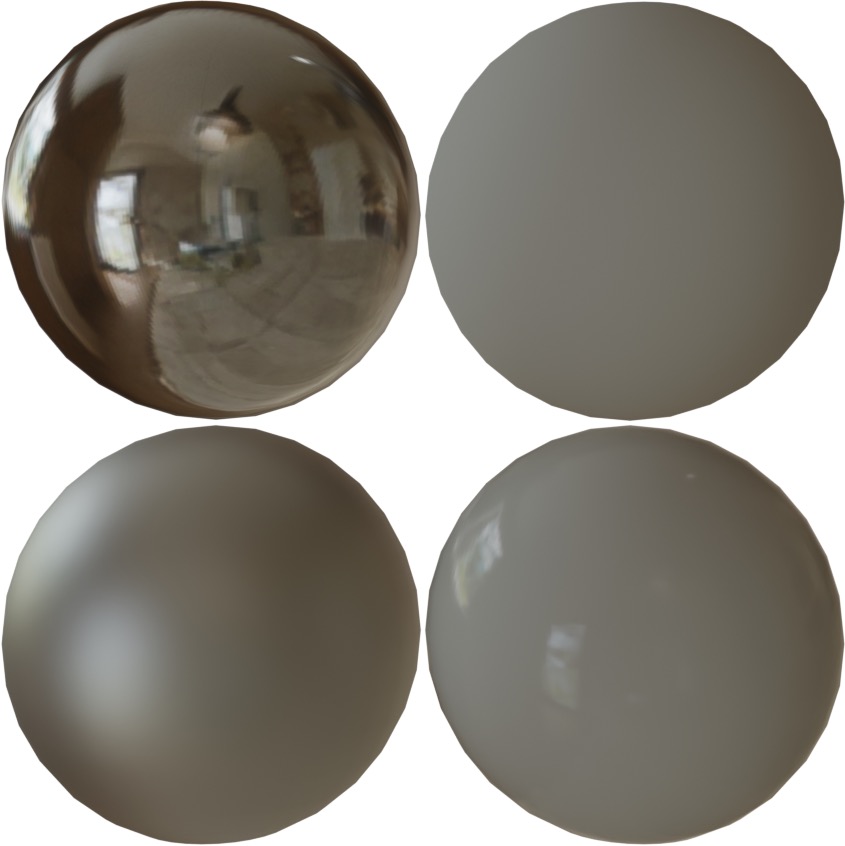}&
    \includegraphics[width=\tmplength]{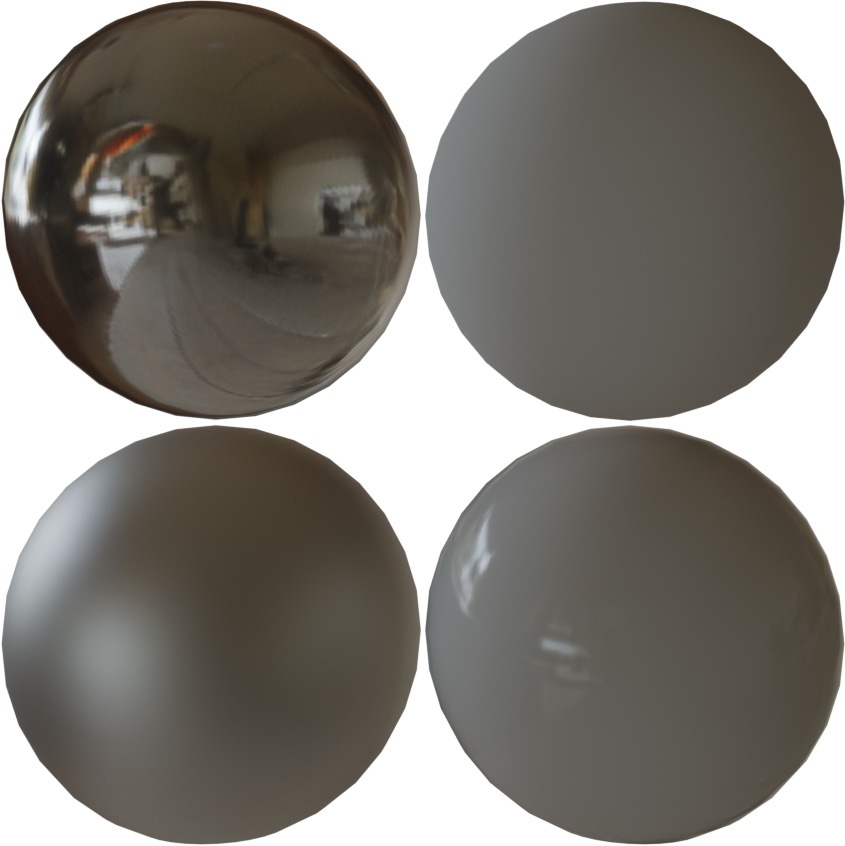}&
    \includegraphics[width=\tmplength]{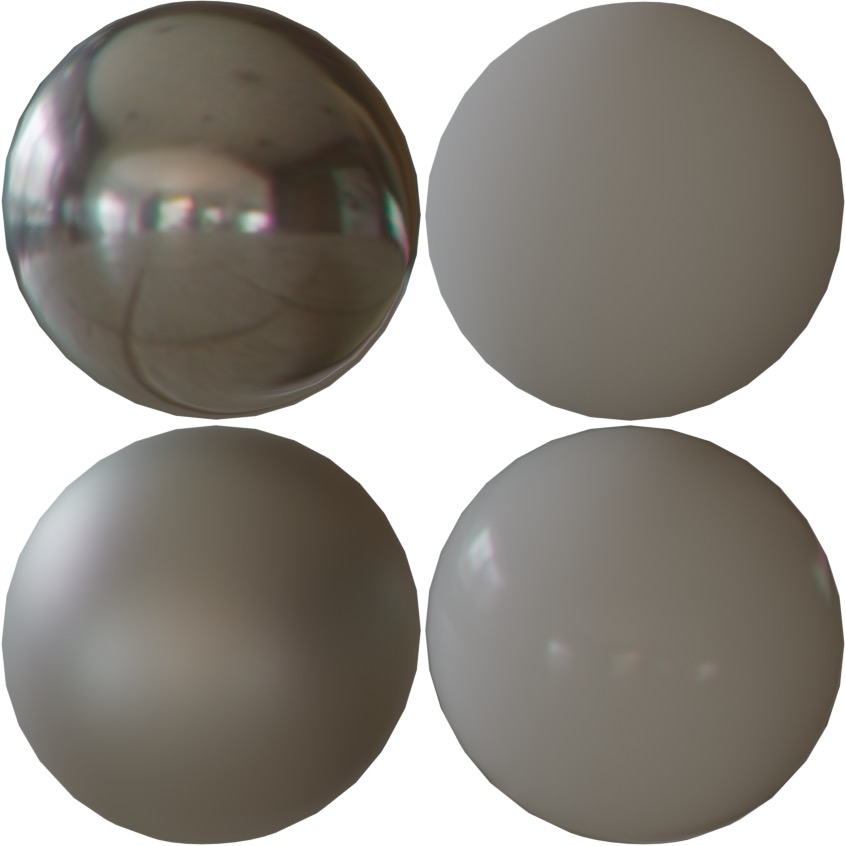}\\
    &
         \color{color5}\rule[4pt]{1.2pt}{1.7cm}
    &
    \includegraphics[width=\tmplength]{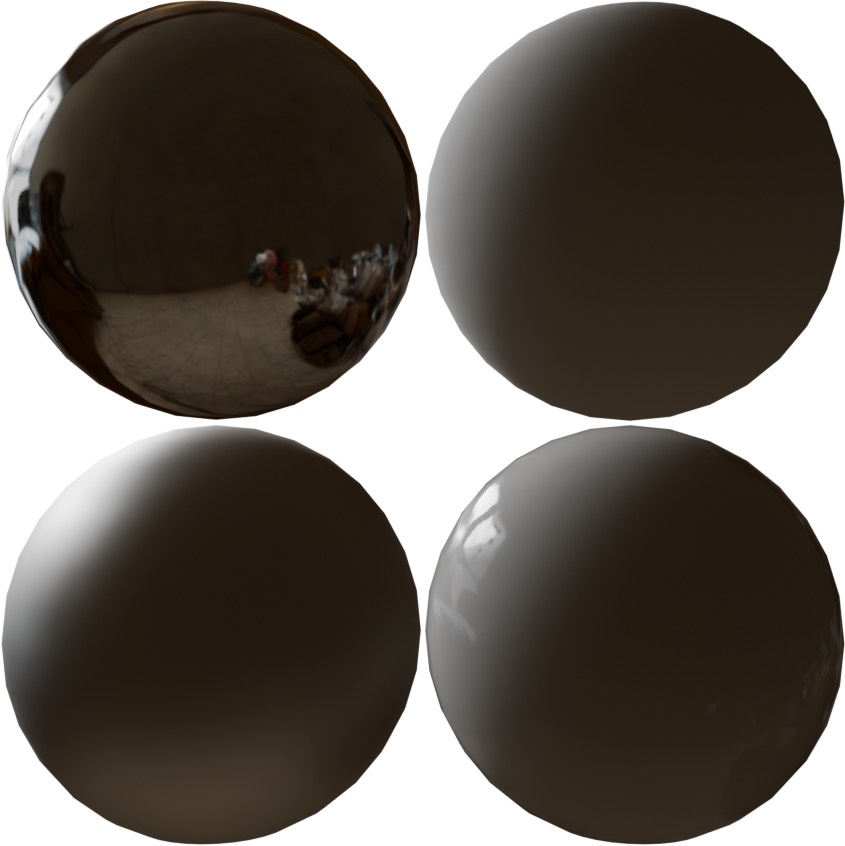}&
    \includegraphics[width=\tmplength]{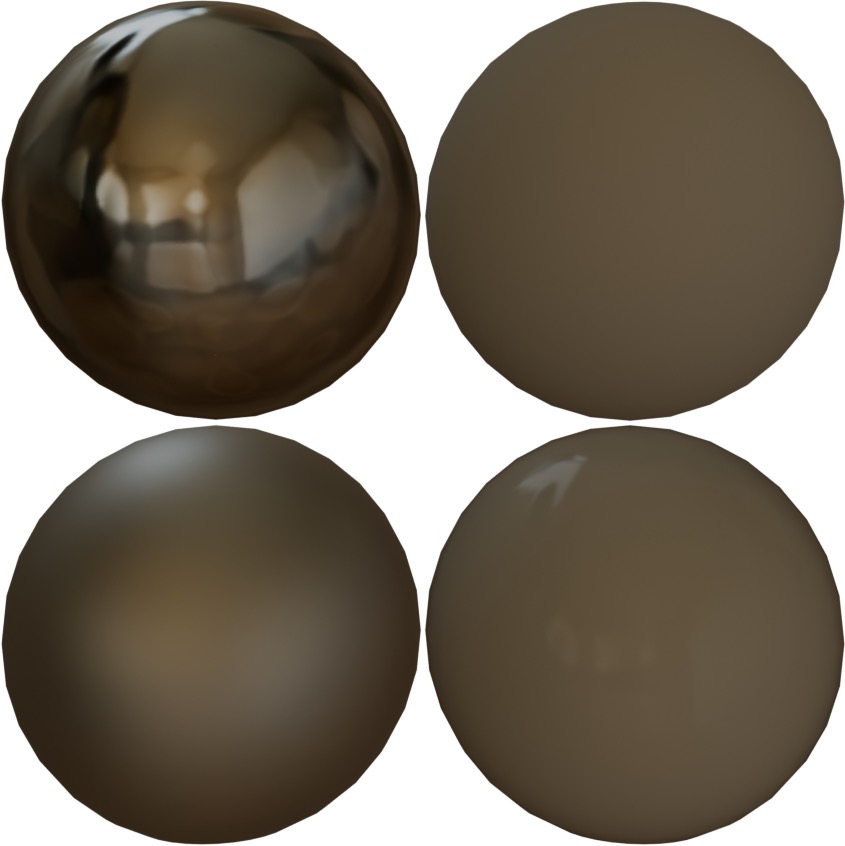}&
    \includegraphics[width=\tmplength]{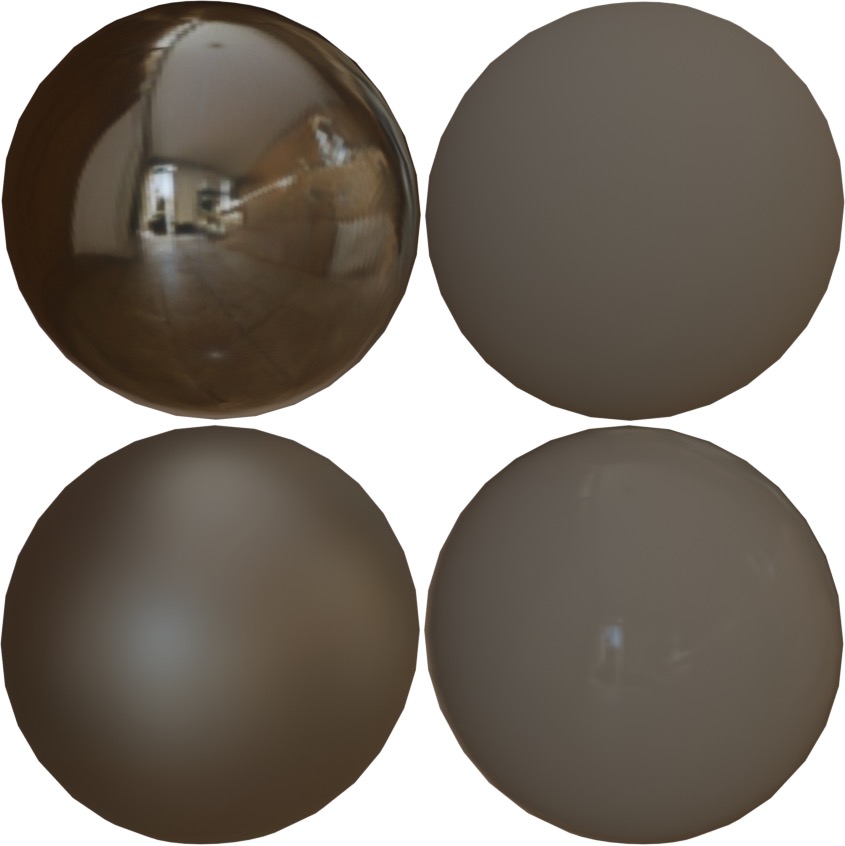}&
    \includegraphics[width=\tmplength]{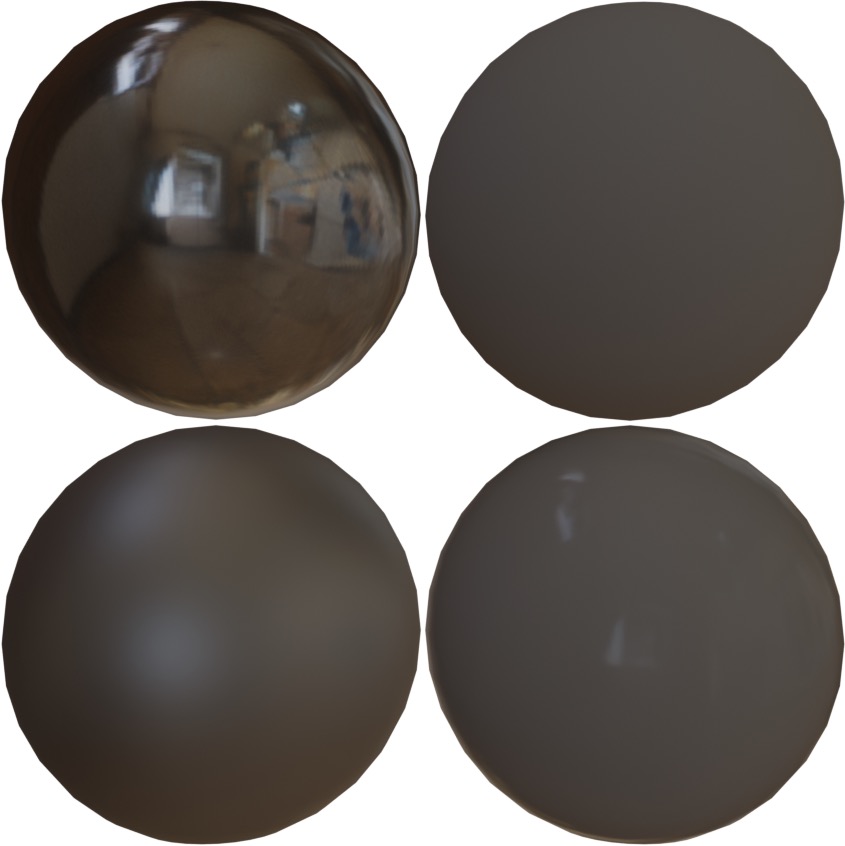}&
    \includegraphics[width=\tmplength]{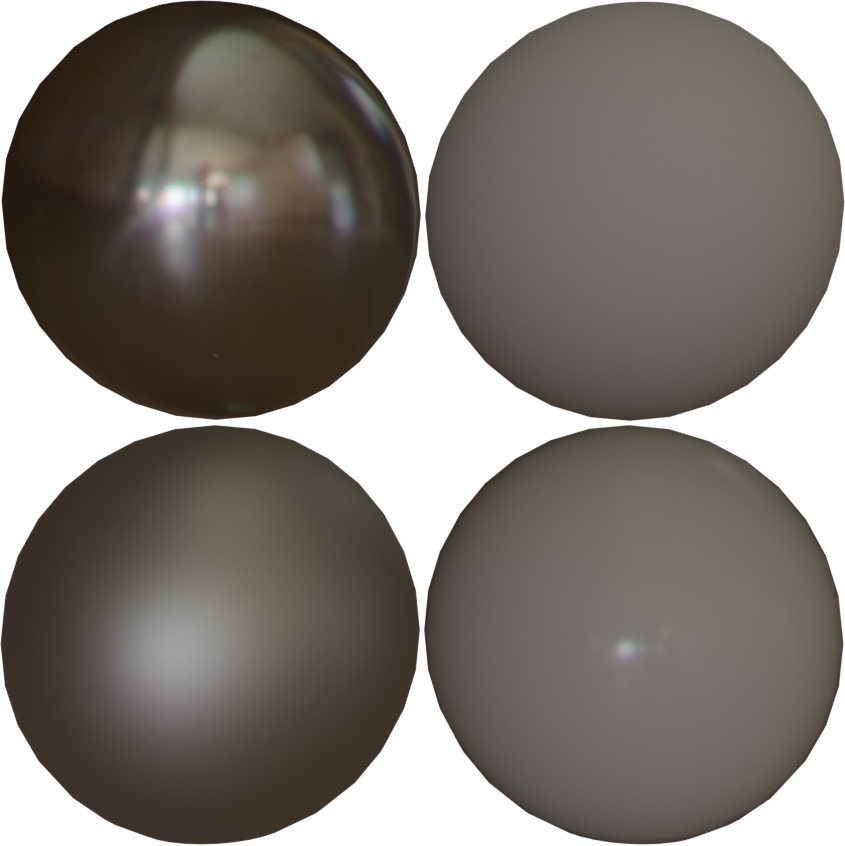}\\
    \multirow[t]{2}{*}{\raisebox{-0.5\height}{\includegraphics[width=3\tmplength]{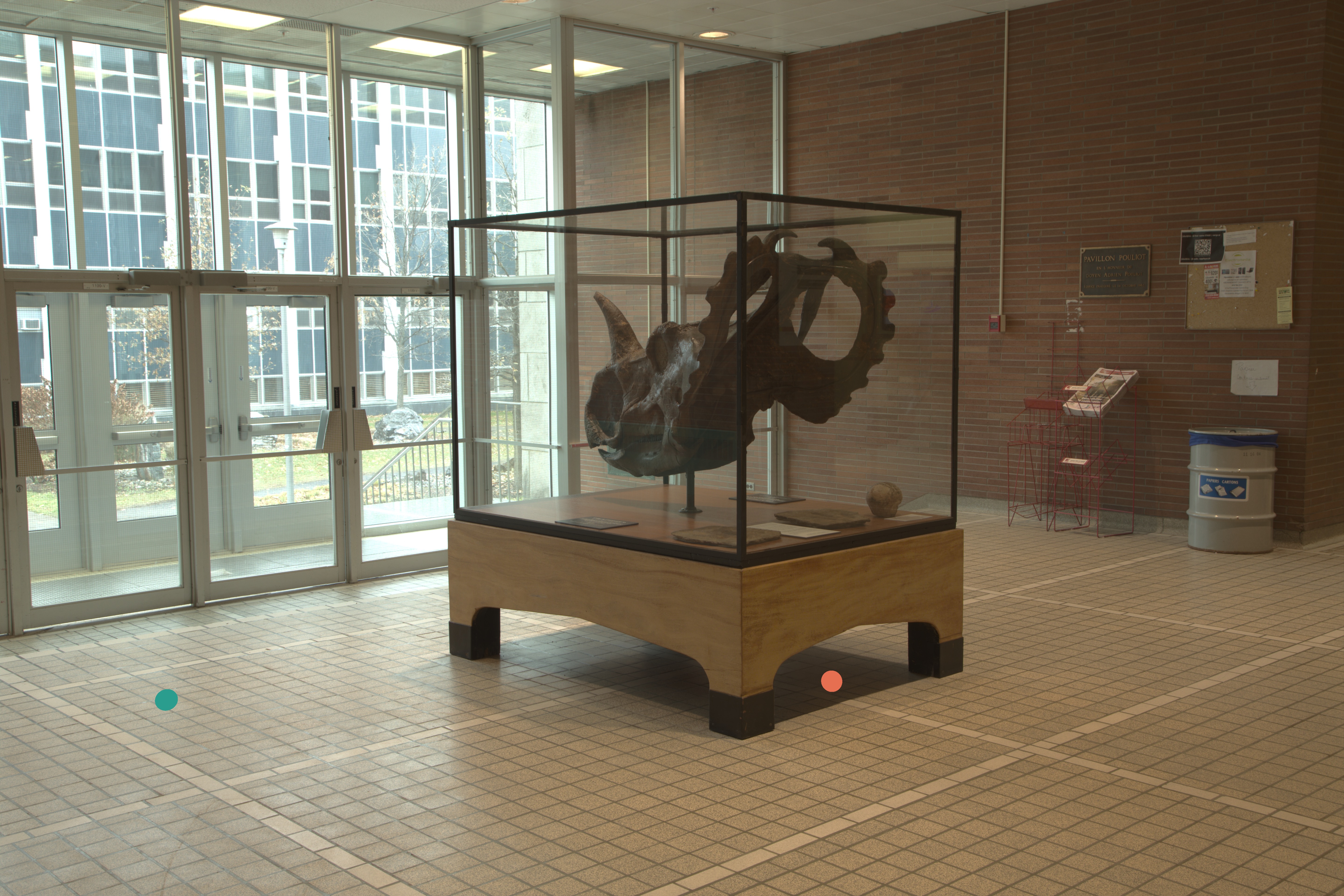}}}&
         \color{color2}\rule[4pt]{1.2pt}{1.7cm} 
    &
    \includegraphics[width=\tmplength]{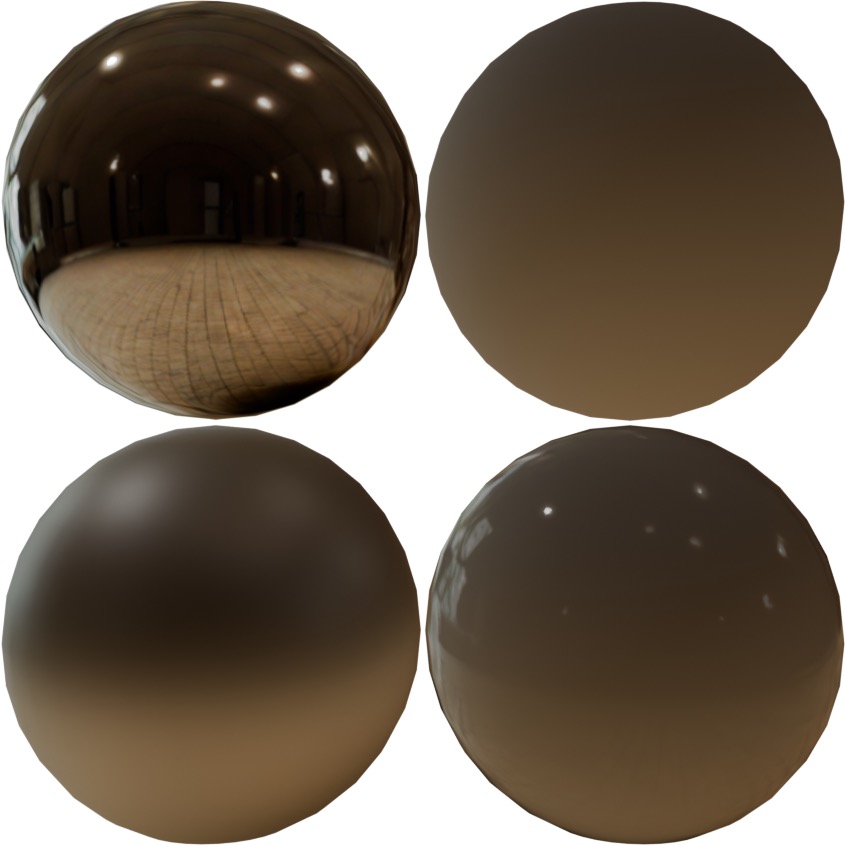}&
    \includegraphics[width=\tmplength]{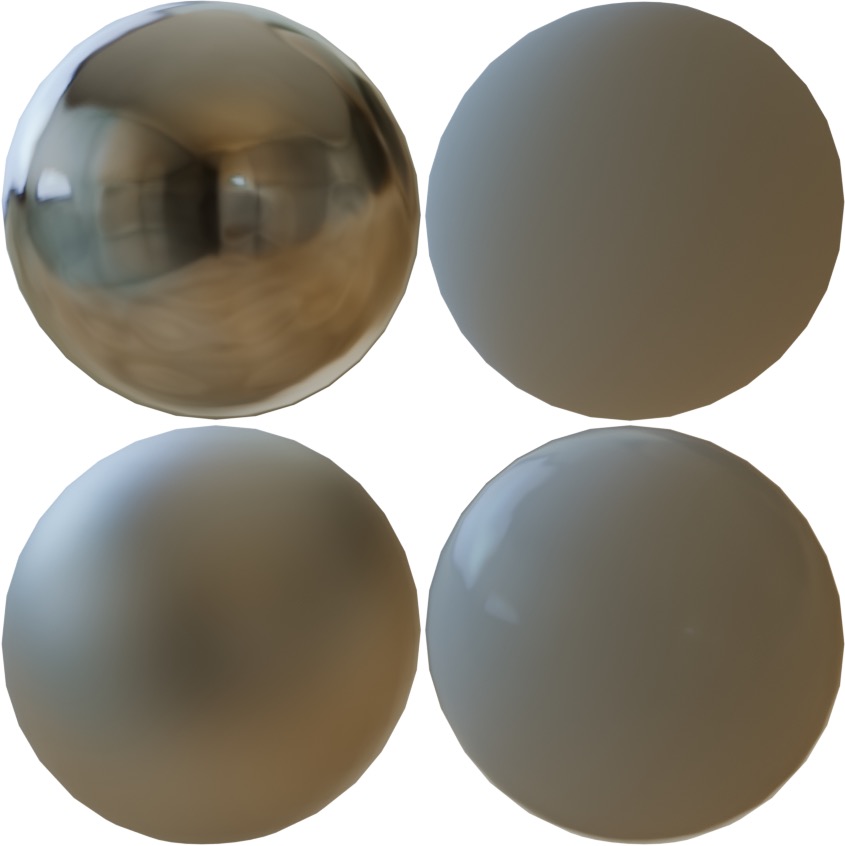}&
    \includegraphics[width=\tmplength]{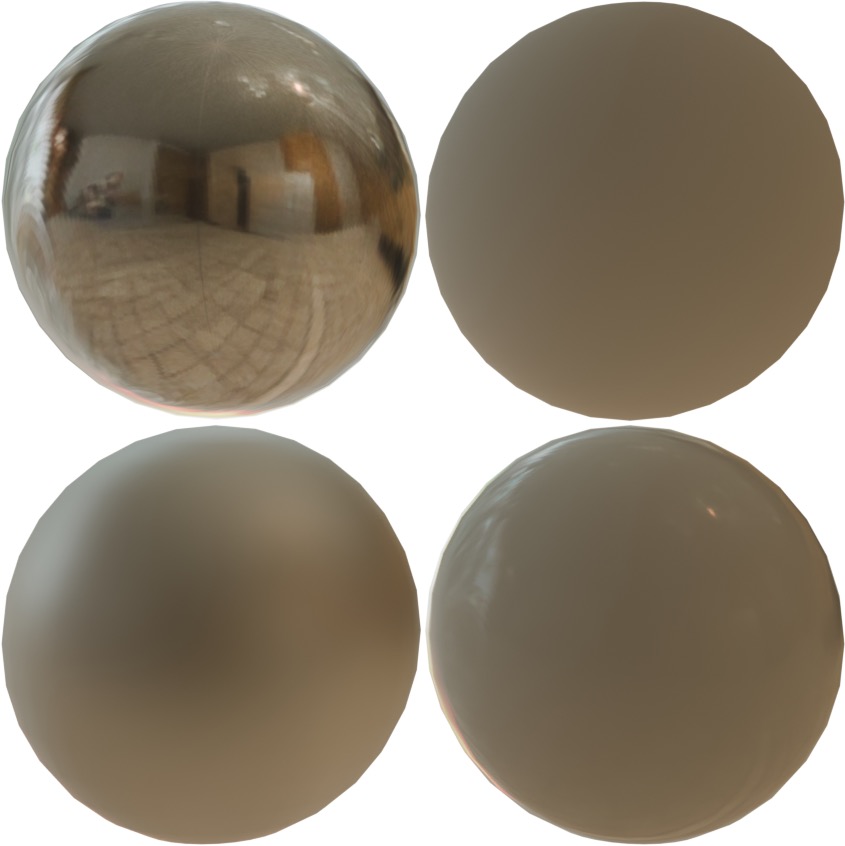}&
    \includegraphics[width=\tmplength]{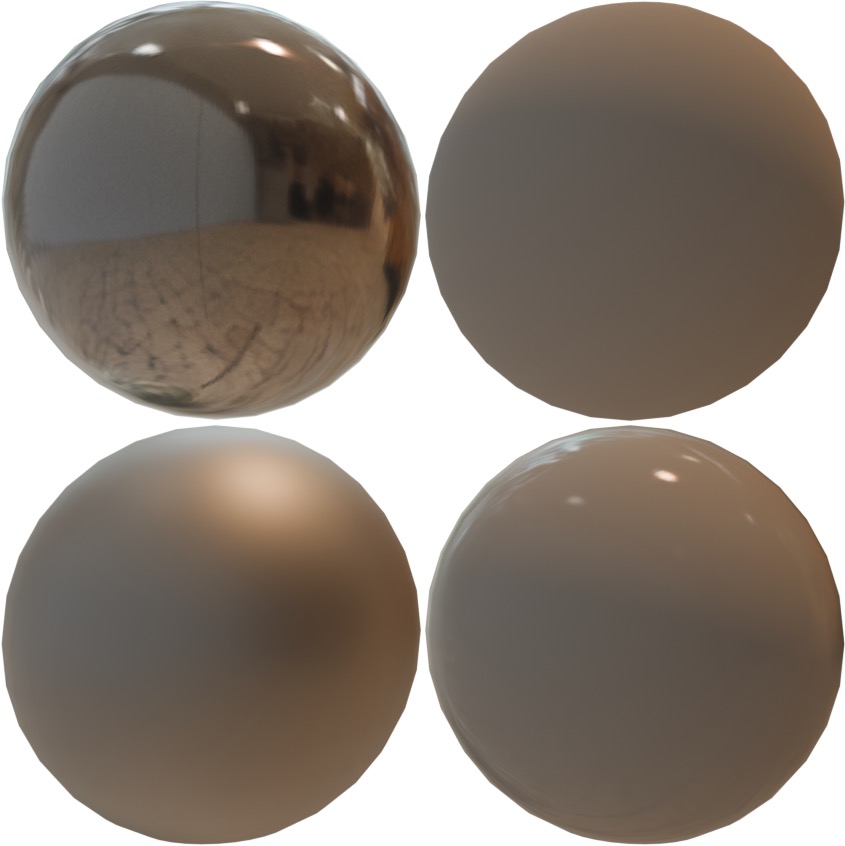}&
    \includegraphics[width=\tmplength]{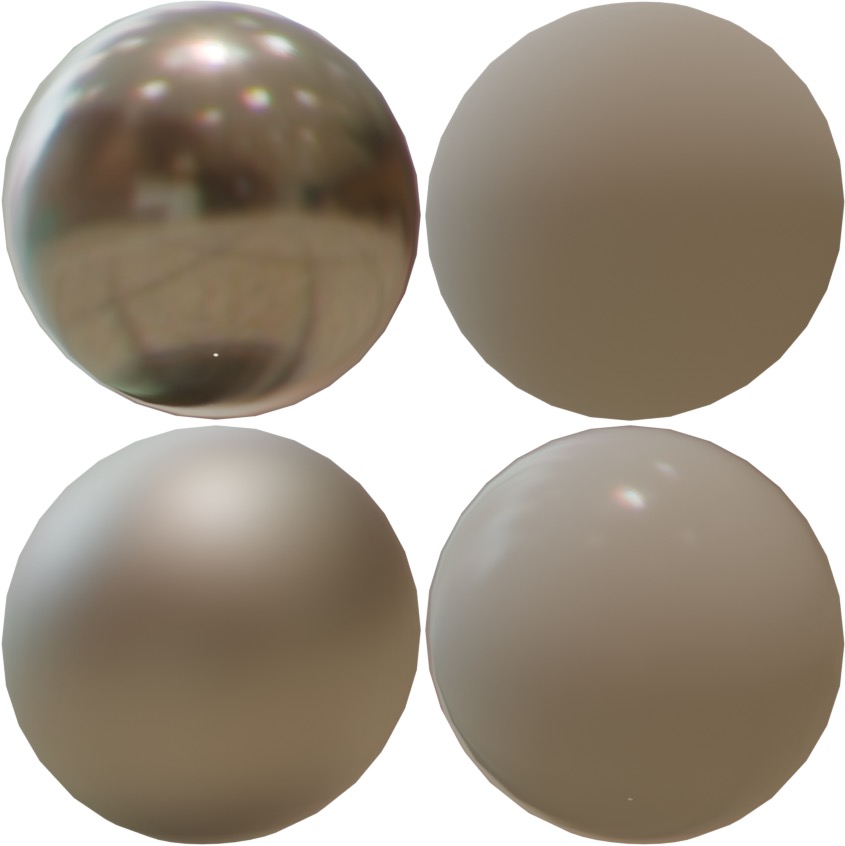}\\
    &
         \color{color5}\rule[4pt]{1.2pt}{1.7cm}
    &
    \includegraphics[width=\tmplength]{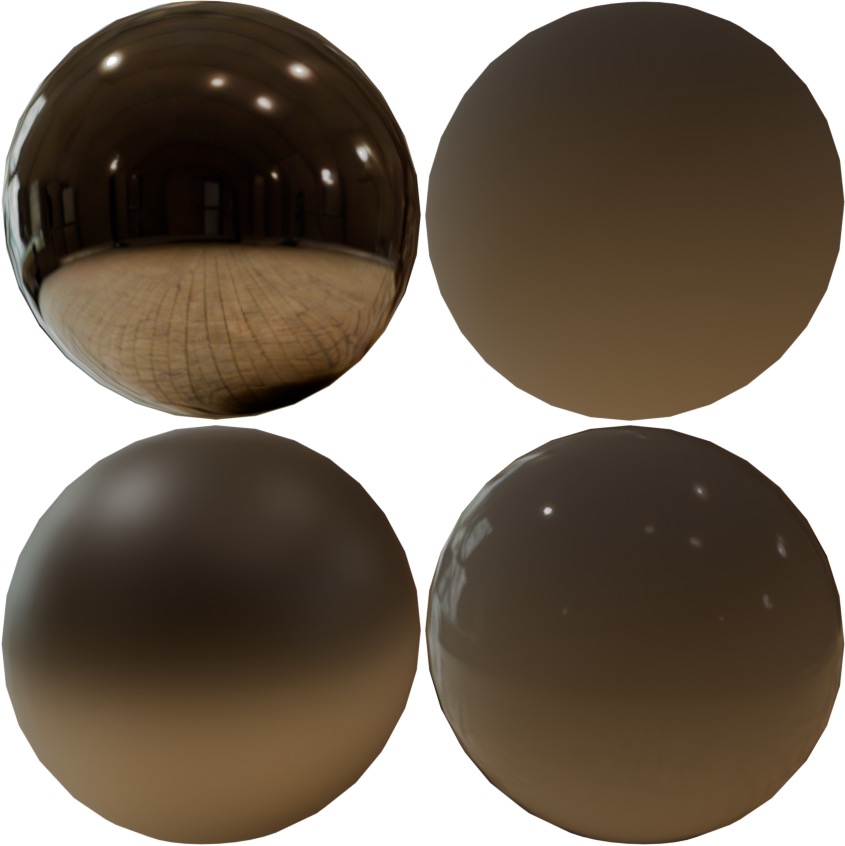}&
    \includegraphics[width=\tmplength]{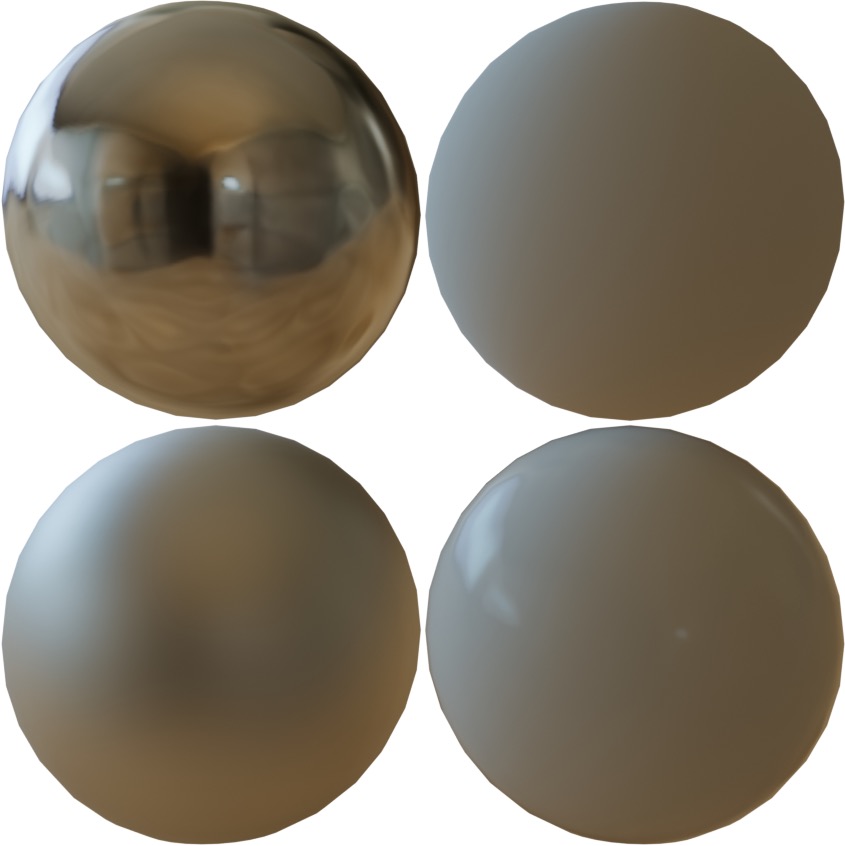}&
    \includegraphics[width=\tmplength]{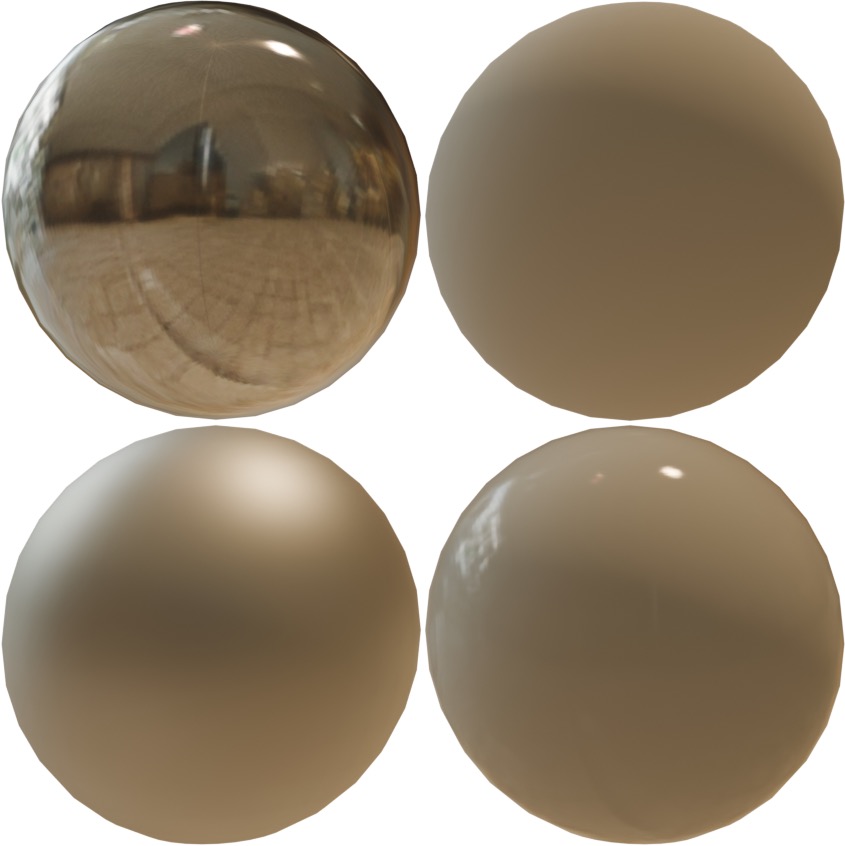}&
    \includegraphics[width=\tmplength]{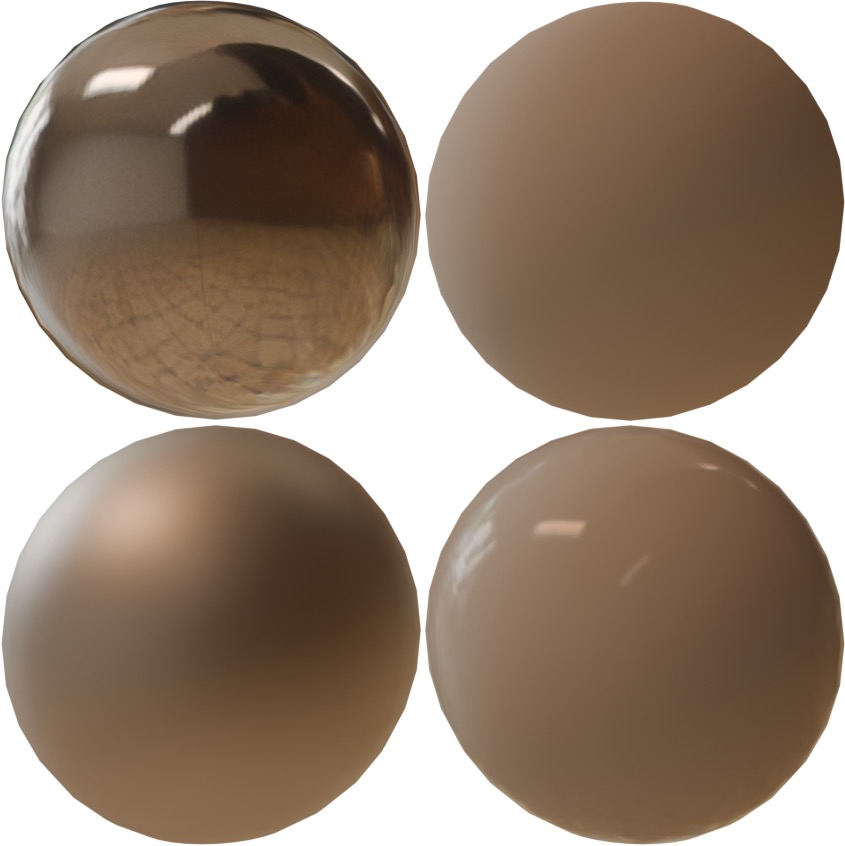}&
    \includegraphics[width=\tmplength]{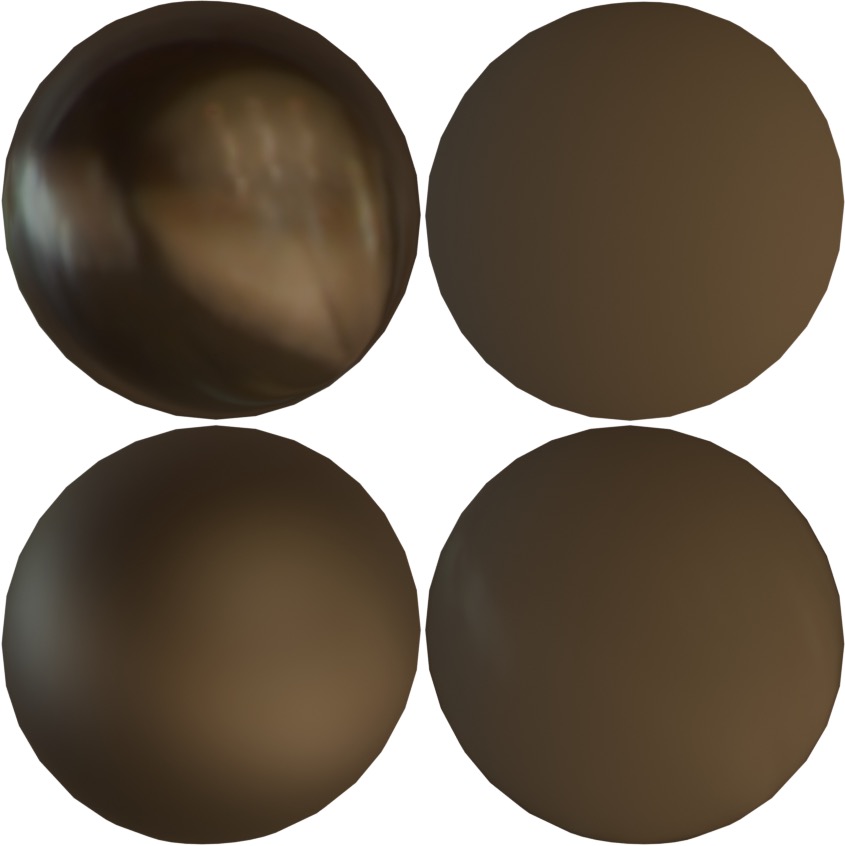}\\
    \end{tabular}
    \caption{Additional sample predictions from the Laval Indoor Spatially Varying test set~\cite{garon2019fast} for, from left to right: DiffusionLight~\cite{Phongthawee2023DiffusionLight}, 4D Lighting~\cite{4DLighting}, and the image and video versions of the proposed \themethod. We visualize predictions by rendering the same four test spheres used for the quantitative metrics (see \cref{tab:quant_single_img}): mirror (top left), diffuse (top right), matte (bottom left) and glossy (bottom right).} 
    \label{fig:single_im_qualitative_laval}
\end{figure*}
Sample predictions from The Laval Indoor Spatially Varying HDR dataset~\cite{garon2019fast} are presented in \cref{fig:single_im_qualitative_laval}.

\begin{figure*}
   \centering
   \footnotesize
   \setlength{\tabcolsep}{2.5pt}
   \setlength{\tmplength}{0.12\linewidth}
   \newlength{\tmpwidth}
   \setlength{\tmpwidth}{2.0cm}

\begin{tabular}{>{\centering\arraybackslash}m{1.1cm} m{\tmpwidth}m{\tmpwidth}m{\tmpwidth}p{0.05cm}m{\tmpwidth}m{\tmpwidth}m{\tmpwidth}}
\multirow{1}{*}{Scene} &
\includegraphics[width=\tmplength]{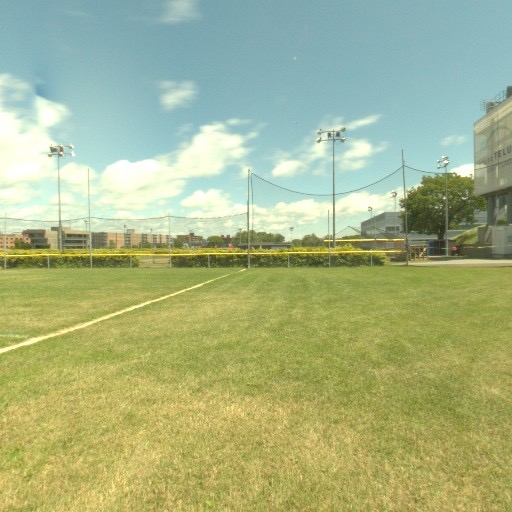} &
\includegraphics[width=\tmplength]{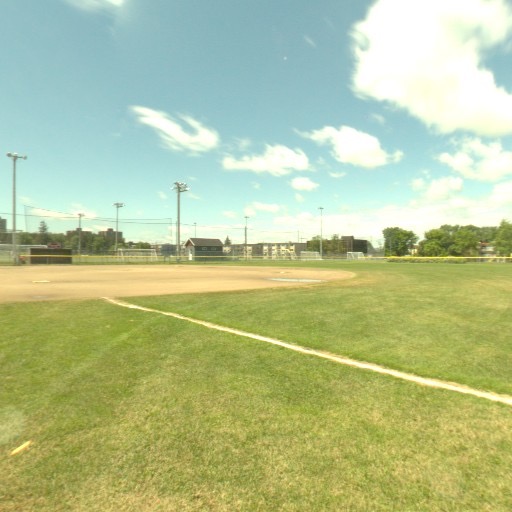} &
\includegraphics[width=\tmplength]{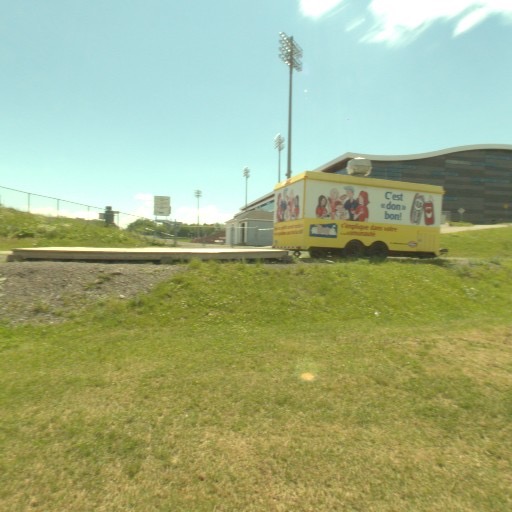}  & &
\includegraphics[width=\tmplength]{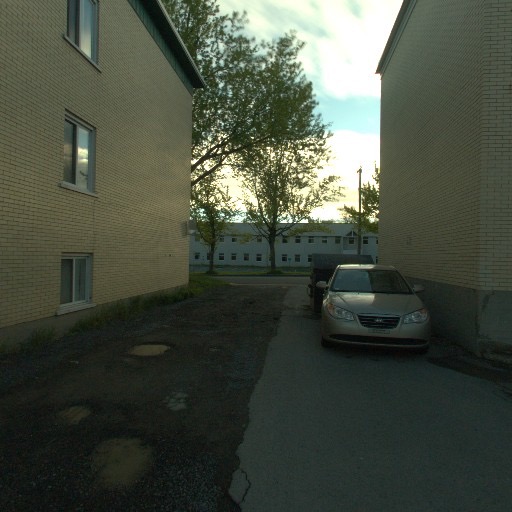} &
\includegraphics[width=\tmplength]{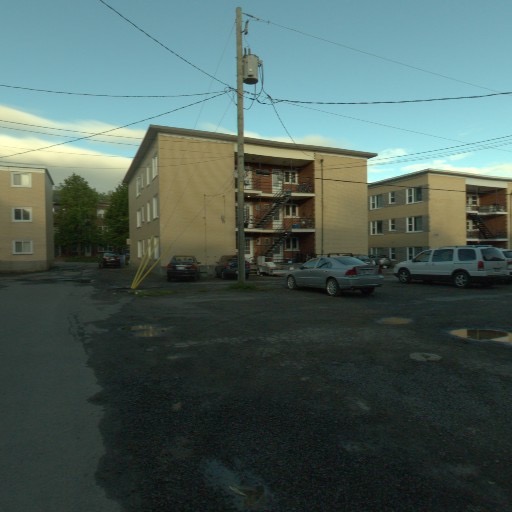} &
\includegraphics[width=\tmplength]{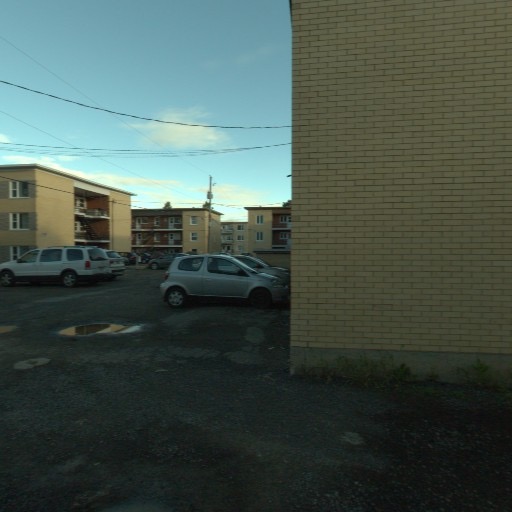} 
\\

Diff.Light &
\includegraphics[width=\tmplength]{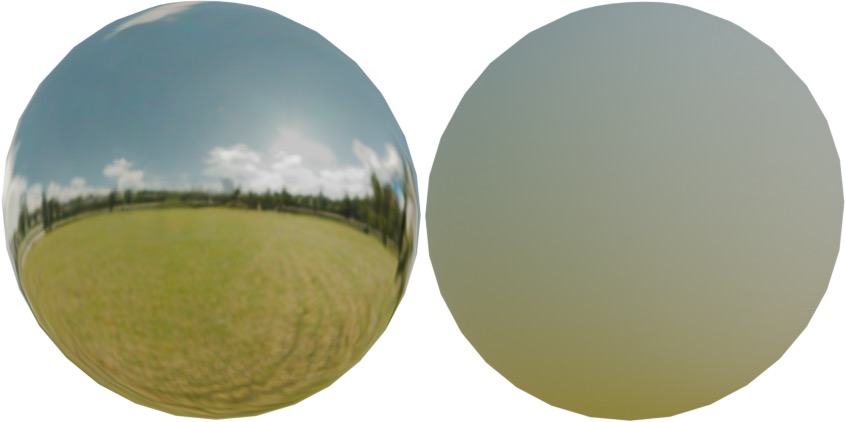} &
\includegraphics[width=\tmplength]{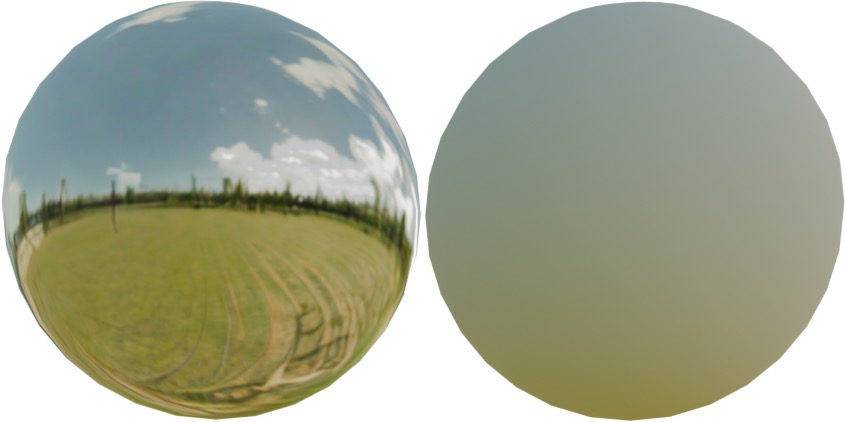} &
\includegraphics[width=\tmplength]{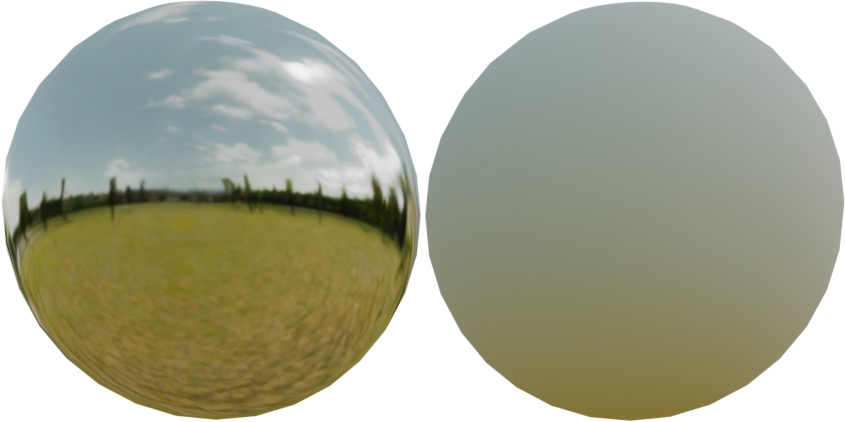} & &
\includegraphics[width=\tmplength]{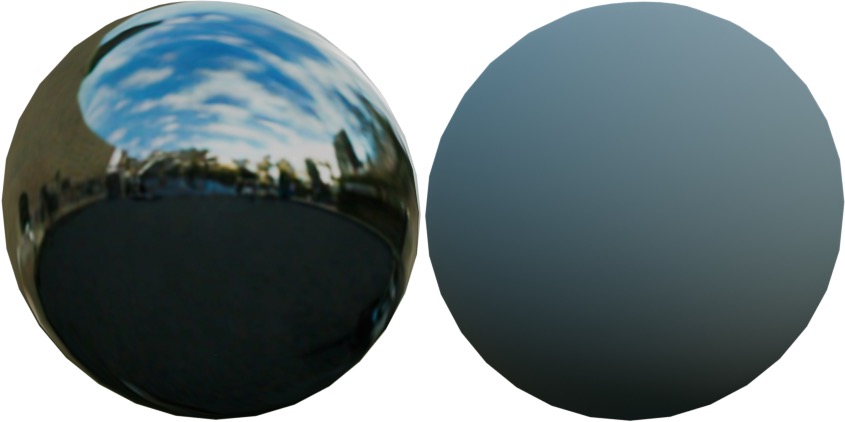} &
\includegraphics[width=\tmplength]{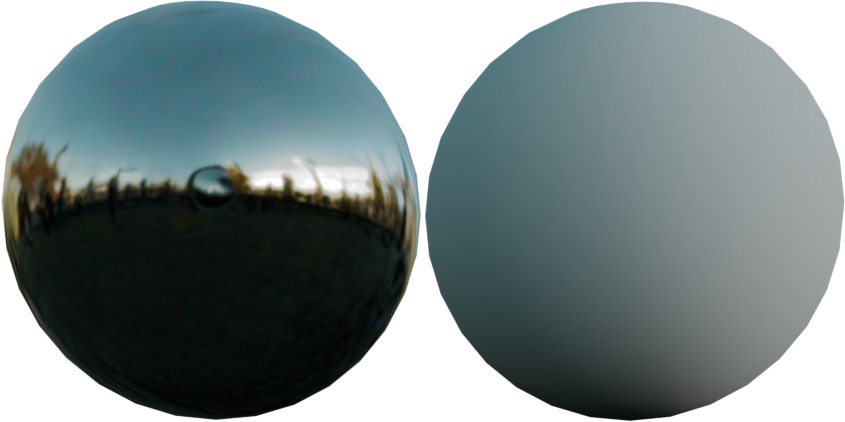} &
\includegraphics[width=\tmplength]{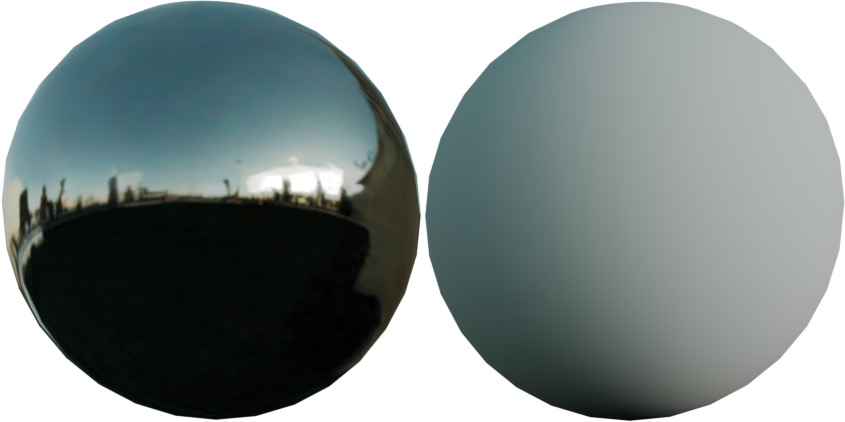} 
\\

4D \mbox{Lighting} &
\includegraphics[width=\tmplength]{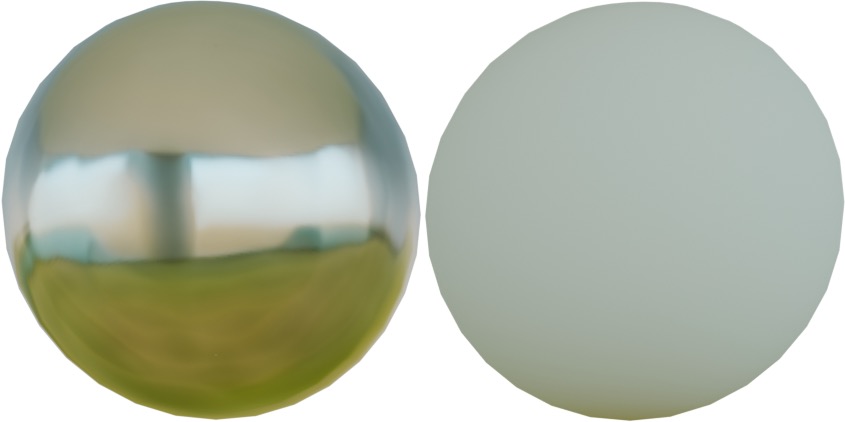} &
\includegraphics[width=\tmplength]{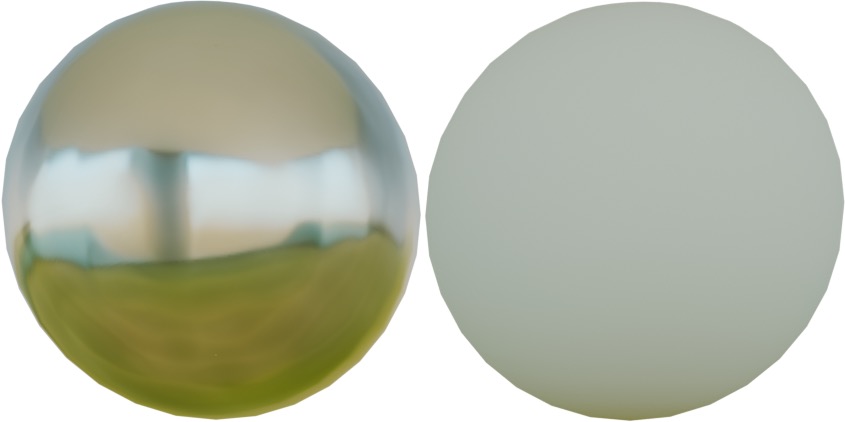} &
\includegraphics[width=\tmplength]{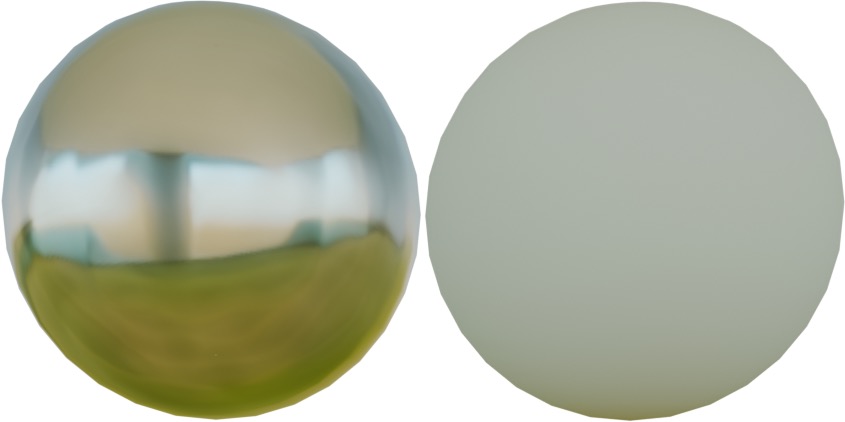} & &
\includegraphics[width=\tmplength]{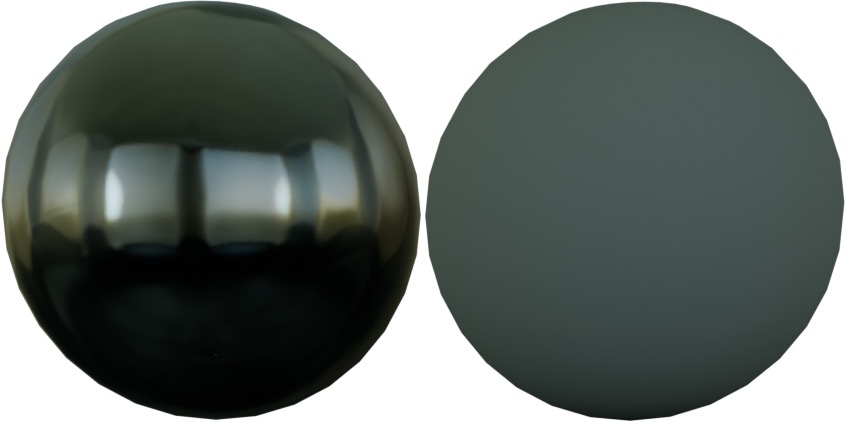} &
\includegraphics[width=\tmplength]{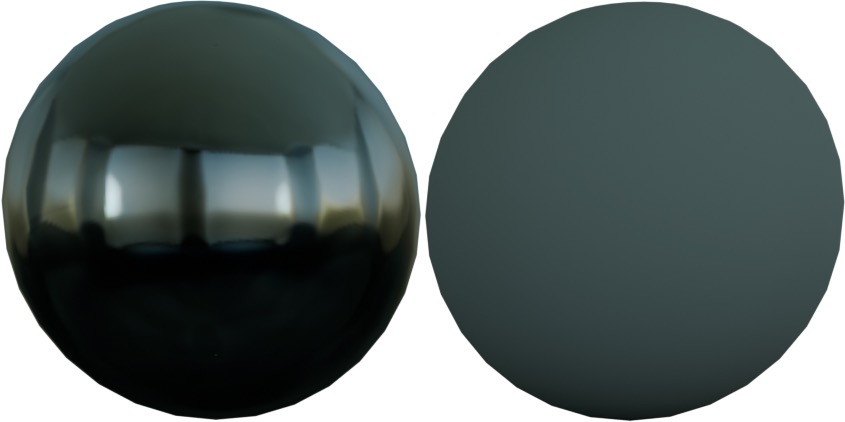} &
\includegraphics[width=\tmplength]{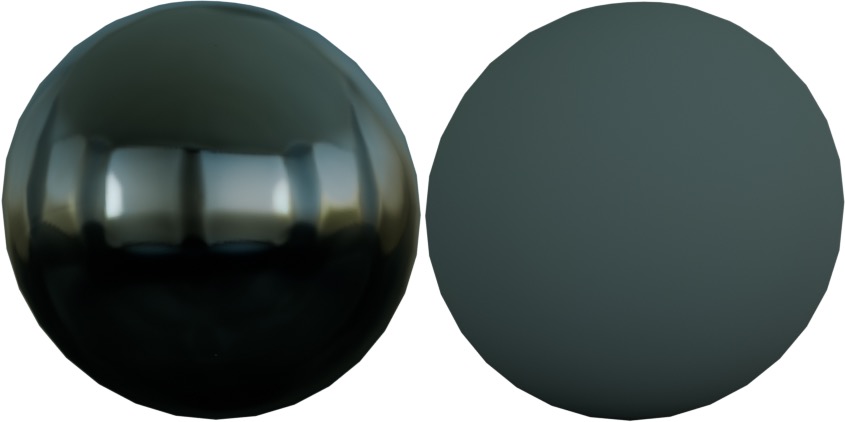} 
\\

\themethod \mbox{(image)} &
\includegraphics[width=\tmplength]{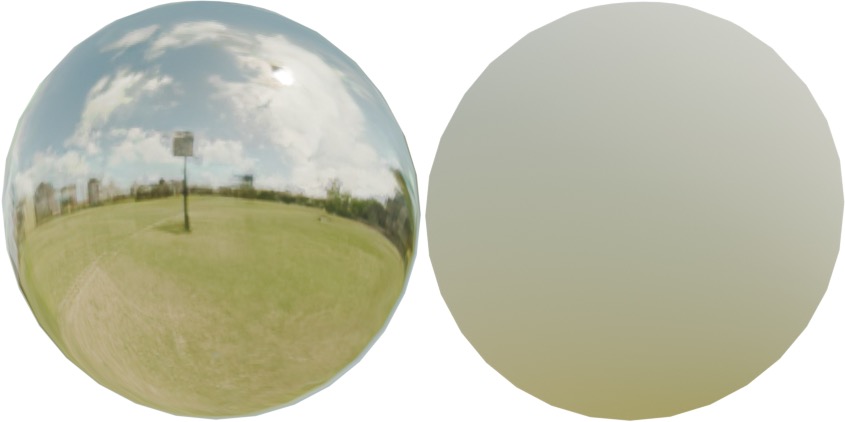} &
\includegraphics[width=\tmplength]{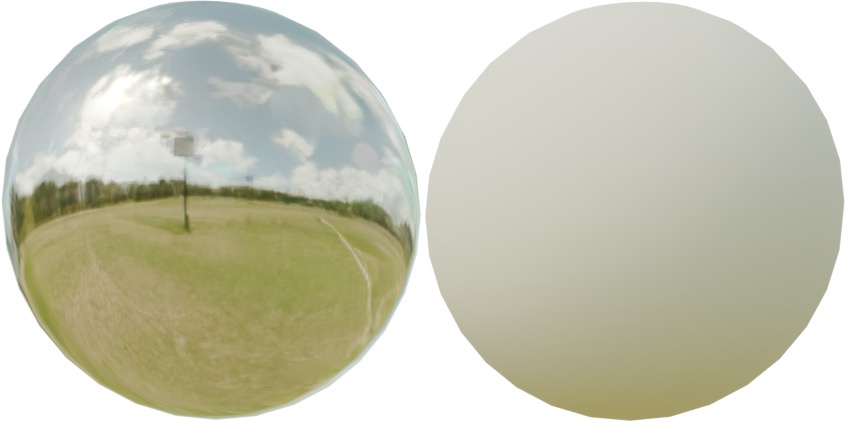} &
\includegraphics[width=\tmplength]{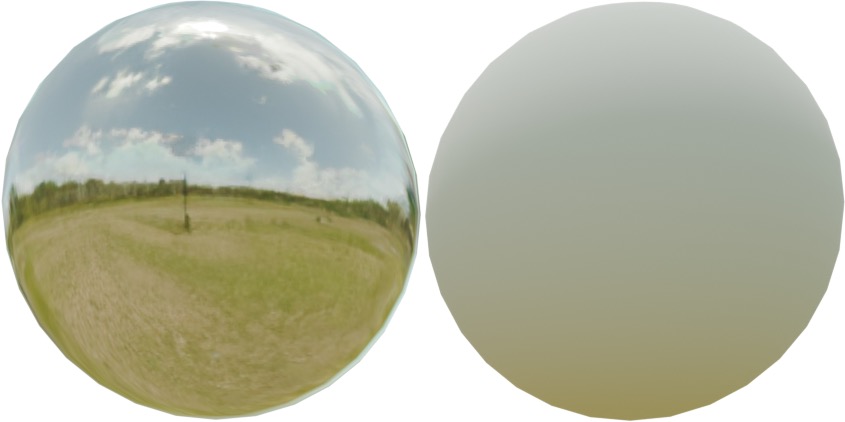} &&
\includegraphics[width=\tmplength]{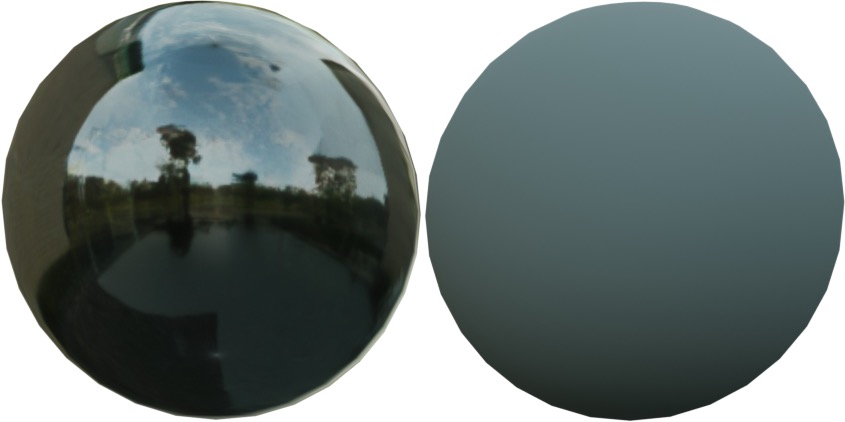} &
\includegraphics[width=\tmplength]{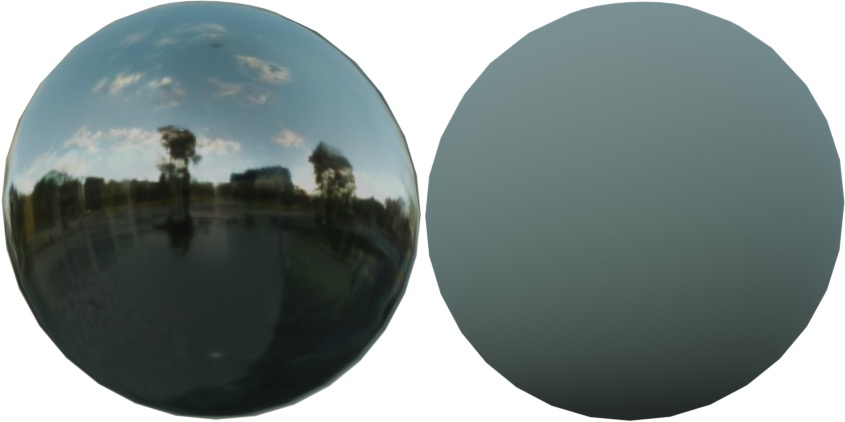} &
\includegraphics[width=\tmplength]{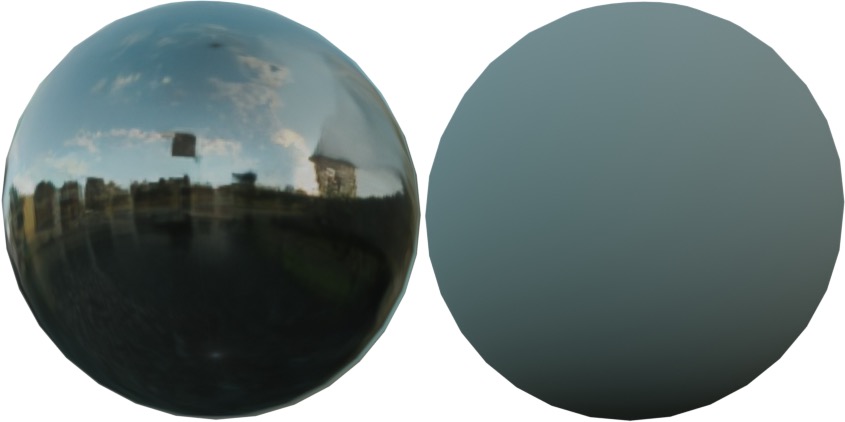} 
\\

\themethod (video) &
\includegraphics[width=\tmplength]{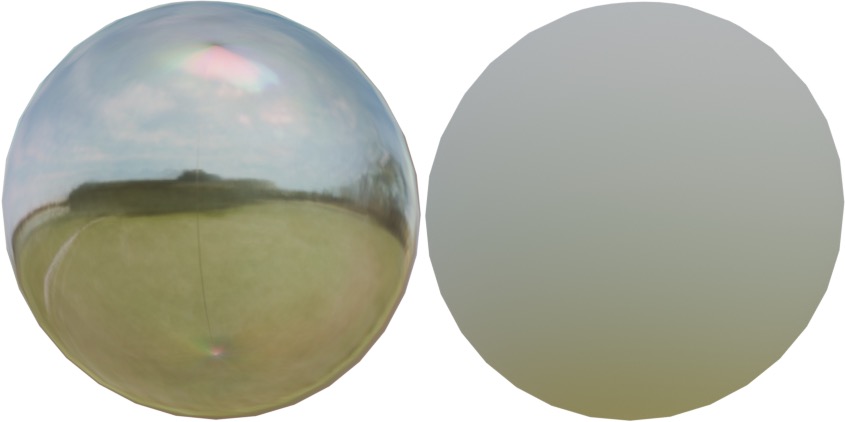} &
\includegraphics[width=\tmplength]{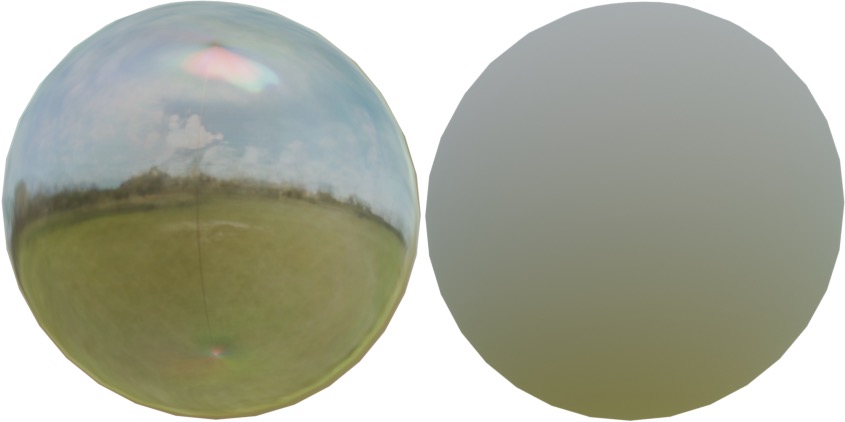} &
\includegraphics[width=\tmplength]{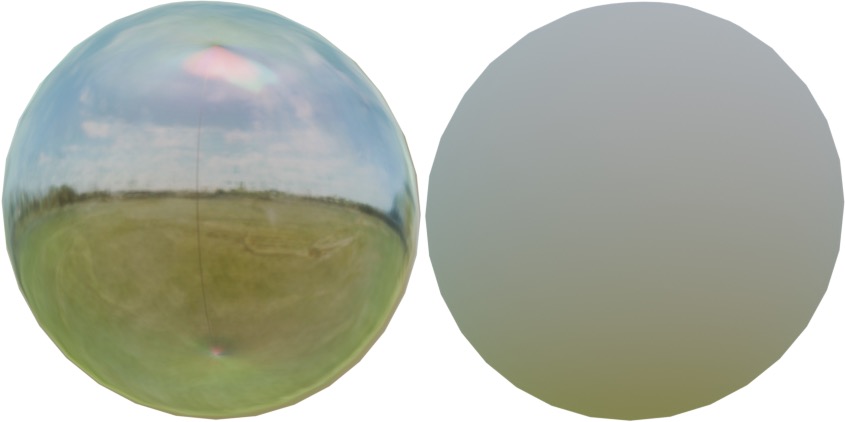} & &
\includegraphics[width=\tmplength]{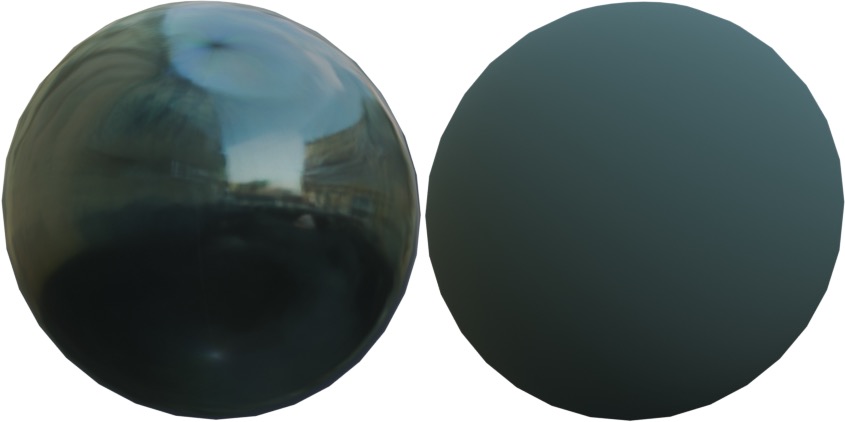} &
\includegraphics[width=\tmplength]{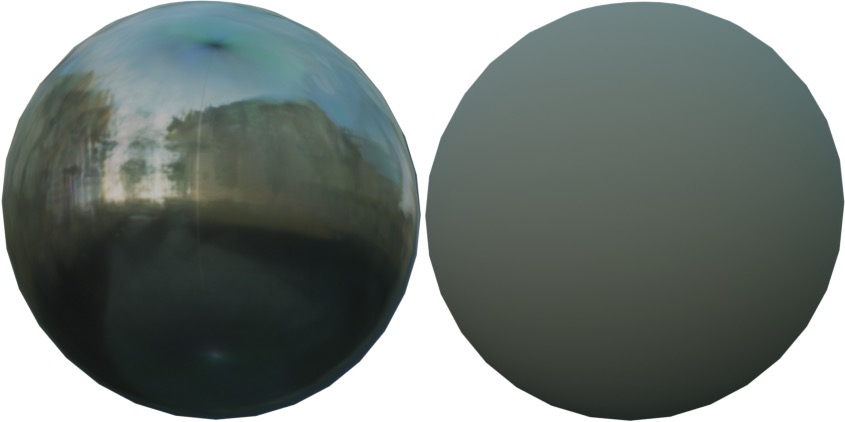} &
\includegraphics[width=\tmplength]{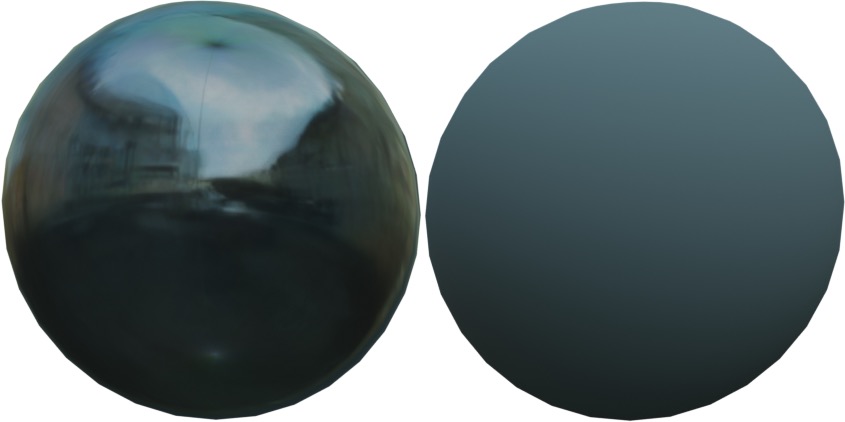} 
\\

GT &
\includegraphics[width=\tmplength]{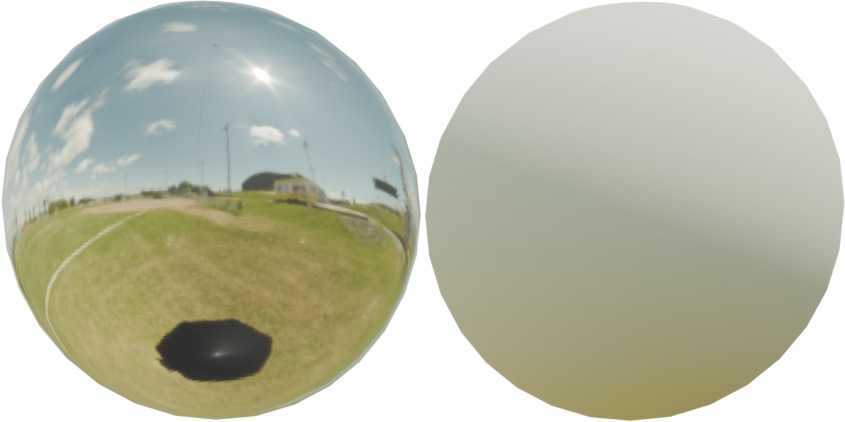} &
\includegraphics[width=\tmplength]{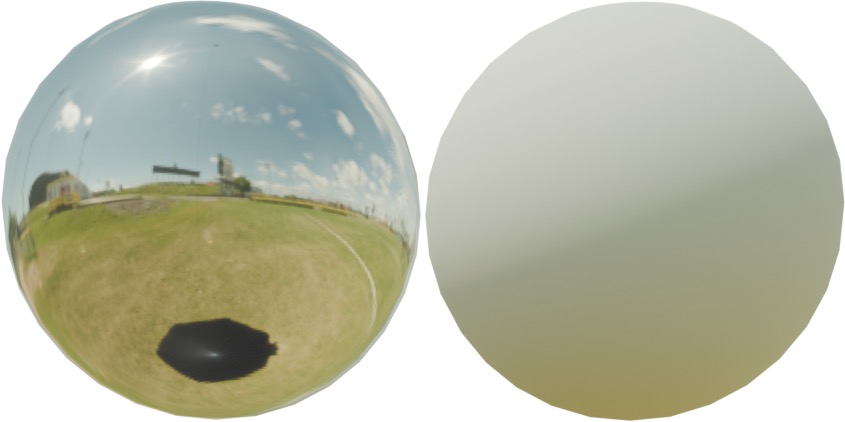} &
\includegraphics[width=\tmplength]{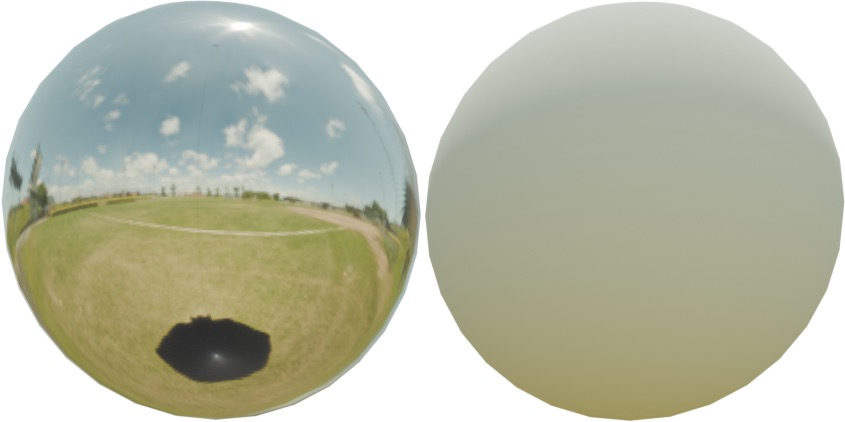} &&
\includegraphics[width=\tmplength]{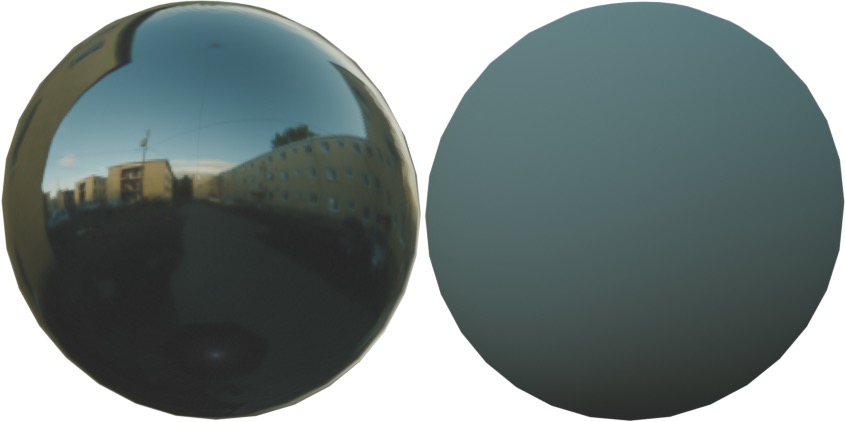} &
\includegraphics[width=\tmplength]{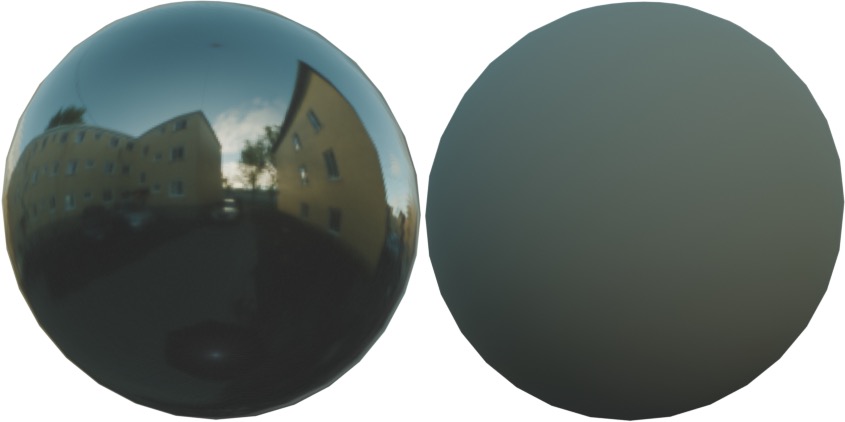} &
\includegraphics[width=\tmplength]{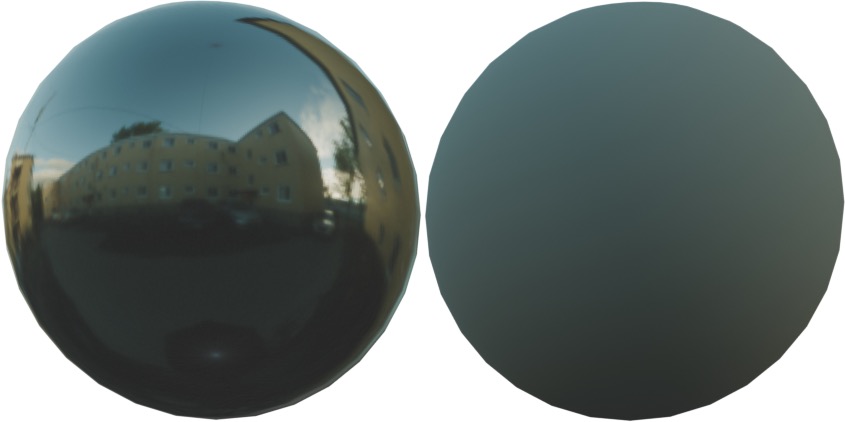} 

\end{tabular}

\caption{\change{Example qualitative prediction results on the Laval Outdoor HDR Dataset~\cite{holdgeoffroy-cvpr-19} with near-camera lighting estimations. 4D Lighting~\cite{4DLighting}, trained exclusively on indoor scenes, cannot appropriately deal with outdoor scenes.}}
\label{fig:sequence_qualitative_outdoor}
\end{figure*}

\change{Sample predictions from the Laval Outdoor HDR Dataset~\cite{holdgeoffroy-cvpr-19} are shown in \cref{fig:sequence_qualitative_outdoor}.}

\begin{figure*}[!th]
   \centering
   \footnotesize
   \setlength{\tabcolsep}{0.5pt}
   \setlength{\tmplength}{0.20\linewidth}
   \setlength{\cbarheight}{1.9cm}
    \begin{tabular}{cccccc}  &
    Input & Mirror & Diffuse & Object Insertion \\
    \rot{ \hspace{18pt} Frame 1} &
    \includegraphics[width=\tmplength]{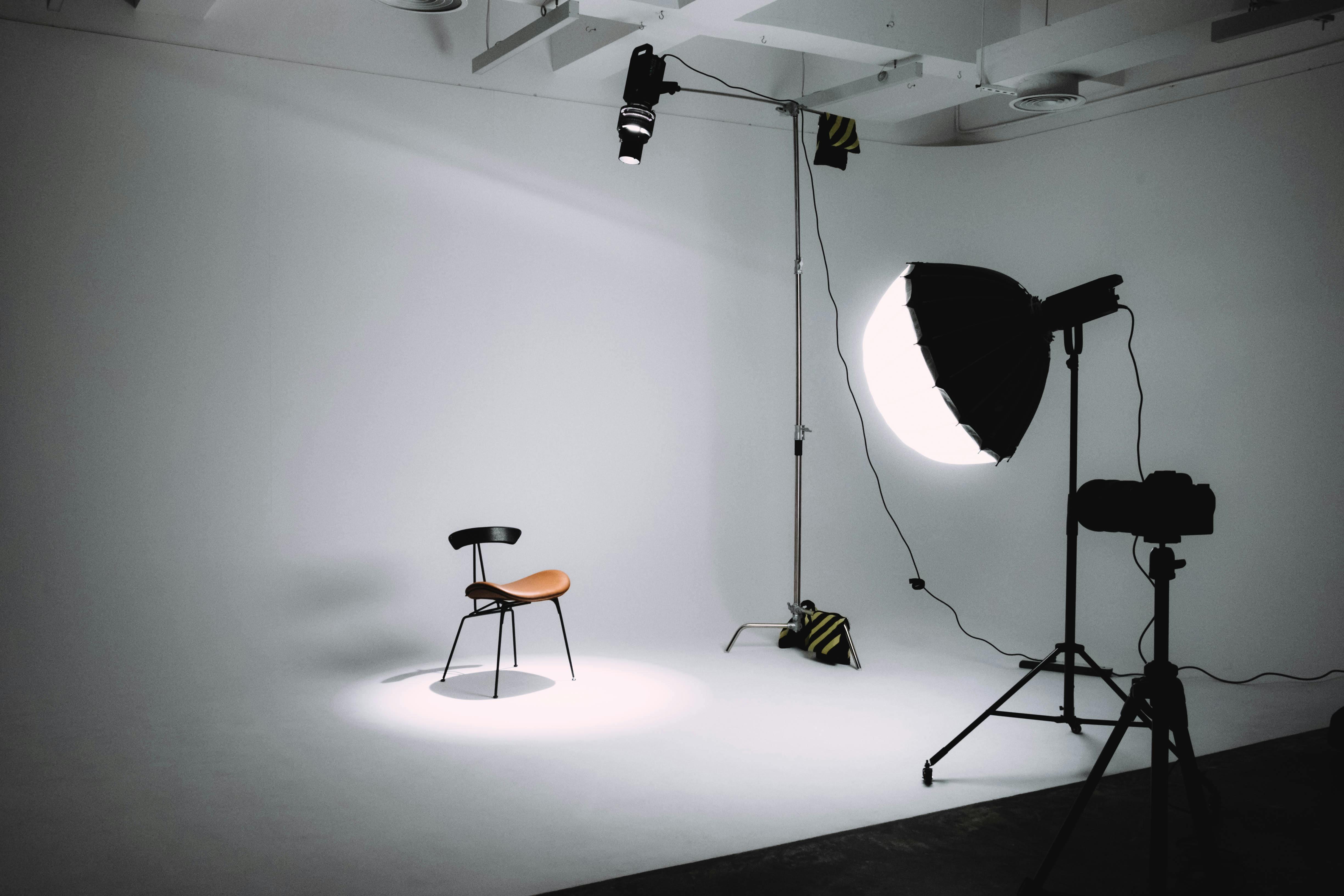} &
    \includegraphics[width=\tmplength]{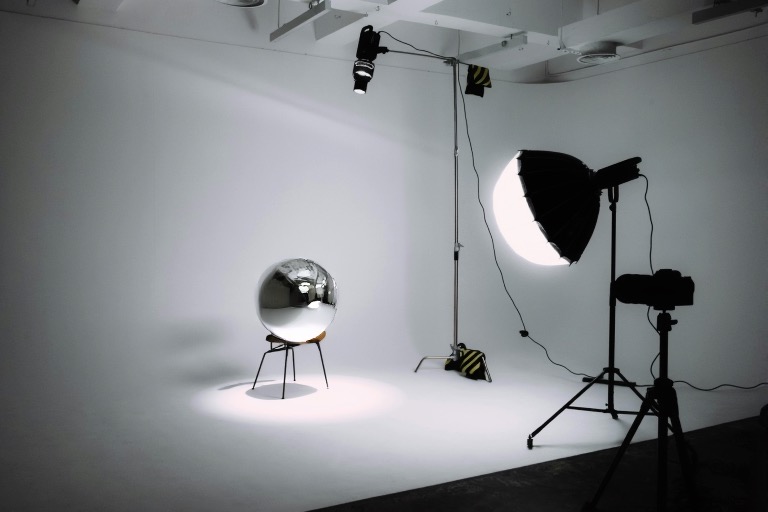}&
    \includegraphics[width=\tmplength]{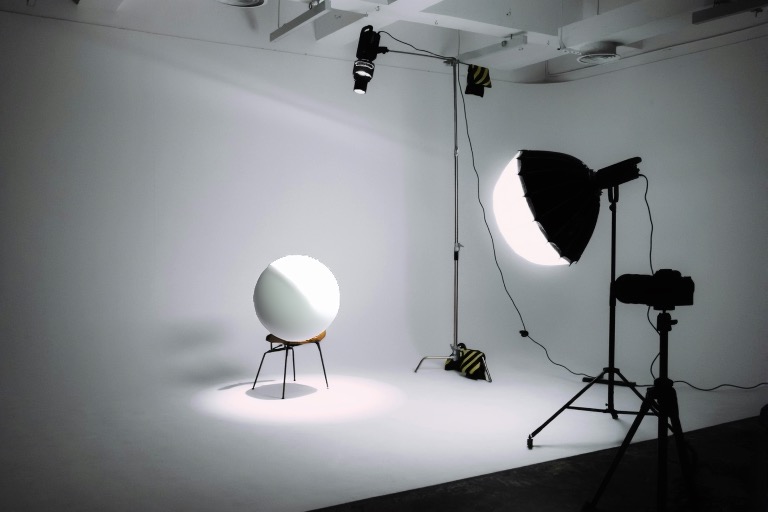}&
    \includegraphics[width=\tmplength]{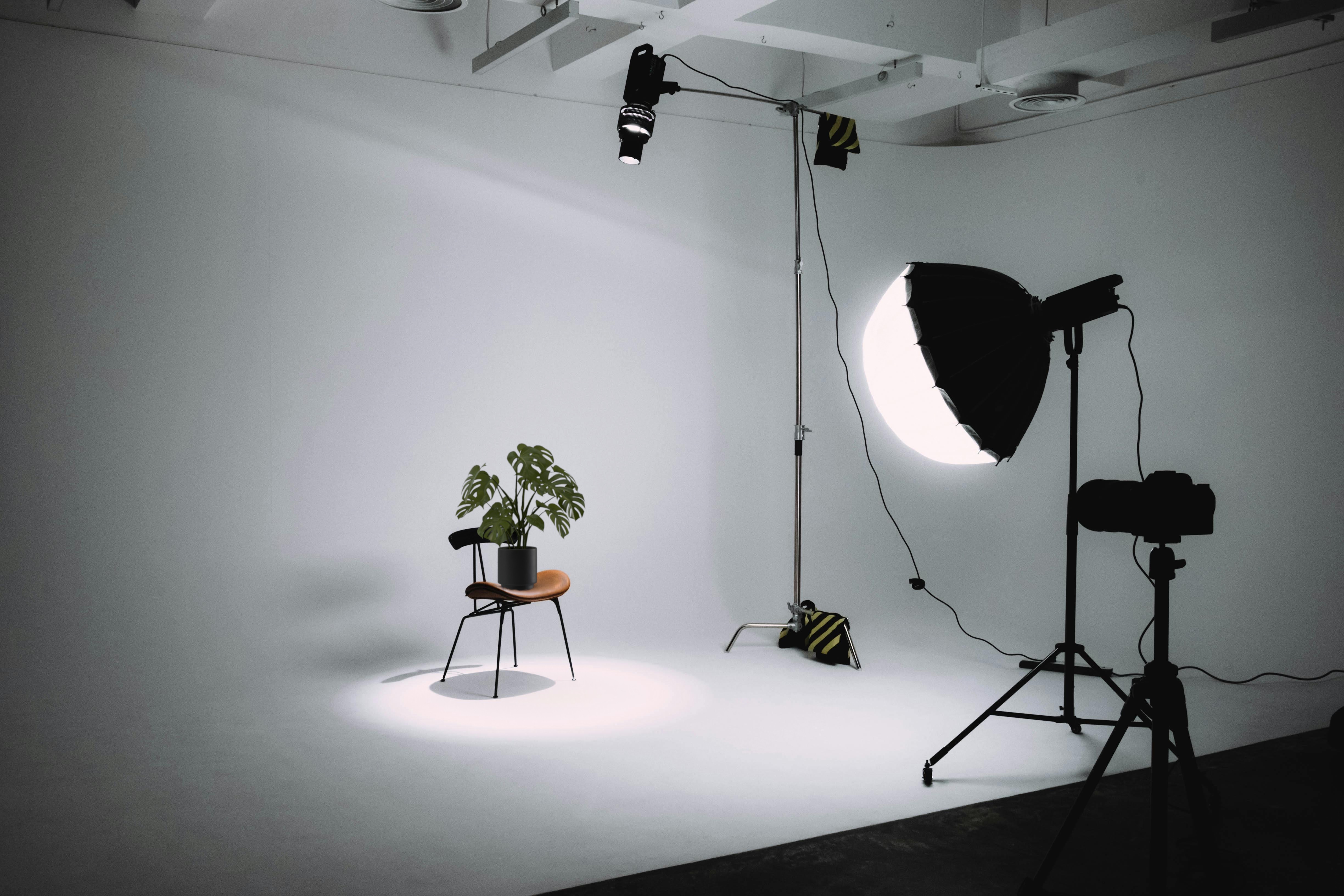}\\ \rot{ \hspace{18pt} Frame 2} &
    \includegraphics[width=\tmplength]{figures/experiments/supp/in-the-wild/spotlight/input/frame_0000.jpg} &
    \includegraphics[width=\tmplength]{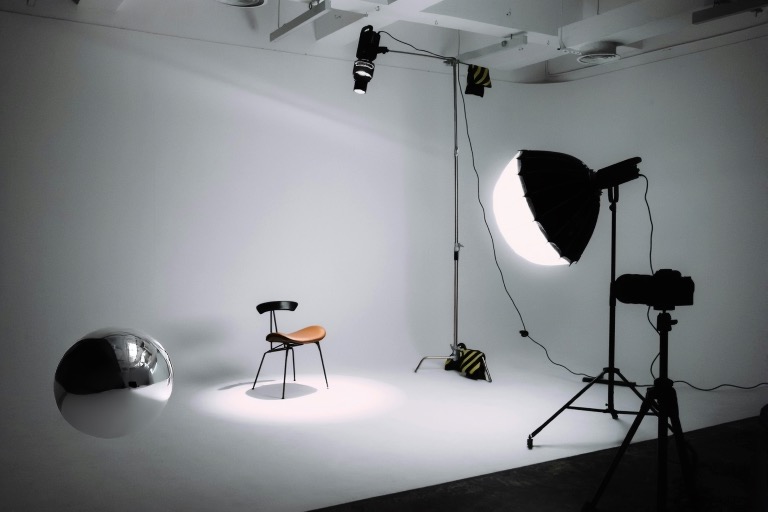}&
    \includegraphics[width=\tmplength]{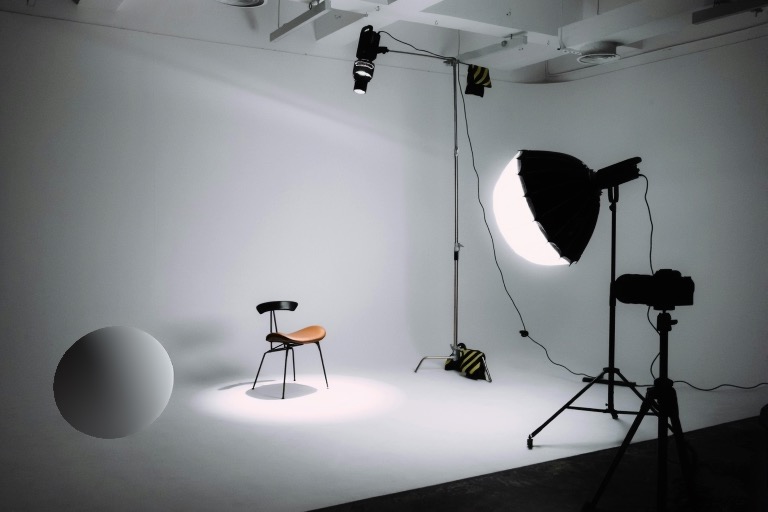}&
    \includegraphics[width=\tmplength]{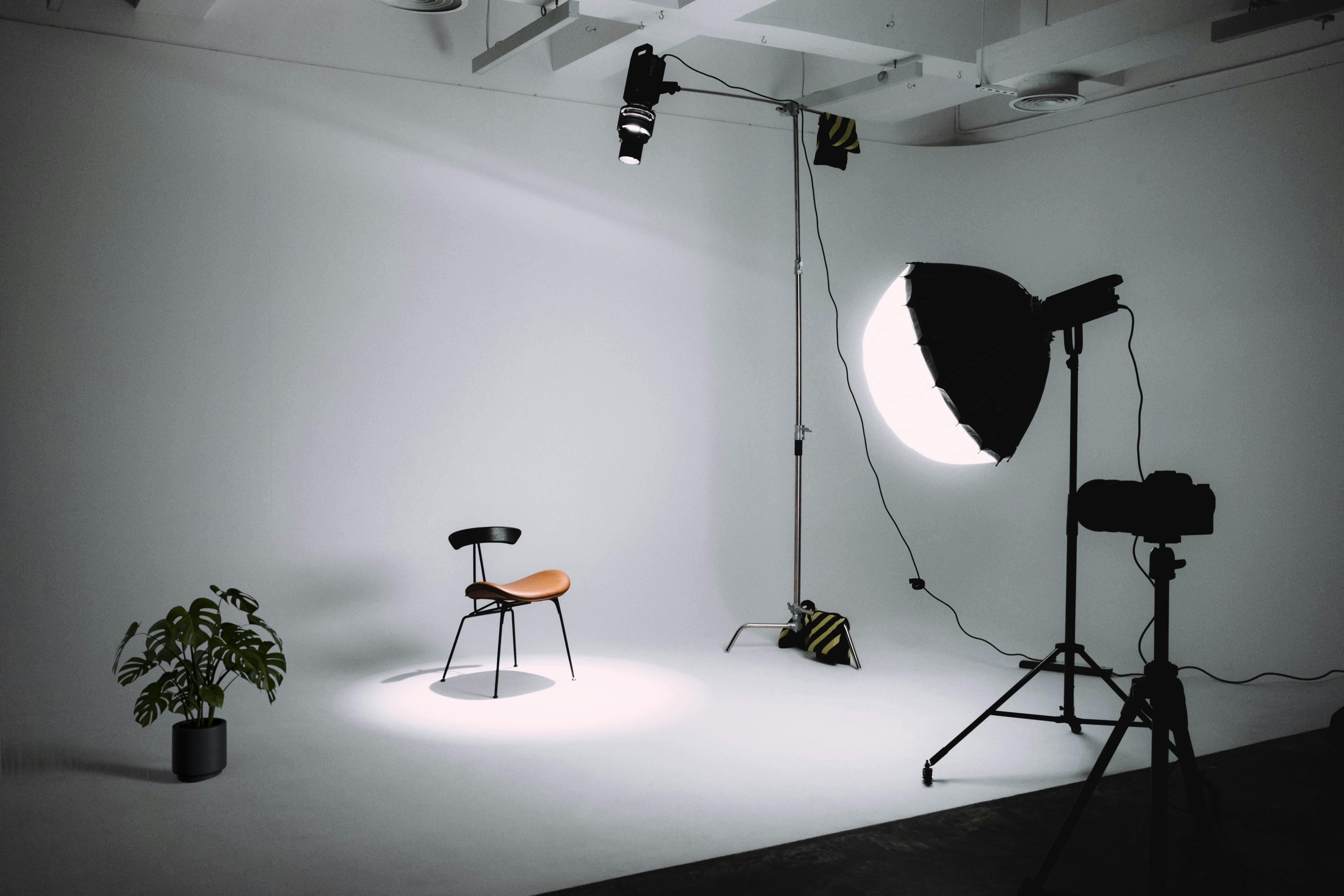}\\
    \rot{ \hspace{18pt} Frame 3} &
    \includegraphics[width=\tmplength]{figures/experiments/supp/in-the-wild/spotlight/input/frame_0000.jpg} &
    \includegraphics[width=\tmplength]{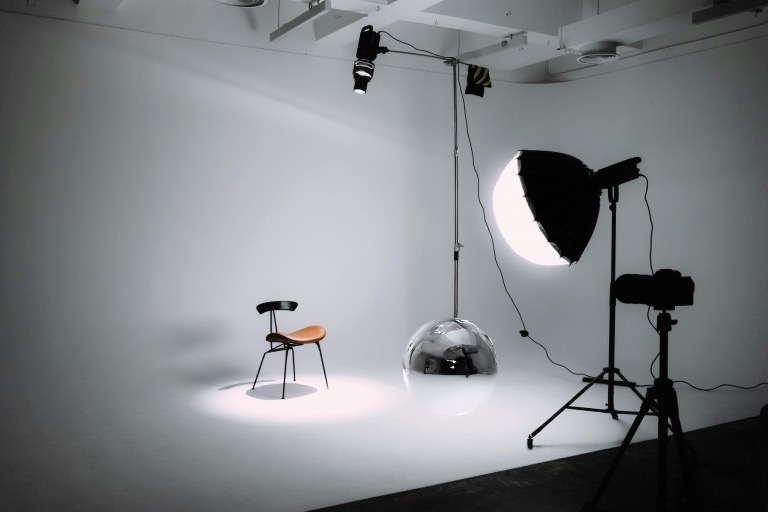}&
    \includegraphics[width=\tmplength]{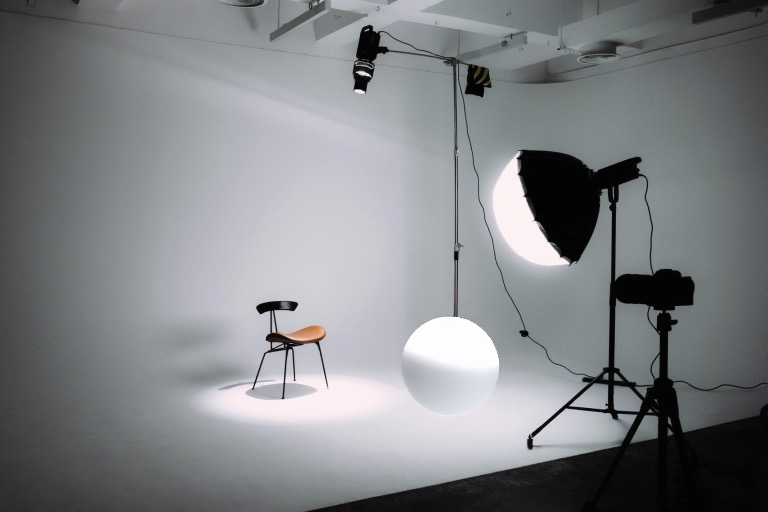}&
    \includegraphics[width=\tmplength]{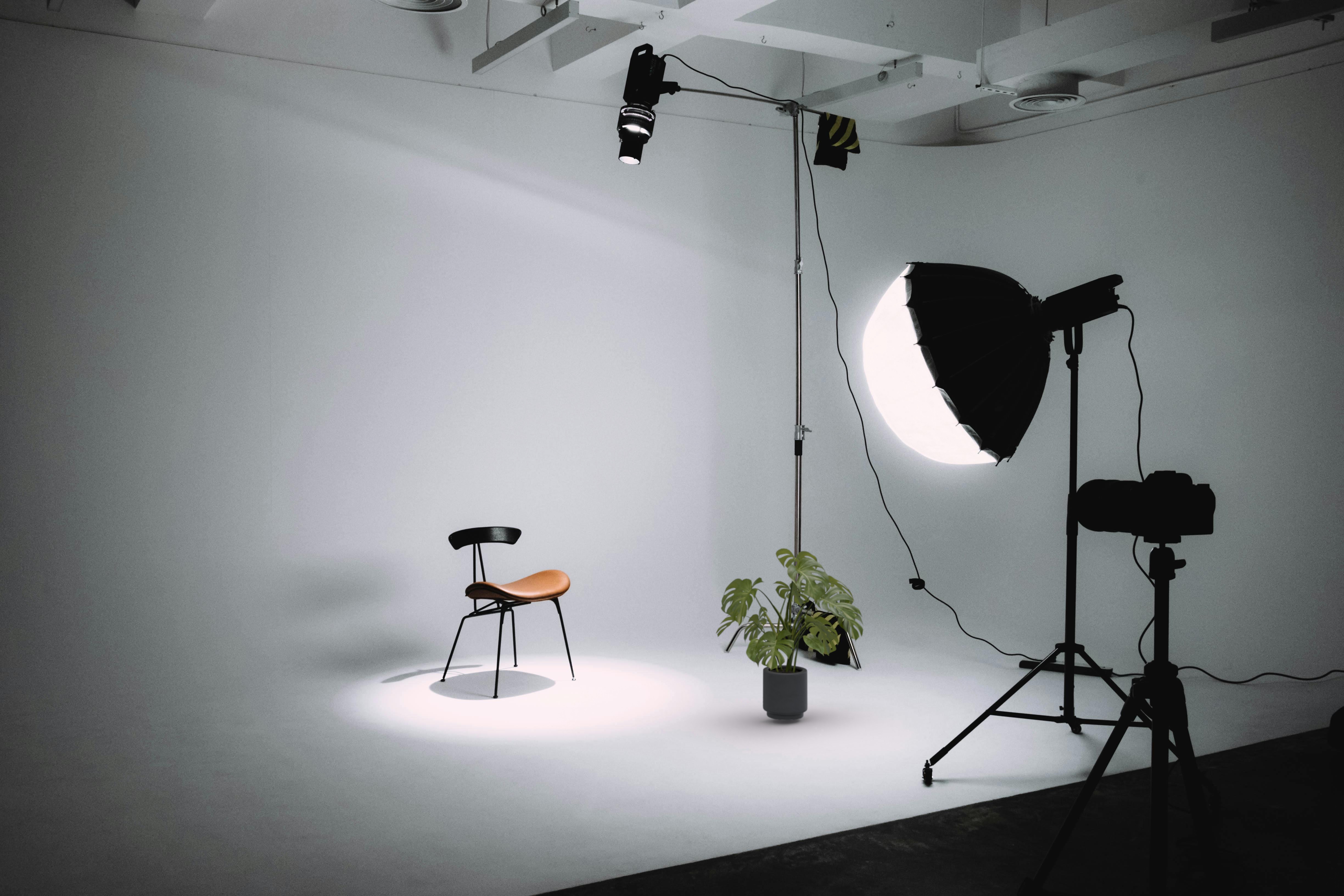}\\ [6pt]
    \rot{ \hspace{12pt} Frame 1} &
    \includegraphics[width=\tmplength]{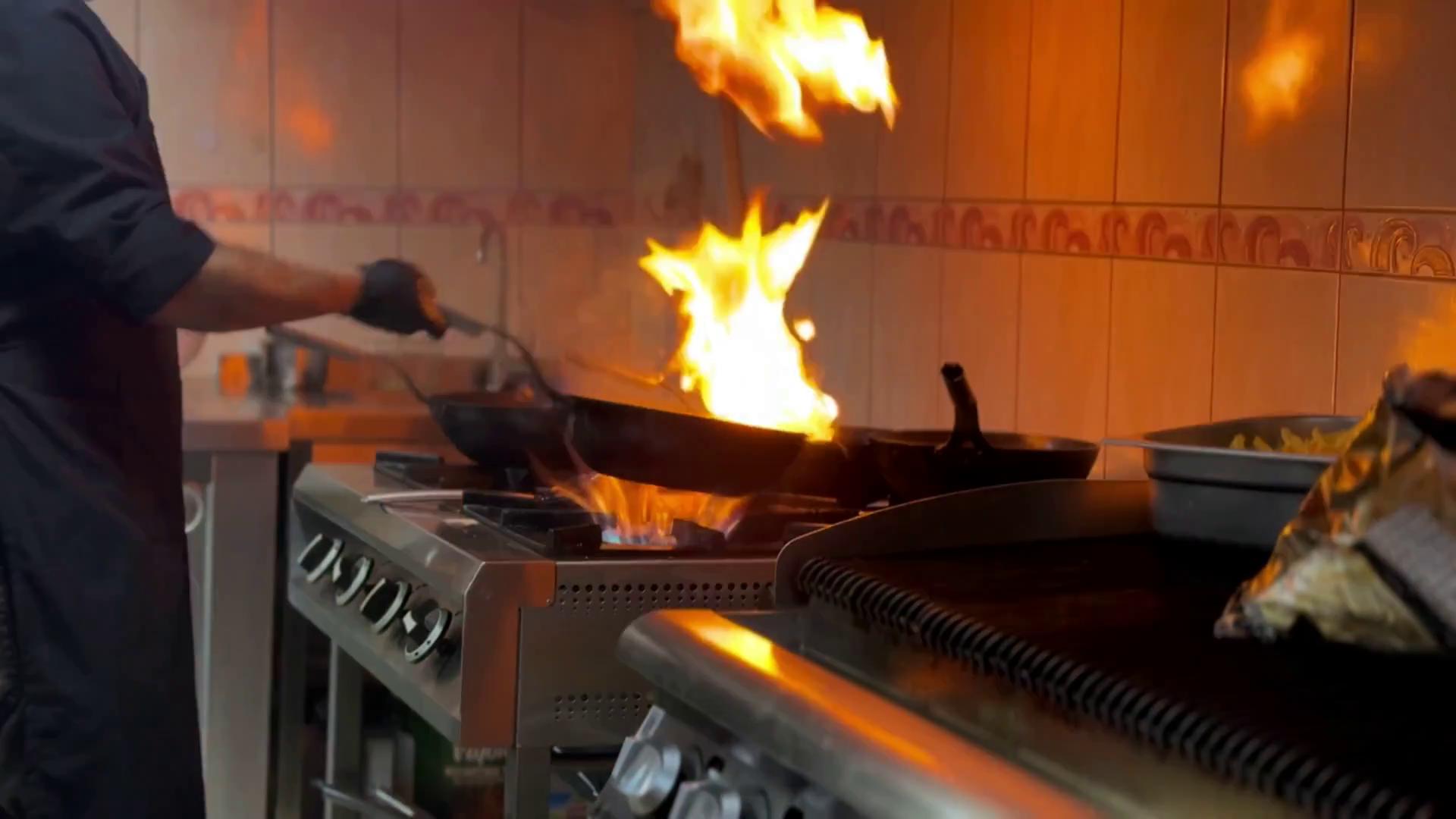} &
    \includegraphics[width=\tmplength]{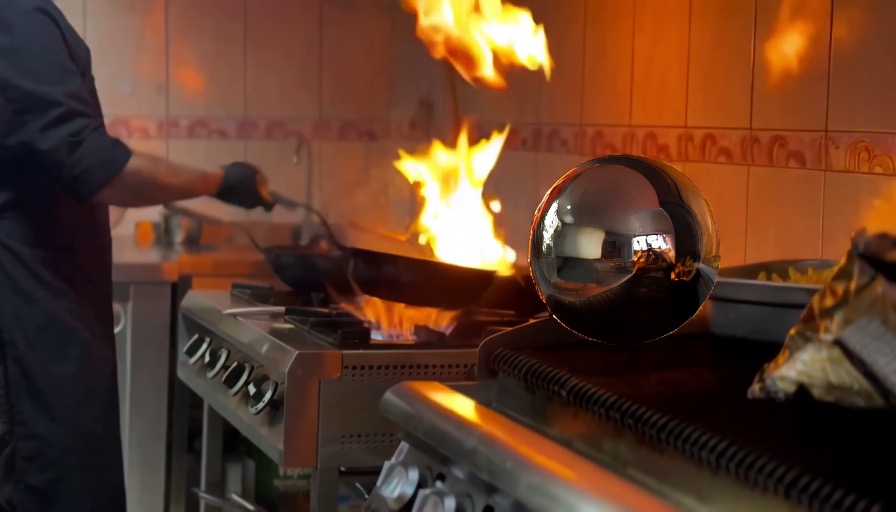}&
    \includegraphics[width=\tmplength]{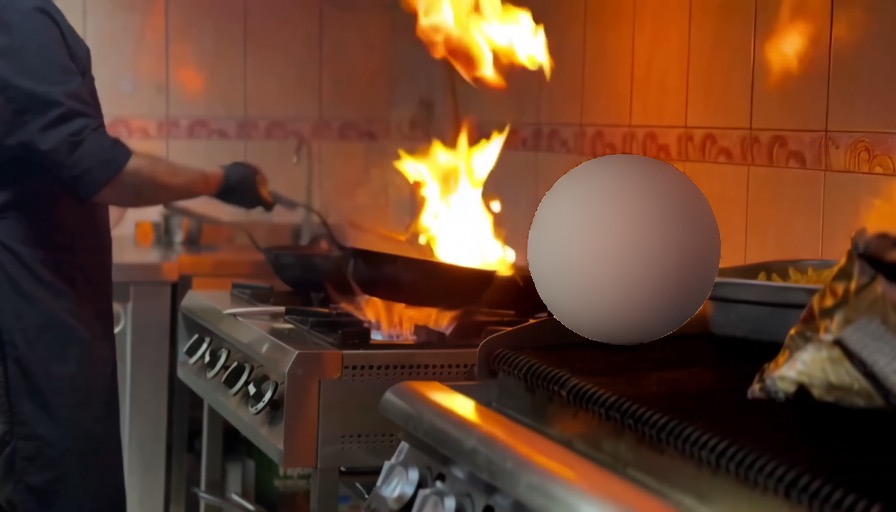}&
    \includegraphics[width=\tmplength]{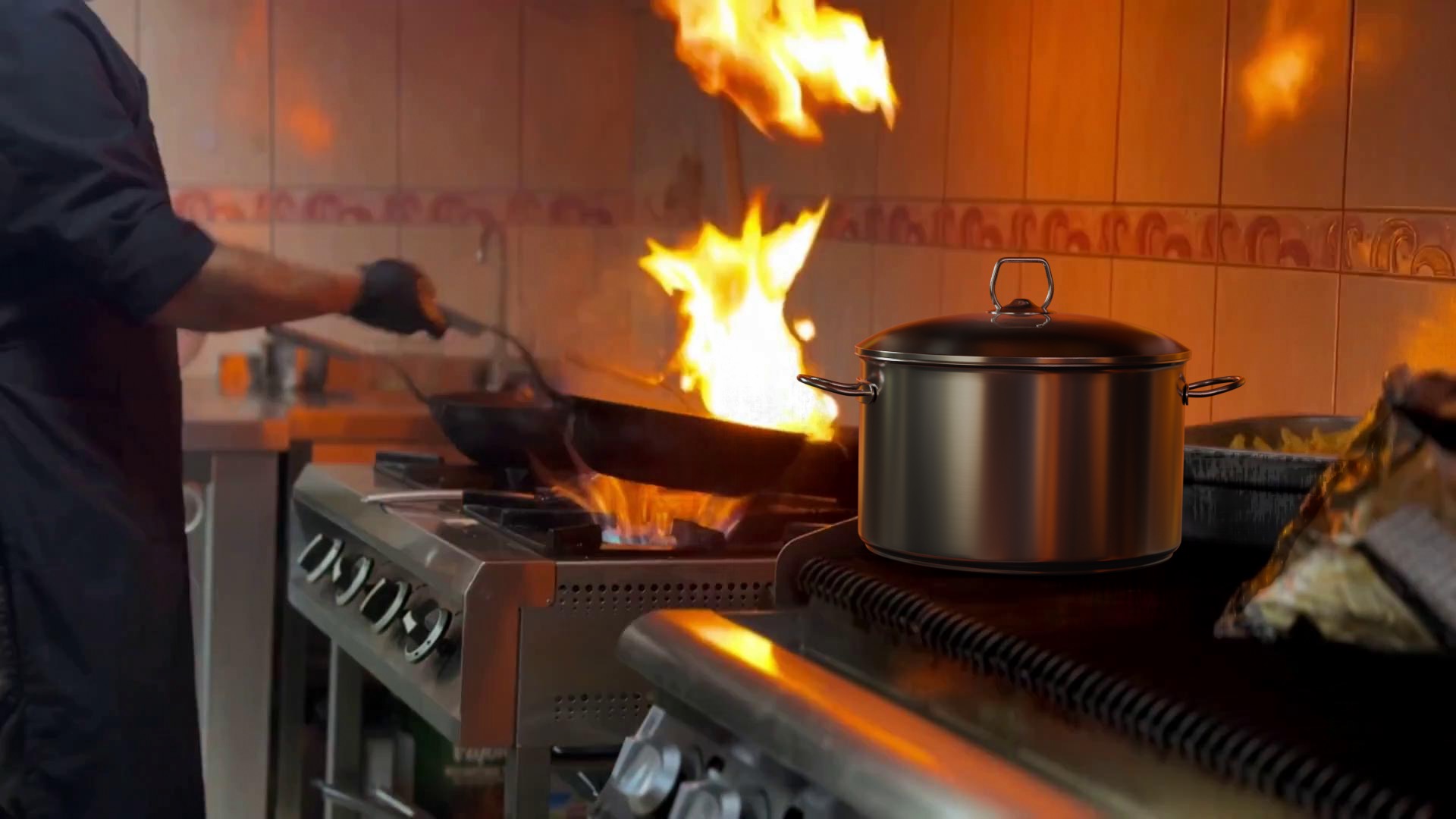}\\ \rot{ \hspace{12pt} Frame 2} &
    \includegraphics[width=\tmplength]{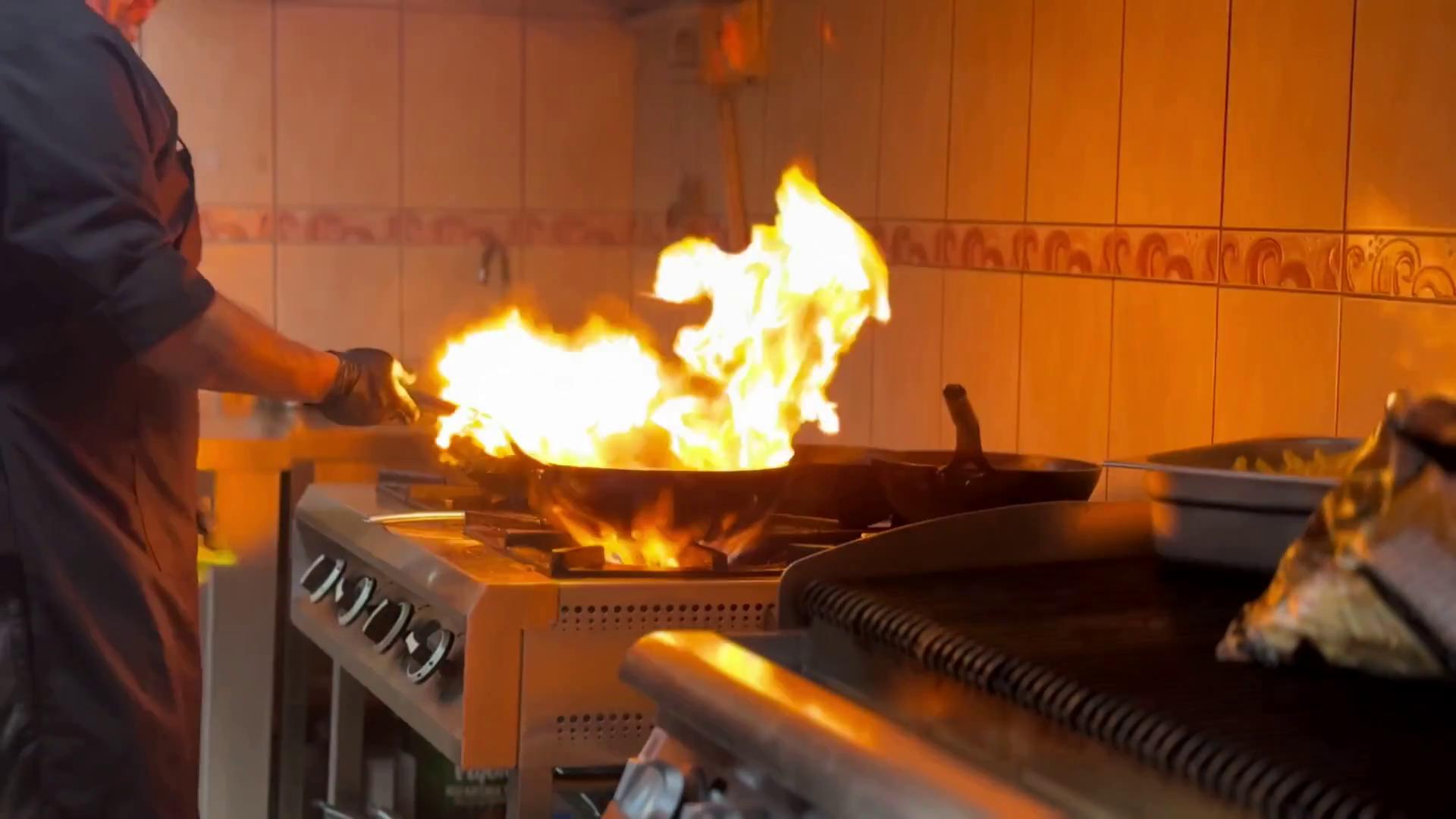} &
    \includegraphics[width=\tmplength]{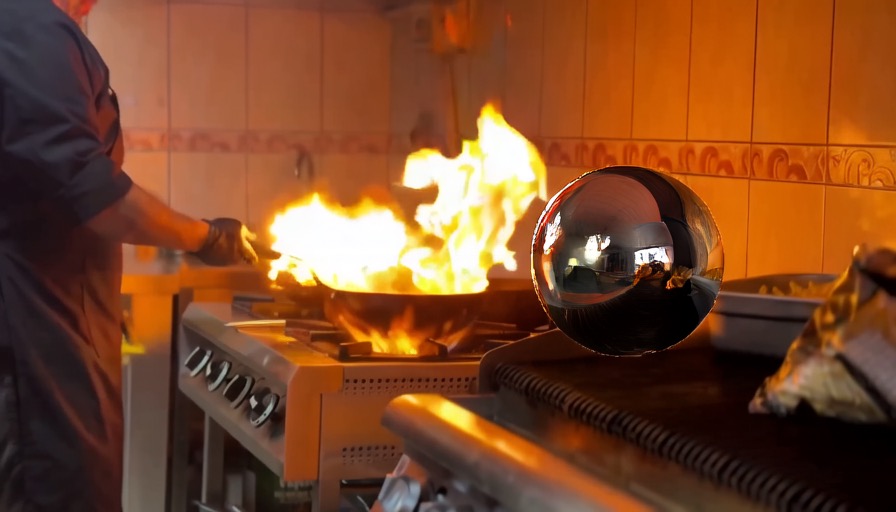}&
    \includegraphics[width=\tmplength]{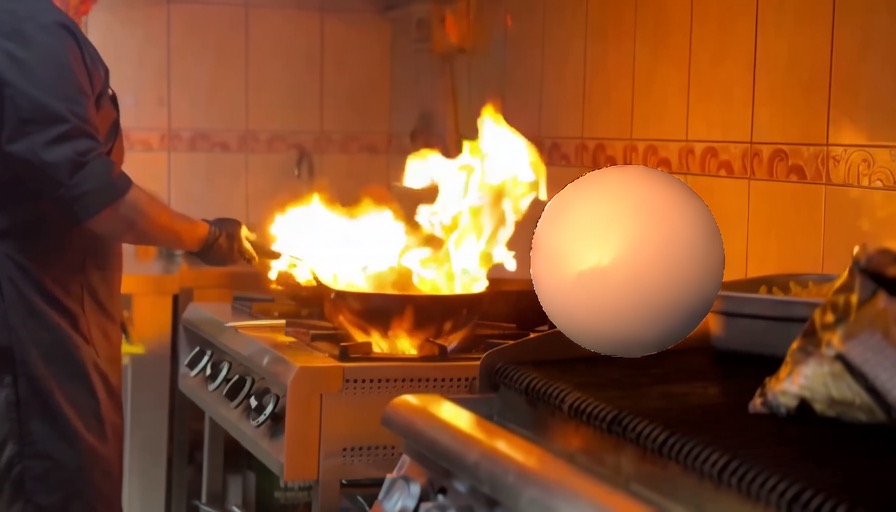}&
    \includegraphics[width=\tmplength]{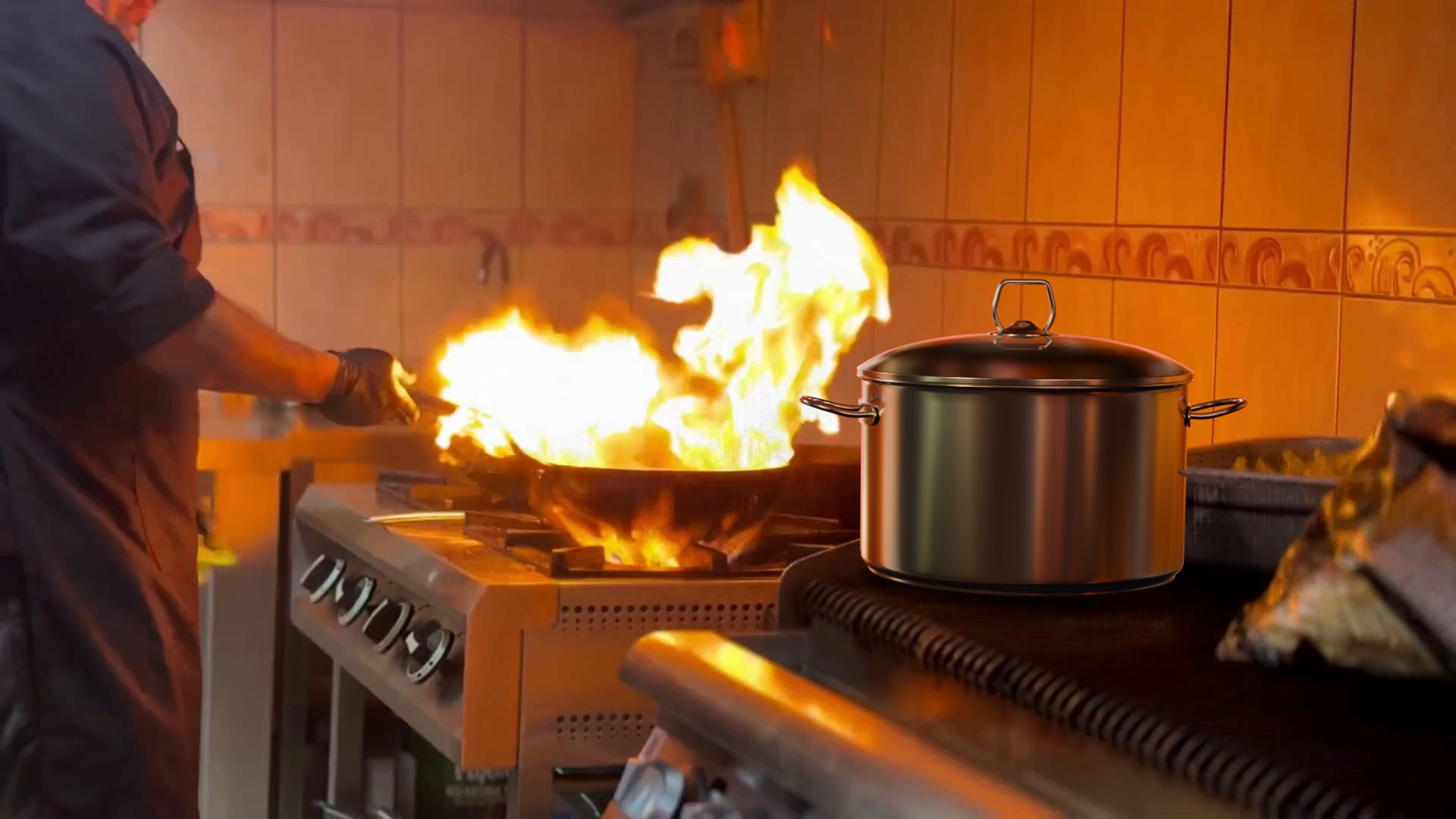}\\
    \rot{ \hspace{12pt} Frame 3} &
    \includegraphics[width=\tmplength]{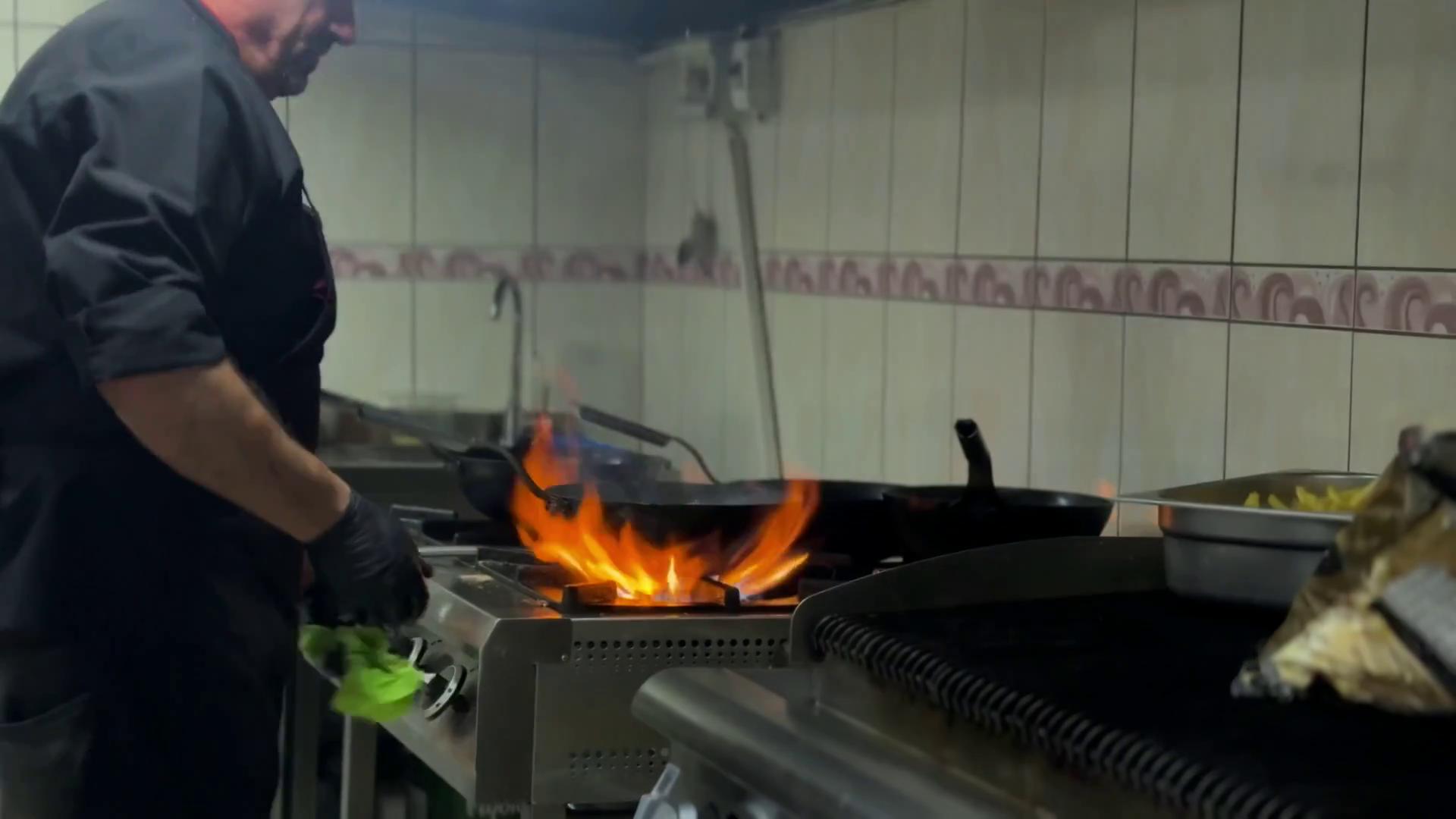} &
    \includegraphics[width=\tmplength]{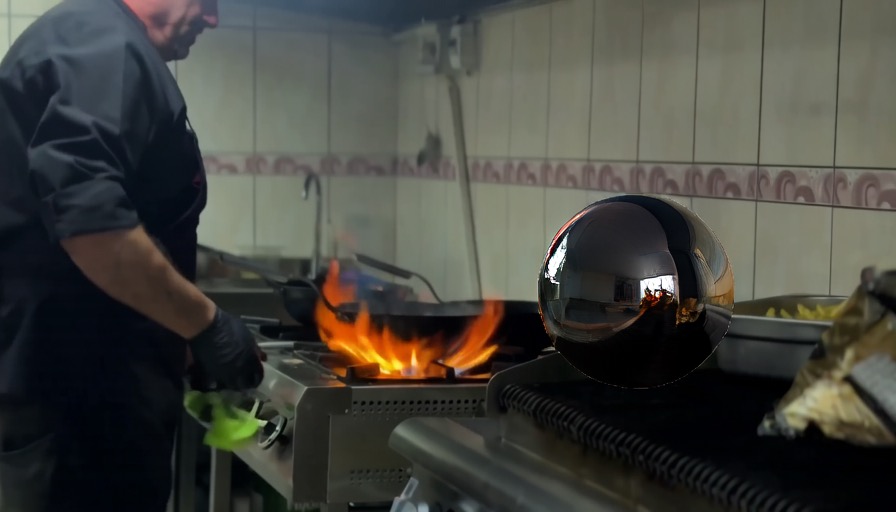}&
    \includegraphics[width=\tmplength]{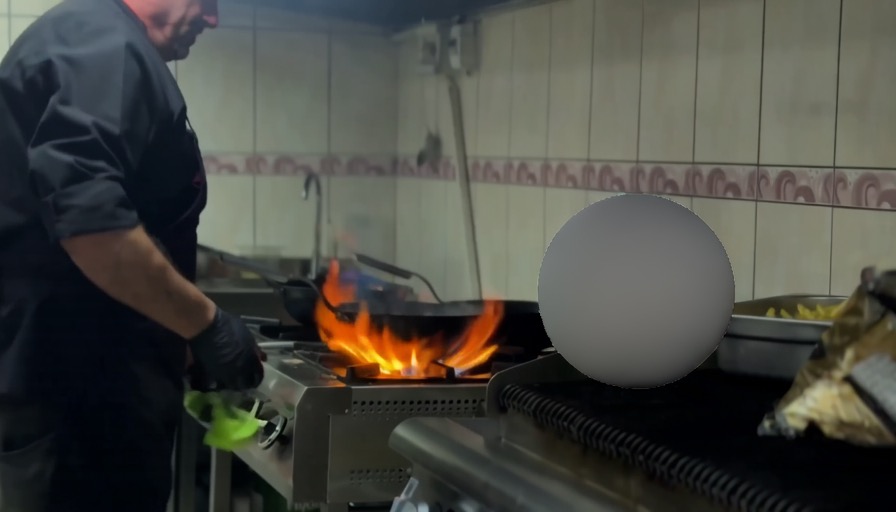}&
    \includegraphics[width=\tmplength]{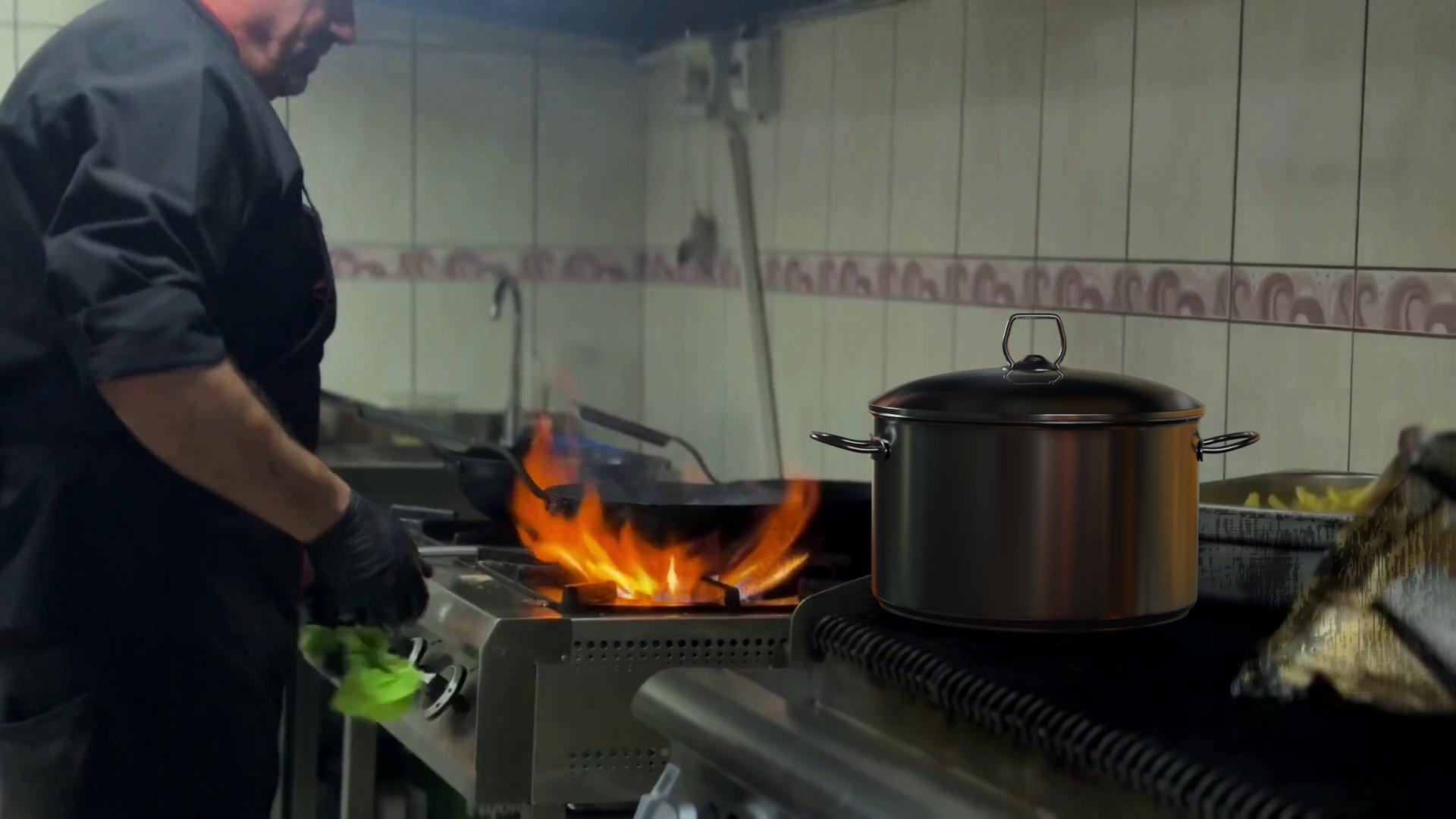}\\[6pt]
    \rot{ \hspace{18pt} Frame 1} &
    \includegraphics[width=\tmplength, trim={0 150 0 0}, clip]{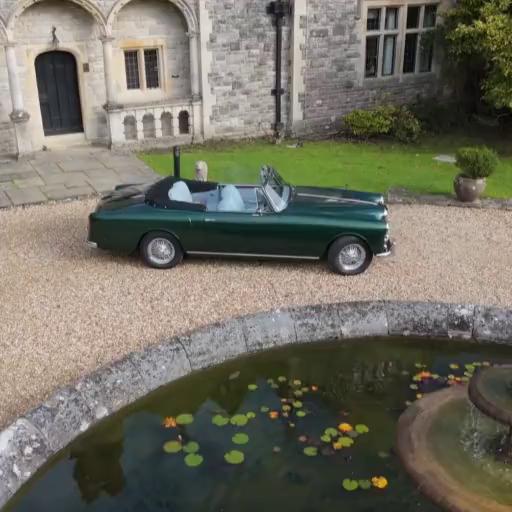} &
    \includegraphics[width=\tmplength, trim={0 150 0 0}, clip]{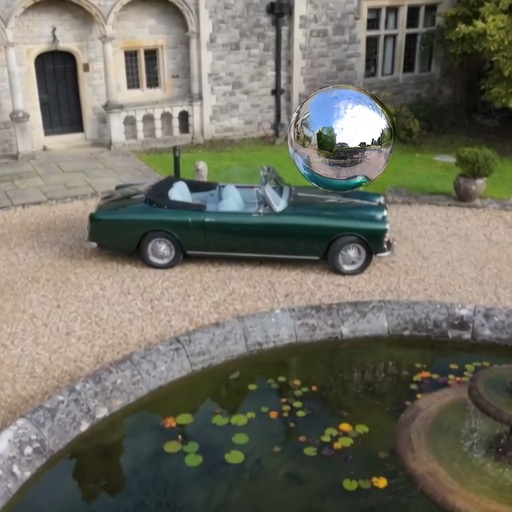}&
    \includegraphics[width=\tmplength, trim={0 150 0 0}, clip]{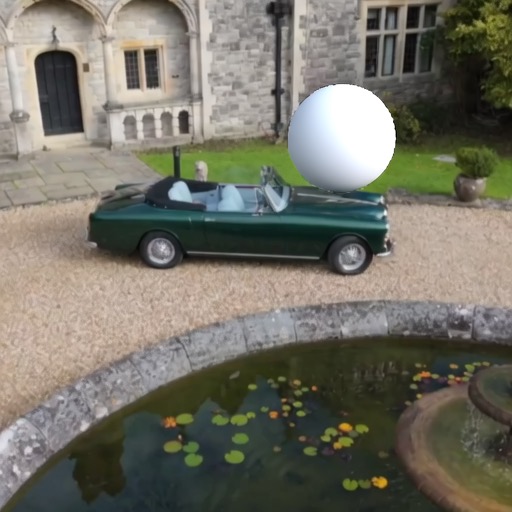}&
    \includegraphics[width=\tmplength, trim={0 150 0 0}, clip]{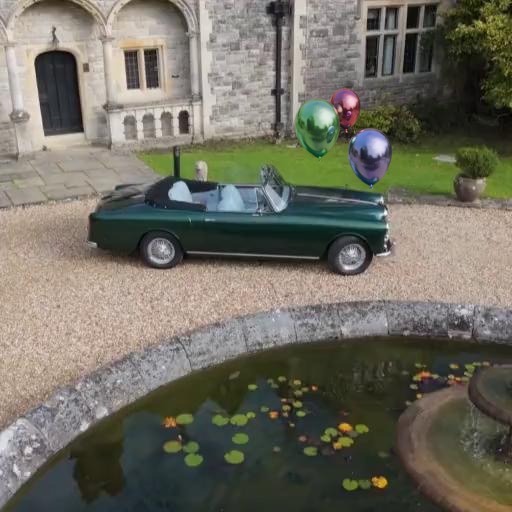}\\ \rot{ \hspace{18pt} Frame 2} &
    \includegraphics[width=\tmplength, trim={0 150 0 0}, clip]{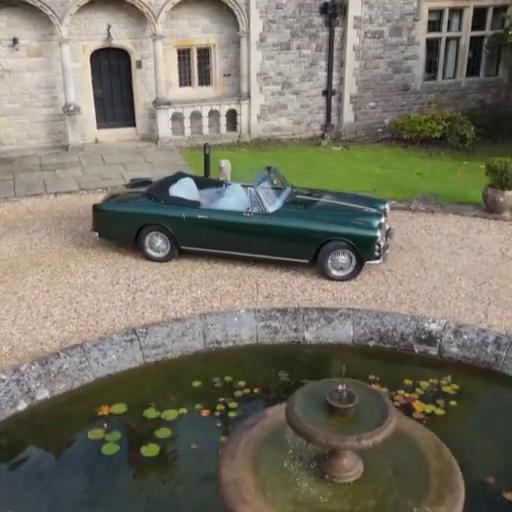} &
    \includegraphics[width=\tmplength, trim={0 150 0 0}, clip]{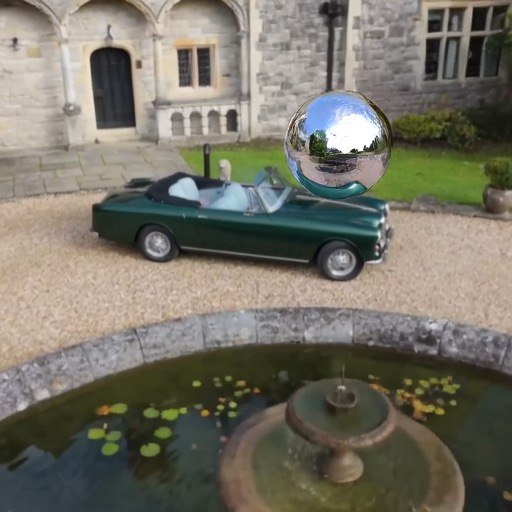}&
    \includegraphics[width=\tmplength, trim={0 150 0 0}, clip]{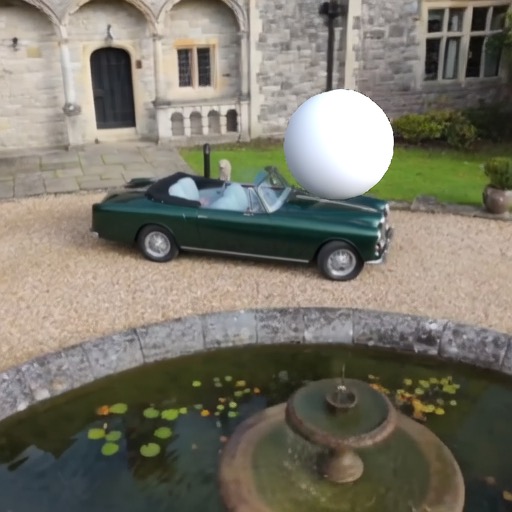}&
    \includegraphics[width=\tmplength, trim={0 150 0 0}, clip]{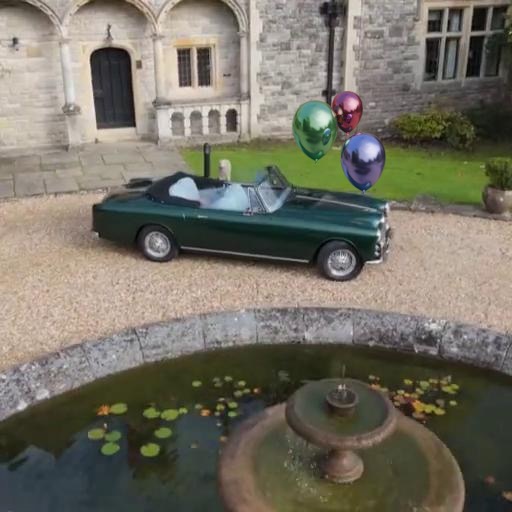}\\
    \rot{ \hspace{18pt} Frame 3} &
    \includegraphics[width=\tmplength, trim={0 150 0 0}, clip]{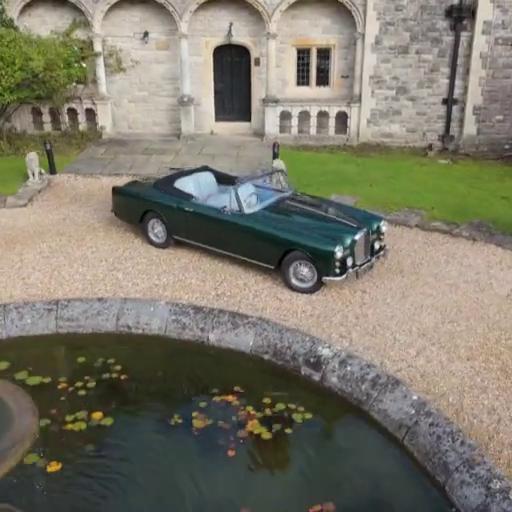} &
    \includegraphics[width=\tmplength, trim={0 150 0 0}, clip]{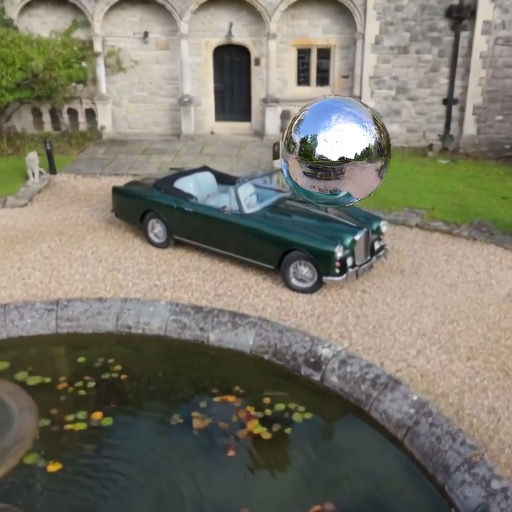}&
    \includegraphics[width=\tmplength, trim={0 150 0 0}, clip]{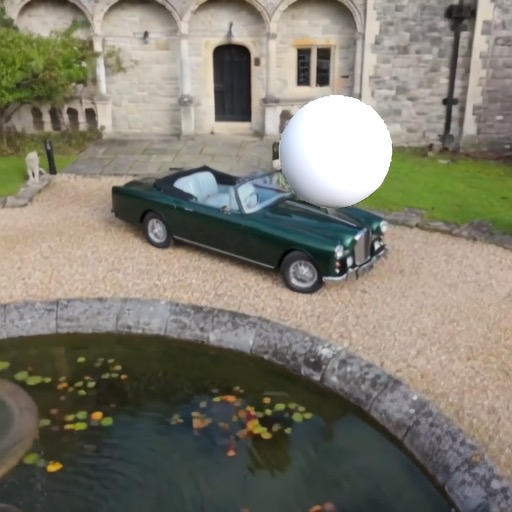}&
    \includegraphics[width=\tmplength, trim={0 150 0 0}, clip]{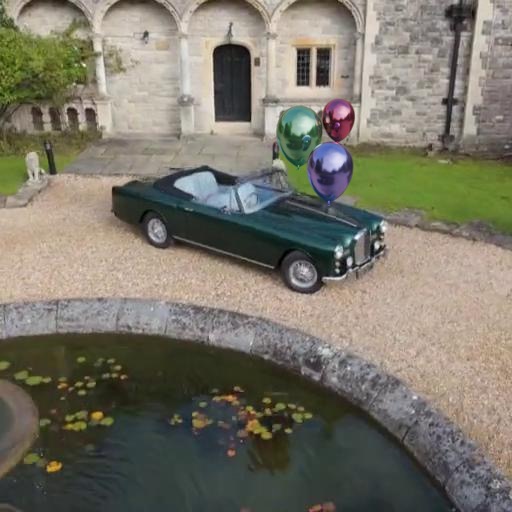}\\
    \end{tabular}
    \caption{Additional examples of our method on in-the-wild images and videos, with from left to right: the input frame, the predicted mirror sphere at EV0, the predicted diffuse sphere at EV0 and the inserted object. The predicted pointcloud is used as shadow catcher.} 
    \label{fig:single_in_the_wild}
\end{figure*}
More in-the-wild results are presented in \cref{fig:single_in_the_wild}. We make use of the predicted pointcloud from the FOV and depthmap as shadow catcher when inserting objects in the scene.

\begin{table*}
\centering
\footnotesize
\setlength{\tabcolsep}{1pt}
\begin{tabular}{lccccccccccccccccccc}
\midrule
& \multicolumn{4}{c}{RMSE$_\downarrow$}  & 
\multicolumn{4}{c}{Si-RMSE$_\downarrow$}  & \multicolumn{4}{c}{SSIM$_\uparrow$}   &  \multicolumn{4}{c}{Ang Err$_\downarrow$} \\
\midrule
 Method      & Mirr & Diff & Gloss & Mat  & Mirr & Diff & Gloss & Mat  & Mirr & Diff & Gloss & Mat  & Mirr & Diff & Gloss & Mat \\
 \midrule
   w/o Diffuse, Geo &   \third{\num{0.328}} & \third{\num{0.231}} & \third{\num{0.246}} & \third{\num{0.269}} & \num{0.651} & \third{\num{0.181}} & \num{0.201} & \num{0.310} & \num{0.796}  & \third{\num{0.960}} & \third{\num{0.943}} & \third{\num{0.941}} & \num{5.22} & \num{3.26} & \num{3.31} & \num{3.46}  \\
 w/o Diffuse   &   \second{\num{0.299}} & \best{\num{0.197}} & \best{\num{0.211}} & \best{\num{0.233}} & \third{\num{0.627}} & \best{\num{0.166}} & \second{\num{0.181}} & \second{\num{0.289}} & \second{\num{0.806}}  & \best{\num{0.969}} & \best{\num{0.953}} & \best{\num{0.952}} & \second{\num{4.81}} & \second{\num{2.95}} & \second{\num{2.97}} & \second{\num{3.14}}  \\
 w/o Geo  &   \num{0.333} & \num{0.238} & \num{0.253} & \num{0.277} & \second{\num{0.614}} & \num{0.182} & \third{\num{0.200}} & \third{\num{0.300}} & \third{\num{0.800}}  & \num{0.958} & \num{0.942} & \num{0.939} & \third{\num{4.95}} & \third{\num{3.05}} & \third{\num{3.09}} & \third{\num{3.35}}  \\
 \themethod (full)  &   \best{\num{0.297}} & \second{\num{0.201}} & \second{\num{0.215}} & \second{\num{0.235}} & \best{\num{0.599}} & \second{\num{0.168}} & \best{\num{0.180}} & \best{\num{0.283}} & \best{\num{0.813}}  & \second{\num{0.967}} & \best{\num{0.953}} & \second{\num{0.951}} & \best{\num{4.55}} & \best{\num{2.61}} & \best{\num{2.63}} & \best{\num{2.88}}  \\
\end{tabular}
\caption{\change{Ablation of the use of the added geometric maps $I_{\text{dir}}$ and $I_{\text{dist}}$ for predictions (see \cref{sec:conditions}) and diffuse sphere for HDRI optimization (see \cref{sec:general_approach}) on Laval Indoor SV with our image model. ``Mirr'' (mirror), ``Diff'' (diffuse), ``Gloss'' (glossy) and ``Mat'' (matte) refer to the different test spheres (see \cref{sec:metrics_datasets}).  Results are color coded by \best{best}, \second{second} and \third{third} best.}}
\label{tab:ablation_laval}
\end{table*}

\end{document}